\documentclass{article}

\usepackage[table]{xcolor}
\usepackage[final]{flageval_baai}
\usepackage[letterpaper,margin=1in]{geometry} %

\usepackage{amsmath}
\usepackage{tabularx}
\usepackage{booktabs}
\usepackage{longtable}
\usepackage{graphicx}
\usepackage{caption}
\usepackage{subcaption}

\usepackage{listings}
\usepackage[table]{xcolor}
\usepackage{wrapfig}
\usepackage{multirow}

\usepackage{tcolorbox}
\definecolor{lightgray}{HTML}{F0F0EB}
\definecolor{lightorange}{HTML}{FFD2A4}
\definecolor{lightblue}{HTML}{A4C7FF}
\definecolor{paleturquoise}{HTML}{AFEEEE}
\definecolor{lightgreen}{HTML}{A4FFAE}
\definecolor{stronggreen}{HTML}{3EFF54}
\definecolor{lightred}{HTML}{FFA4A4}
\definecolor{strongred}{HTML}{FF7171}

\lstdefinestyle{codeblock}{
  language=Python,
  basicstyle=\ttfamily\small,
  numbers=left,
  numberstyle=\tiny,
  stepnumber=1,
  numbersep=6pt,
  breaklines=true,
  frame=single,
  rulecolor=\color{black!20},
  backgroundcolor=\color{gray!5},
  tabsize=2,
  showstringspaces=false,
}

\usepackage{url}
\usepackage[colorlinks=true,
            linkcolor=blue,
            citecolor=blue,
            urlcolor=blue,
            filecolor=blue,
            breaklinks=true]{hyperref}
\usepackage{cleveref}

\fancyheadoffset[R,L]{0cm}

\title{ATOM-Bench: A Real-World Benchmark for Atomic Skills and Compositional Generalization in Manipulation Policies}

\author{%
\begin{minipage}{\textwidth}
\centering
\normalfont
\textbf{Zenan Wu}$^{1,2,*,\ddagger}$\quad
\textbf{Bingqing Wei}$^{1,2,*,\ddagger}$\quad
\textbf{Lu Liu}$^{1,*}$\quad
\textbf{Zheqi He}$^{1,\dagger}$\quad
\textbf{Xi Wang}$^{1}$\quad
\textbf{Jiakang Liu}$^{1}$\\[4pt]
\textbf{Zehui Li}$^{1,2}$\quad
\textbf{Guocai Yao}$^{1}$\quad
\textbf{Jing-Shu Zheng}$^{1}$\quad
\textbf{Xi Yang}$^{1}$\quad
\textbf{Yongtao Wang}$^{2}$\\[6pt]
$^{1}$Beijing Academy of Artificial Intelligence\qquad $^{2}$Peking University \\ [6pt]
\{liulu, zqhe\}@baai.ac.cn \\ [6pt]
\textbf{Homepage:} \url{https://flageval-baai.github.io/AtomBenchPage/}
\end{minipage}%
}

\newcommand{\blfootnote}[1]{%
  \begingroup
  \renewcommand{\thefootnote}{}%
  \footnotetext{#1}%
  \endgroup
}

\begin{document}
\maketitle
\blfootnote{$^{*}$Equal contribution.\quad $^{\dagger}$Corresponding author and project leader.\quad $^{\ddagger}$Work done during an internship at BAAI.}

\begin{abstract}
Generalist manipulation policies are increasingly presented as foundation models for robotic control, but their real-world generalization remains difficult to diagnose. A policy may succeed on demonstrated tasks while still failing to execute fine-grained atomic skills or recombine learned skills in new task structures. We introduce \textbf{ATOM-Bench}, a real-world benchmark for evaluating both atomic skills and compositional generalization in manipulation policies. ATOM-Bench factorizes tabletop manipulation into motor atoms and instruction atoms, and contains 30 atomic tasks and 24 held-out compositional tasks across paired single-arm and dual-arm robot tracks. We collect 3,000 human demonstrations for atomic fine-tuning and release both the demonstration data and evaluation rollout data to support reproducible real-world evaluation. Policies are fine-tuned on atomic tasks and evaluated on both atomic skill acquisition and held-out compositional tasks. We further introduce Atomic Score (AS) and Compositional Failure Share (CFS) to distinguish failures caused by weak atomic skills from failures caused by limited compositional reuse. Through 2,700 physical rollouts on five representative manipulation policies, we find that current policies can acquire simple instruction-grounding skills, but still struggle with fine-grained motor atoms, counting, and logical filtering. More importantly, strong atomic performance does not reliably transfer to held-out compositional tasks. ATOM-Bench provides a diagnostic testbed for studying whether failures arise from weak motor execution, poor instruction grounding, or limited compositional reuse.

\end{abstract}

\section{Introduction}
	
Recent manipulation policies~\citep{black2025pi0,physicalintelligence2025pi05,wu2026pragmatic,cai2026xiaomi,bi2025motusunifiedlatentaction,ye2026worldactionmodelszeroshot} are increasingly presented as foundation models for robotic control. They have demonstrated impressive behaviors across common manipulation, long-horizon instruction following, and multi-task learning. These systems suggest an appealing possibility: a robot policy may acquire reusable physical and semantic skills, and then recombine them to solve new tasks. However, this promise remains difficult to verify on real robots. Strong demonstrations, high simulation scores, or successful task-specific fine-tuning do not necessarily show that a policy has learned composable skills rather than task templates or dataset-specific regularities.

Existing benchmarks make this verification difficult, revealing a growing evaluation tension for manipulation foundation models. Simulation benchmarks provide scale, controllability, and fast iteration, but their large programmatically generated task distributions can be rapidly absorbed by high-capacity policies, while their physical challenges remain less demanding than real-world contact, perception, and execution. As a result, high simulation success increasingly provides limited evidence of deployable real-robot generalization. Real-world evaluation gives stronger evidence, but it is expensive, difficult to automate, and hard to scale. Recent real-robot efforts make important progress: RoboArena \citep{atreya2025roboarena} studies distributed pairwise policy ranking, AutoEval \citep{zhouautoeval} reduces human involvement with learned success classifiers and reset policies, RoboChallenge \citep{yakefu2025robochallenge} builds an online evaluation system for scalable real-robot testing, and GM-100 \citep{wang2026great} expands real-robot task diversity with 100 detail-oriented tasks. However, these benchmarks primarily evaluate policies over task suites or ranking protocols. They provide limited control over the skills a policy has seen during adaptation, making it difficult to determine whether a model has learned reusable atomic skills that can be transferred to unseen compositional tasks. ATOM-Bench targets this diagnostic gap by controlling the atomic training distribution and evaluating both atomic skill acquisition and held-out compositional generalization on real robots.

We focus on two related questions. First, can current manipulation policies acquire reliable atomic skills, including both physical motor skills and language-conditioned instruction skills? Second, after these atomic skills are learned, can they be recombined to solve held-out compositional tasks that were never demonstrated during fine-tuning? This setting is more diagnostic than broad task-suite averages, because it lets us pinpoint \emph{where} a policy fails rather than only \emph{that} it fails.

\begin{figure}[t]
\centering
\includegraphics[width=\linewidth]{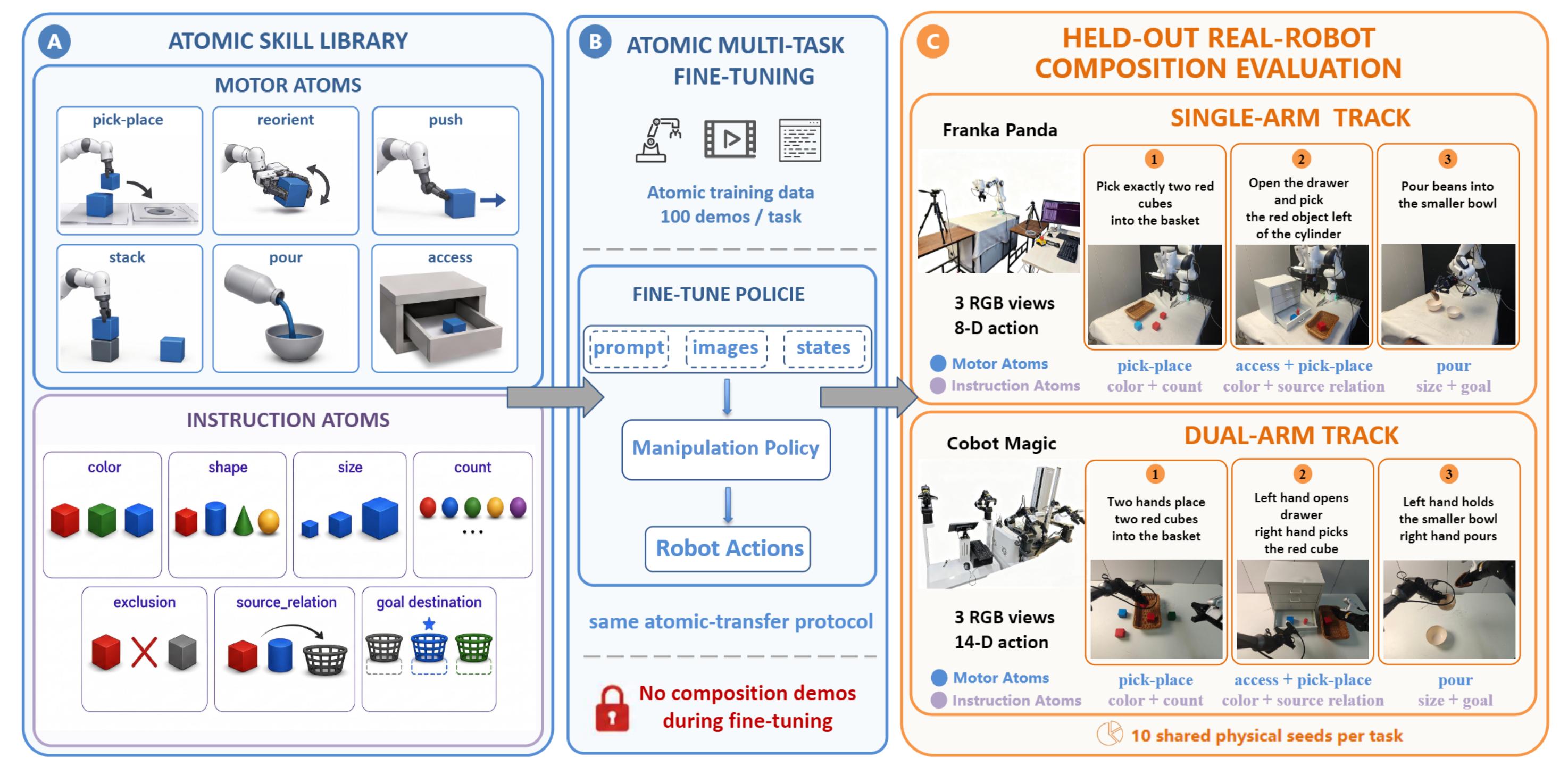}
\caption{
    \textbf{Overview of ATOM-Bench.}
    Tabletop manipulation tasks are factorized into \emph{motor atoms} and \emph{instruction atoms}. Policies are fine-tuned on atomic tasks and evaluated on both atomic skill acquisition and held-out compositional generalization.
}
\label{fig:intro}
\end{figure}
To study these questions, we introduce \textbf{ATOM-Bench}, a real-world benchmark for atomic skills and compositional generalization in manipulation policies. As illustrated in Figure~\ref{fig:intro}, ATOM-Bench factorizes tabletop manipulation into \emph{motor atoms}, which describe physical operations, and \emph{instruction atoms}, which describe semantic or logical constraints in language. These atoms are used to build atomic tasks for adaptation and held-out compositional tasks for evaluation. Policies are fine-tuned on atomic tasks and then evaluated on both the atomic tasks themselves and held-out compositional tasks that combine multiple motor and instruction atoms. We further introduce Atomic Score (AS) and Compositional Failure Share (CFS) to separate failures caused by weak atomic skills from failures caused by compositional failures. ATOM-Bench contains two paired real-world robot tracks: a single-arm track on Franka Panda and a dual-arm track on Cobot Magic. Each dual-arm task is designed as the bimanual counterpart of a single-arm task, adding hand-role assignment, inter-arm coordination, and ordered action sequences.

Our experiments show that atomic skills and compositional generalization remain challenging in different ways. Atomic tasks that require precise control, such as pouring, reorientation, and pushing, are difficult for current policies. Instruction atoms that require set-level reasoning, such as counting and exclusion, also remain challenging. Even when policies acquire useful atomic skills, their performance drops sharply on held-out compositional tasks. These results suggest that real-world generalization in manipulation policies remains far from solved.
Our contributions are:
\begin{itemize}
    \item We introduce \textbf{ATOM-Bench}, a real-world robot benchmark with 3,000 human demonstrations across two robotic platforms.
    \item We design a paired two-track task suite with 30 atomic tasks and 24 compositional tasks, covering single-arm object-centric manipulation and dual-arm coordinated manipulation.
    \item We introduce AS and CFS to distinguish atomic-skill deficits from compositional failures.
    \item We evaluate five representative manipulation policies through 2,700 physical rollouts and show that both atomic skills and compositional transfer remain challenging.
\end{itemize}

\section{Related Work}
\textbf{Simulated and compositional manipulation benchmarks.}
Simulation benchmarks provide scalable and controllable evaluation for robot learning. 
VIMA-Bench~\citep{jiang2023vima}, LIBERO~\citep{liu2023libero}, and VLABench~\citep{zhang2024vlabench} study multi-task learning, language-conditioned manipulation, long-horizon execution, and knowledge-intensive reasoning in simulated environments. 
Recent benchmarks further show that strong VLA performance can hide memorization and weak compositional generalization: VLA-Arena~\citep{zhang2025vlaarena}, LIBERO-PRO~\citep{zhou2025liberopro}, and RoboHiMan~\citep{chen2025robohiman} evaluate policies under compositional, visual, language, or task perturbations. 
These works motivate ATOM-Bench, but they mainly evaluate in simulation or digital-twin settings. 
ATOM-Bench instead evaluates both atomic skill execution and compositional generalization directly on physical robots.

\textbf{Real-world manipulation benchmarks.}
Real-world benchmarks evaluate policies directly on physical robots. 
RoboArena~\citep{atreya2025roboarena} performs distributed pairwise policy comparison, AutoEval~\citep{zhouautoeval} automates success scoring and scene resets, RoboChallenge~\citep{yakefu2025robochallenge} provides large-scale online evaluation, and GM-100~\citep{wang2026great} broadens real-robot task coverage. 
SceneReplica~\citep{khargonkar2024scenereplica} and FurnitureBench~\citep{heo2023furniturebench} emphasize reproducible scenes and long-horizon assembly. 
These efforts improve scalability, automation, ranking, reproducibility, or task diversity. 
In contrast, ATOM-Bench fine-tunes policies on controlled atomic tasks and evaluates them on both atomic and held-out compositional tasks, revealing failures in atomic execution and compositional reuse.

\textbf{Generalist manipulation policies.}
Generalist manipulation policies such as the $\pi$ series~\citep{black2025pi0,physicalintelligence2025pi05,physicalintelligence2025pistar06,physicalintelligence2026pi07}, OpenVLA~\citep{kim2025openvla}, LingBot-VLA~\citep{wu2026pragmatic}, GR00T~\citep{gr00tn1_2025,gear2025gr00tn16}, SmolVLA~\citep{shukor2025smolvla}, Motus~\citep{bi2025motusunifiedlatentaction}, and DreamZero~\citep{ye2026worldactionmodelszeroshot}, have shown increasingly strong performance on language-conditioned robot tasks.
However, it remains unclear whether their generalization comes from reusable physical and semantic skills or from task templates and dataset-specific regularities. 
ATOM-Bench provides a controlled real-world protocol for this question by jointly measuring atomic-task performance and held-out compositional performance under the same adaptation setting.

\section{ATOM-Bench}

\subsection{Task Factorization and Task Groups}
\label{sec:task_factorization}

ATOM-Bench uses atoms as a practical abstraction for controlled real-robot evaluation.
We choose atoms by summarizing recurring task factors in prior manipulation benchmarks~\citep{yu2019metaworld,nasiriany2024robocasa,luo2025fmb,wang2025roboeval,paulius2019manipulation,paulius2020motion,srivastava2022behavior,zhang2024vlabench} and referring-expression grounding datasets~\citep{kazemzadeh2014referitgame,mao2016refcocog,yu2016modelingcontext}.
The selected atoms are common in manipulation tasks, expose different physical or semantic failure modes, can be reliably instantiated on real robots, and can be recombined into held-out compositional tasks.
The full taxonomy and design rationale are provided in Appendix~\ref{app:atom_taxonomy}.

We divide these factors into two types: \emph{motor atoms} and \emph{instruction atoms}.
Motor atoms specify the physical operation required by the task, including pick-and-place, reorientation, pushing, stacking, pouring, or articulated-object access.
Instruction atoms specify how the target object or destination is described in language, including color, shape, size, source relation, goal destination, counting or exclusion.
Instruction atoms are always evaluated through executable robot tasks, typically with pick-and-place or its bimanual counterpart as the carrier action.

Based on these atoms, each platform contains 7 instruction-atom tasks, 8 motor-atom tasks, and 12 held-out compositional tasks.
The single-arm and dual-arm task suites are paired: each Cobot Magic task is designed as the bimanual counterpart of a Franka Panda task, with additional hand-role assignment and inter-arm coordination.
The \emph{Motor Set} tests one physical skill at a time with minimal semantic ambiguity.
The \emph{Instruction Set} tests one language constraint at a time under a simple carrier action.
The \emph{Composition Set} contains held-out tasks that combine one or more motor atoms with multiple instruction atoms.

\subsection{Data Collection}
\label{sec:data_collection}

\begin{figure}[t]
\centering
\includegraphics[width=0.5\linewidth]{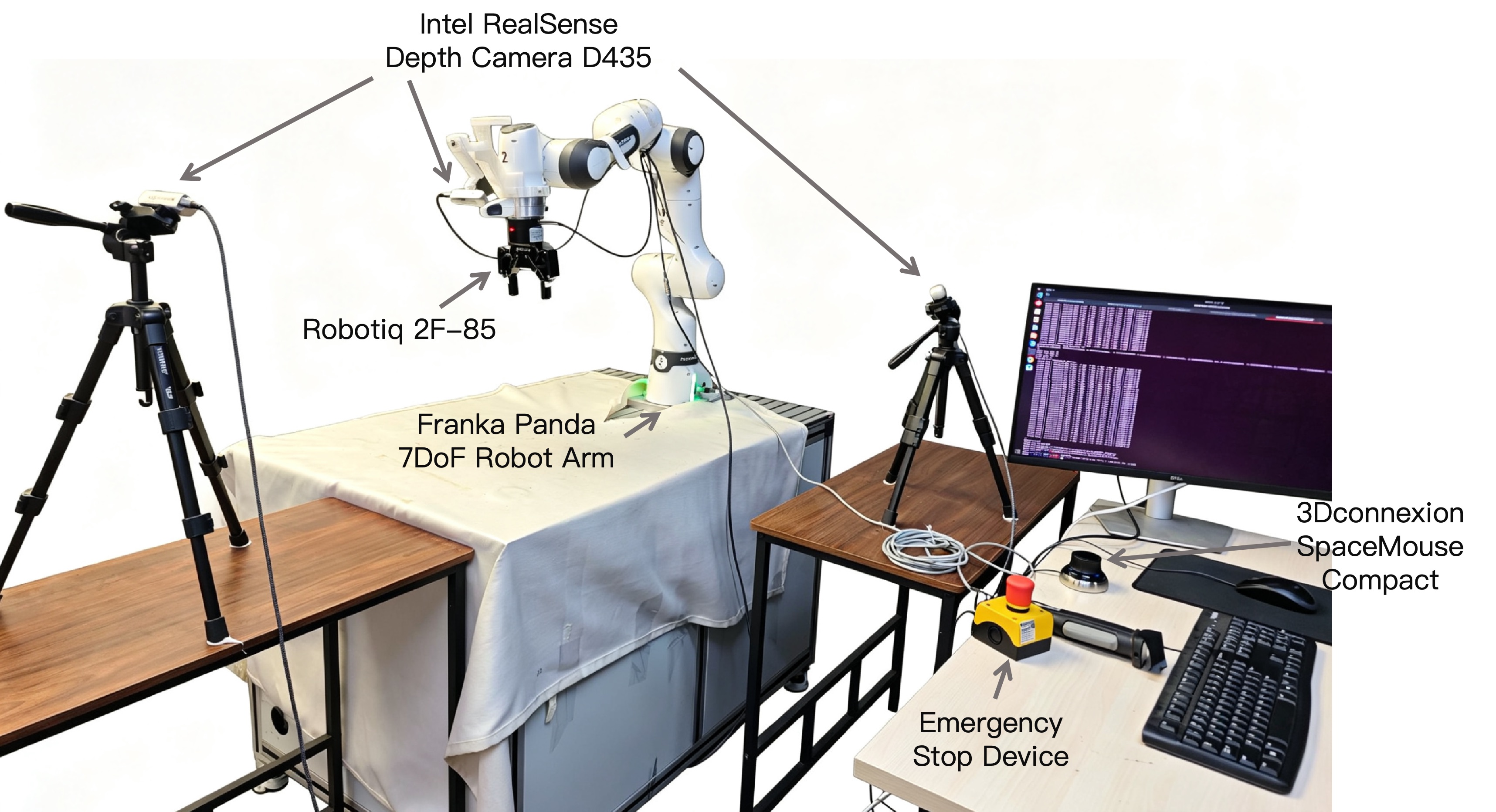}\hfill
\includegraphics[width=0.48\linewidth]{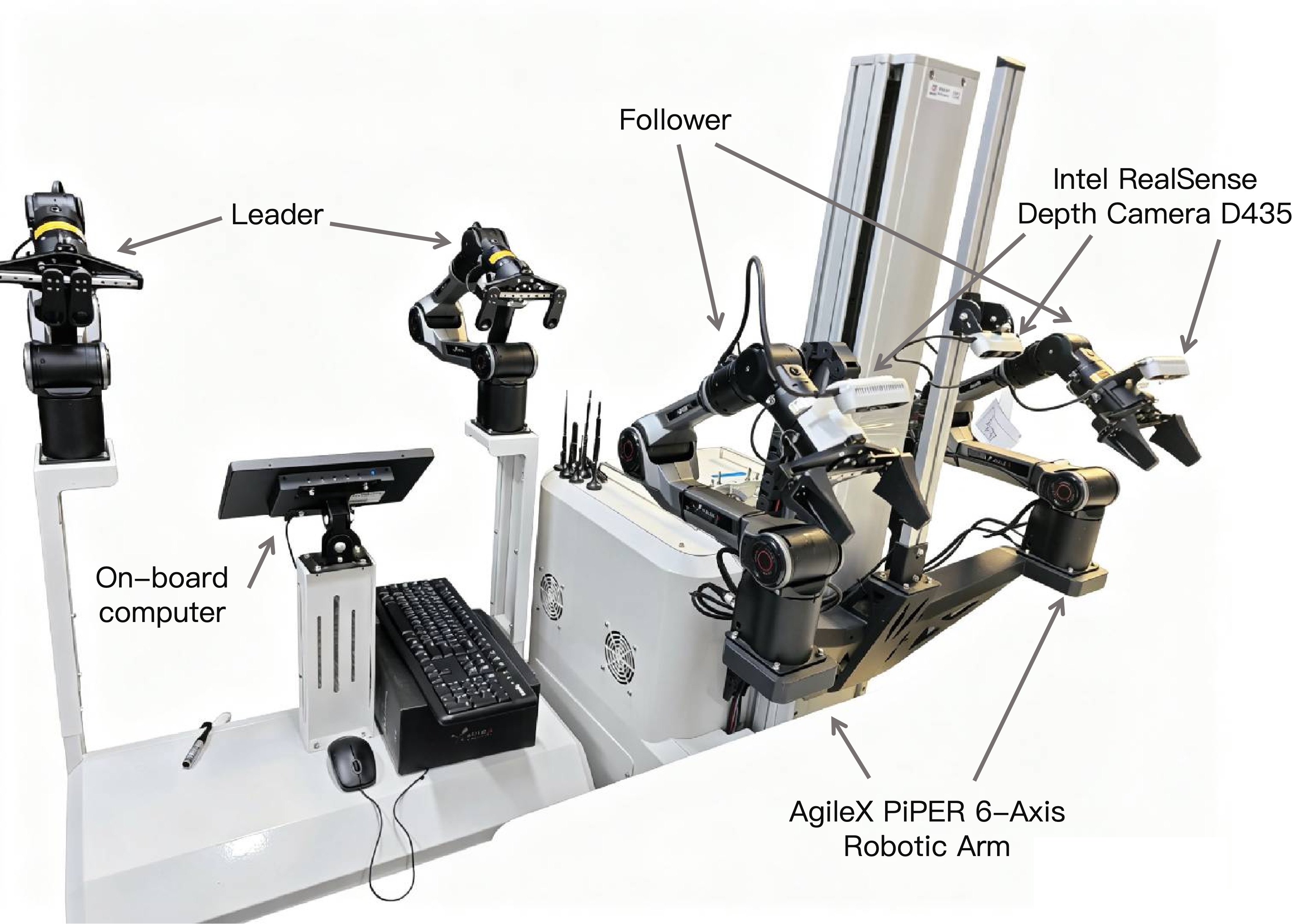}
\caption{Data collection platforms used in ATOM-Bench. \textbf{Left:} the Franka Panda with single-arm. \textbf{Right:} the Agilex Cobot Magic with dual-arm.}
\label{fig:platforms}
\end{figure}

We collect data on two robotic platforms: Franka Panda for single-arm tasks and Agilex Cobot Magic for dual-arm tasks. As shown in Figure~\ref{fig:platforms}, the Franka Panda setup uses a 7-DoF Franka Panda arm with a Robotiq 2F-85 gripper and three Intel RealSense cameras mounted at the left, front, and wrist viewpoints. The wrist camera is attached with a 3D-printed mount following~\citep{taunyazov2020event}.

The Agilex Cobot Magic setup is an open-source dual-arm system built upon Mobile ALOHA~\citep{fu2024mobile}. It supports coordinated control of two robotic arms and uses three Intel RealSense cameras mounted at the left-arm, right-arm, and front viewpoints.

For each atomic task, we collect 100 expert demonstrations through human teleoperation. Franka Panda is controlled with a 3D SpaceMouse, while Agilex Cobot Magic is teleoperated through its master arms. During teleoperation, we record synchronized observations and robot actions at 30 Hz. In total, we collect approximately 9 hours of demonstrations on Franka Panda and 12.5 hours on Agilex Cobot Magic.

\subsection{Metrics}
\label{sec:metrics}

ATOM-Bench uses five metrics: Success Rate (SR), Process Success Rate (PSR), Atomic Score (AS), Compositional Failure Share (CFS), and Transfer Gap (TG).
Each task is evaluated over 10 shared physical test seeds, and results are macro-averaged across tasks.

\textbf{Success Rate.}
Success Rate (SR) is the primary task-completion metric. 
A rollout is counted as successful only if it satisfies the task completion predicate. 
SR is the percentage of successful rollouts among all evaluation trials.

\textbf{Process Success Rate.}
Since real-robot failures are often partial, PSR captures intermediate progress.
Each task is annotated with a small set of score points (e.g., select the correct object, reach the target); PSR is the per-rollout fraction of points achieved, averaged across rollouts.

\textbf{Atomic Score.}
For a compositional task $X$, let $A(X)$ be the set of its participating motor and instruction atoms. 
We define the Atomic Score (AS) as the mean PSR of these atoms,
\[
\mathrm{AS}(X)=\frac{1}{|A(X)|}\sum_{a \in A(X)} \mathrm{PSR}(a).
\]
When an atom has multiple representative atomic tasks, $\mathrm{PSR}(a)$ is the mean PSR of those tasks.
AS measures how well the policy performs the required atoms in isolation.

\textbf{Compositional Failure Share.}
We measure how much of the compositional failure cannot be explained by weak atomic skills using
\[
\mathrm{CFS}(X)=\frac{\max(0,\mathrm{AS}(X)-\mathrm{PSR}(X))}{1-\mathrm{PSR}(X)}.
\]
A low CFS means that the failure is mostly explained by weak atomic skills, while a high CFS means that the required atoms are learned in isolation but fail after composition.

\textbf{Transfer Gap.}
TG is the SR difference between single-task and multi-task atomic fine-tuning.
Positive TG favors task specialization, while negative TG favors joint atomic training.

\section{Experiments}
We design our experiments to evaluate generalist manipulation policies on two axes: atomic skill acquisition and compositional generalization, under a shared real-world adaptation protocol. Our experiments answer four questions:
(1) How well do current policies acquire atomic motor and instruction skills on real robots?
(2) How well do those acquired atomic skills compose into unseen held-out tasks?
(3) What are the dominant failure modes of current policies?
(4) How does atomic-transfer fine-tuning compare with single-task fine-tuning?

\subsection{Experimental Setup}
We evaluate five manipulation policies in ATOM-Bench: Pi0.5, LingBot-VLA, GROOT N1.6, SmolVLA, and Motus. Each policy is evaluated on both robotic platforms described in Section~\ref{sec:data_collection}.

We use two fine-tuning protocols. \emph{Atomic-transfer fine-tuning} is the primary protocol: a single policy is jointly fine-tuned on all 15 atomic tasks and evaluated on both the atomic tasks and the held-out Composition Set. \emph{Single-task atomic fine-tuning} trains an independent checkpoint on the 100 demonstrations of one atomic task and evaluates it only on that task, serving as a specialization baseline. Because this requires one training run per atomic task, we report it only for Pi0.5. We also evaluate robustness with perturbations; details are provided in Appendix~\ref{app:perturbation}. 

For each task, policies are evaluated under 10 fixed physical test seeds with shared initial object placements. We reproduce each seed using a mask-guided placement procedure, as detailed in Appendix~\ref{app:seed_placement}. Hyperparameters and training details are listed in Appendix~\ref{app:sft_config}.

\subsection{Atomic Skill Acquisition}
\label{sec:rq1}

\begin{figure*}[t]
\centering
\includegraphics[width=\linewidth]{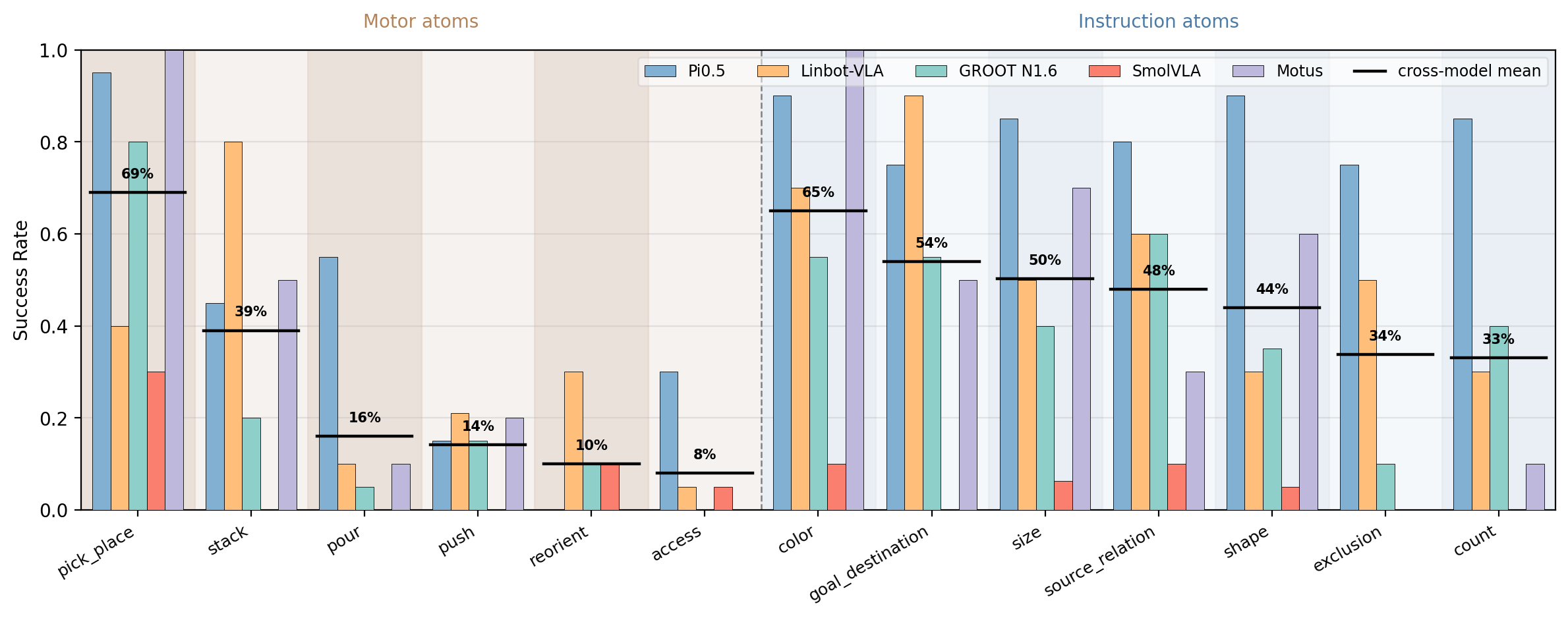}
\caption{
    Per-atom Success Rate under atomic-transfer fine-tuning.
    Each bar is one model's SR over all tasks instantiating that atom on both platforms.
}
\label{fig:rq1_atom_sr}
\end{figure*}

We first evaluate whether policies can acquire atomic skills under the atomic-transfer protocol and results are summarized in Table~\ref{tab:rq1_atomic_skill} and Figure~\ref{fig:rq1_atom_sr}.

\begin{table*}[t]
\centering
\small
\caption{
    \textbf{Atomic skill acquisition under atomic-transfer fine-tuning.}
    Mean SR and PSR (\%) over the Motor and Instruction Sets.
}
\label{tab:rq1_atomic_skill}
\begin{tabular}{l cccc cccc}
\toprule
\small
& \multicolumn{4}{c}{\textbf{Franka Panda}} & \multicolumn{4}{c}{\textbf{Cobot Magic}} \\
\cmidrule(lr){2-5} \cmidrule(lr){6-9}
& \multicolumn{2}{c}{Motor} & \multicolumn{2}{c}{Instruction} & \multicolumn{2}{c}{Motor} & \multicolumn{2}{c}{Instruction} \\
\cmidrule(lr){2-3} \cmidrule(lr){4-5} \cmidrule(lr){6-7} \cmidrule(lr){8-9}
\textbf{Model} & SR & PSR & SR & PSR & SR & PSR & SR & PSR \\
\midrule
Pi0.5 & \textbf{46.2} & \textbf{56.2} & \textbf{94.3} & \textbf{95.7} & \textbf{45.0} & \textbf{72.0} & \textbf{71.4} & \textbf{83.2} \\
Motus & 36.2 & 46.2 & 67.1 & 78.7 & 35.0 & 59.3 & 50.0 & 67.5 \\
LingBot-VLA & 37.5 & 46.5 & 54.3 & 60.5 & 26.2 & 42.8 & 21.4 & 47.1 \\
GROOT N1.6 & 28.8 & 47.1 & 57.1 & 69.5 & 23.8 & 41.0 & 27.1 & 52.1 \\
SmolVLA & 17.5 & 29.8 & 11.4 & 31.2 & 10.0 & 32.2 & 5.7 & 29.3 \\
\bottomrule
\end{tabular}
\end{table*}

Overall, instruction atoms are easier than motor atoms. For example, Pi0.5 reaches 94.3\% SR on Franka Panda instruction atoms and 71.4\% SR on Cobot Magic instruction atoms, while its motor SR is only 46.2\% and 45.0\%, respectively. This is expected because most instruction atoms are evaluated through simple carrier actions such as pick-and-place, while motor atoms require the policy to execute the physical skill itself. Among motor atoms, pick-and-place is clearly the easiest, with a cross-model mean SR of $74\%$. In contrast, access ($9\%$), reorientation ($11\%$), pushing ($16\%$), and pouring ($20\%$) have much lower cross-model mean SR, showing that precise control and object pose control remain difficult.

Instruction atoms also show different levels of difficulty. Simple visual grounding tasks, such as color, size, and source relation, achieve higher success rates, while count and exclusion are more challenging. This suggests that current policies can often use basic visual attributes to select objects, but still struggle with set-level constraints and logical filtering. Across models, Pi0.5 performs best on both platforms, and Motus is overall the second-strongest. LingBot-VLA is competitive on Franka Panda motor atoms ($37.5\%$ SR, slightly above Motus) but trails on Cobot Magic; GROOT N1.6 shows mid-pack performance on both platforms; SmolVLA trails in every setting. Finally, PSR is often higher than SR, especially for motor tasks, indicating that many failures are partial: policies can complete early stages but fail to satisfy the final success condition.

\subsection{Compositional Generalization}
\label{sec:rq2}

\begin{table}[t]
\centering
\resizebox{0.7\textwidth}{!}{\begin{tabular}{l cccc cccc}
\toprule
& \multicolumn{4}{c}{\textbf{Franka Panda}} & \multicolumn{4}{c}{\textbf{Cobot Magic}} \\
\cmidrule(lr){2-5} \cmidrule(lr){6-9}
\textbf{Model} & SR & PSR & AS & CFS & SR & PSR & AS & CFS \\
\midrule
Pi0.5 & 15.8 & 30.4 & 83.3 & 73.7 & 16.7 & 42.6 & 79.5 & 56.8 \\
Motus & 10.8 & 26.5 & 69.3 & 49.4 & 7.5 & 31.5 & 68.5 & 48.3 \\
LingBot-VLA & 3.3 & 12.1 & 60.3 & 54.5 & 0.0 & 24.4 & 47.2 & 29.3 \\
GROOT N1.6 & 3.3 & 11.6 & 66.3 & 61.2 & 1.7 & 13.7 & 48.2 & 38.8 \\
SmolVLA & 0.0 & 6.6 & 33.3 & 28.3 & 0.0 & 6.3 & 32.5 & 27.3 \\
\bottomrule
\end{tabular}
}
\caption{Atomic-to-compositional transfer. Mean SR, PSR, AS, CFS (\%) over the Composition Set, per platform. AS is the atomic-baseline ceiling; CFS is the share of compositional failure attributable to composition rather than to weak atoms.}
\label{tab:rq2_composition}
\end{table}

\begin{figure}[t]
\centering
\includegraphics[width=0.5\textwidth]{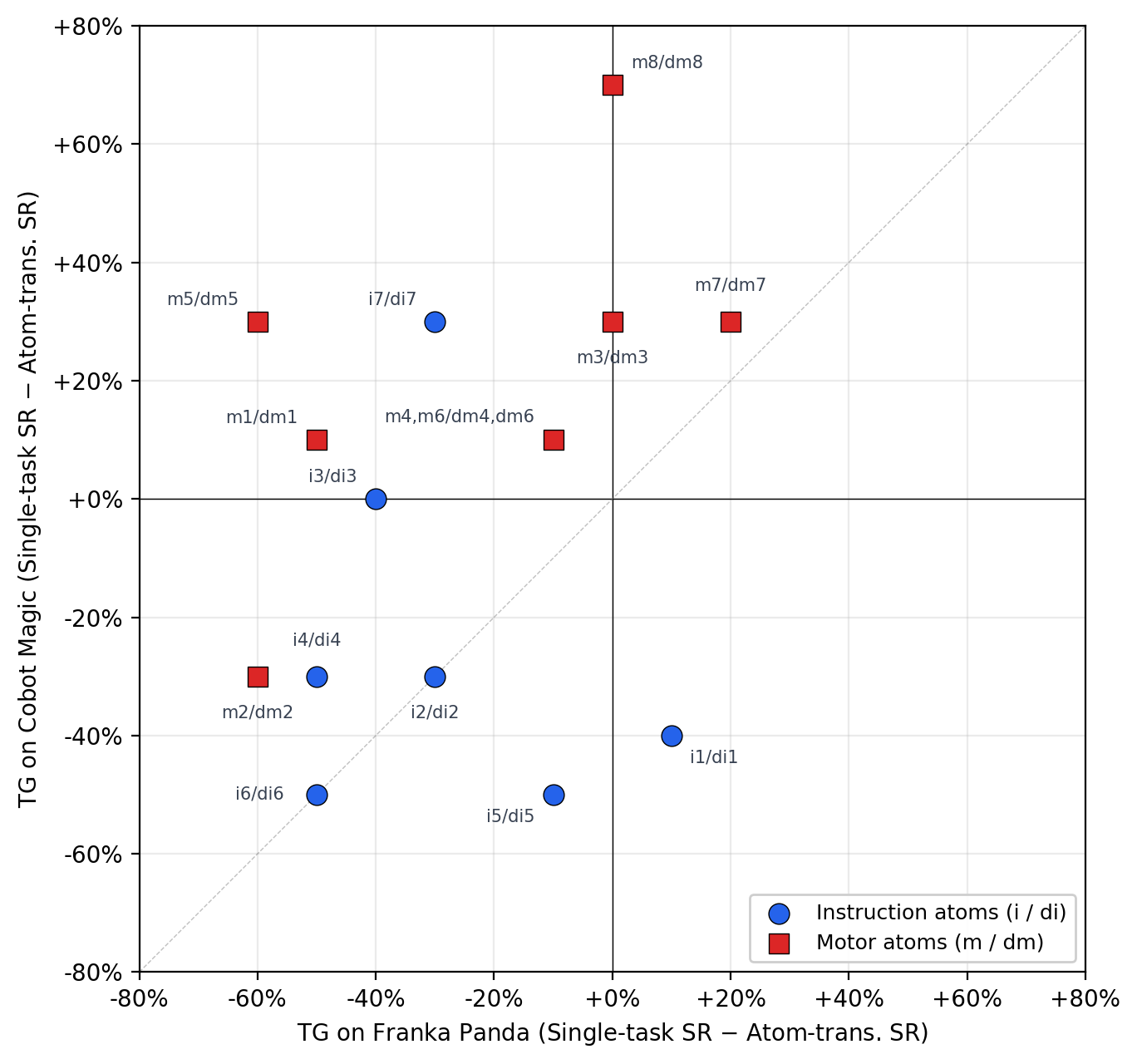}
\caption{Transfer Gap on paired atomic tasks of Pi0.5. Each point pairs a Franka atom with its dual-arm counterpart.}
\label{fig:rq4_transfer_gap}
\end{figure}

We next evaluate whether learned atomic skills transfer to held-out compositional tasks under the atomic-transfer protocol.
As shown in Table~\ref{tab:rq2_composition}, although most models achieve non-trivial AS, their compositional success rates remain low. Pi0.5 reaches high AS on both Franka Panda and Cobot Magic ($83.3\%$ and $79.5\%$), yet its compositional SR is only $15.8\%$ and $16.7\%$. Motus shows the same pattern. LingBot-VLA and GROOT N1.6 on Franka Panda also have moderate AS ($60.3\%$ and $66.3\%$) that does not translate into compositional success (both $3.3\%$ SR). The gap between AS and compositional SR is consistent across models, showing that learning atomic skills does not automatically yield the ability to recombine them into unseen tasks.

CFS further shows that this gap is not only caused by weak atomic skills. Pi0.5 has the highest PSR on both tracks and the highest CFS as well, indicating that most of its remaining failure comes from composition itself rather than from unlearned atoms. Motus shows a similar profile with CFS around $48$--$49\%$ on both platforms despite a lower atomic ceiling. In contrast, SmolVLA has both low AS and CFS, suggesting that its failures are still largely atom-limited. Across models, PSR is consistently higher than SR, showing that policies often complete some intermediate steps but fail to finish the full compositional task. Overall, these results suggest that current policies can acquire some useful atomic skills, but still struggle to reliably compose them in real-world settings.

\subsection{Failure Analysis}
\label{sec:rq3}

\begin{figure*}[t]
\centering
    \centering
    \includegraphics[width=\linewidth]{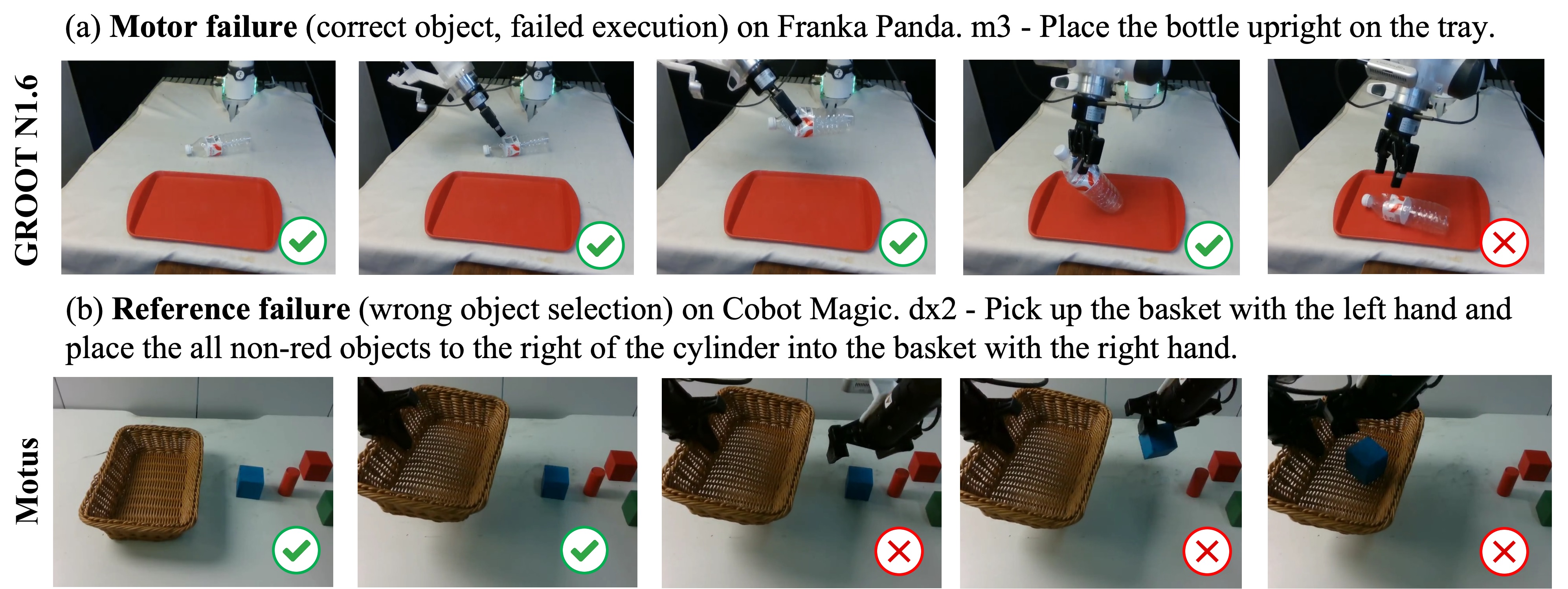}
    \caption{Representative failure cases.}
\label{fig:failure_examples}
\end{figure*}

\begin{figure*}[t]
\centering
\includegraphics[width=\linewidth]{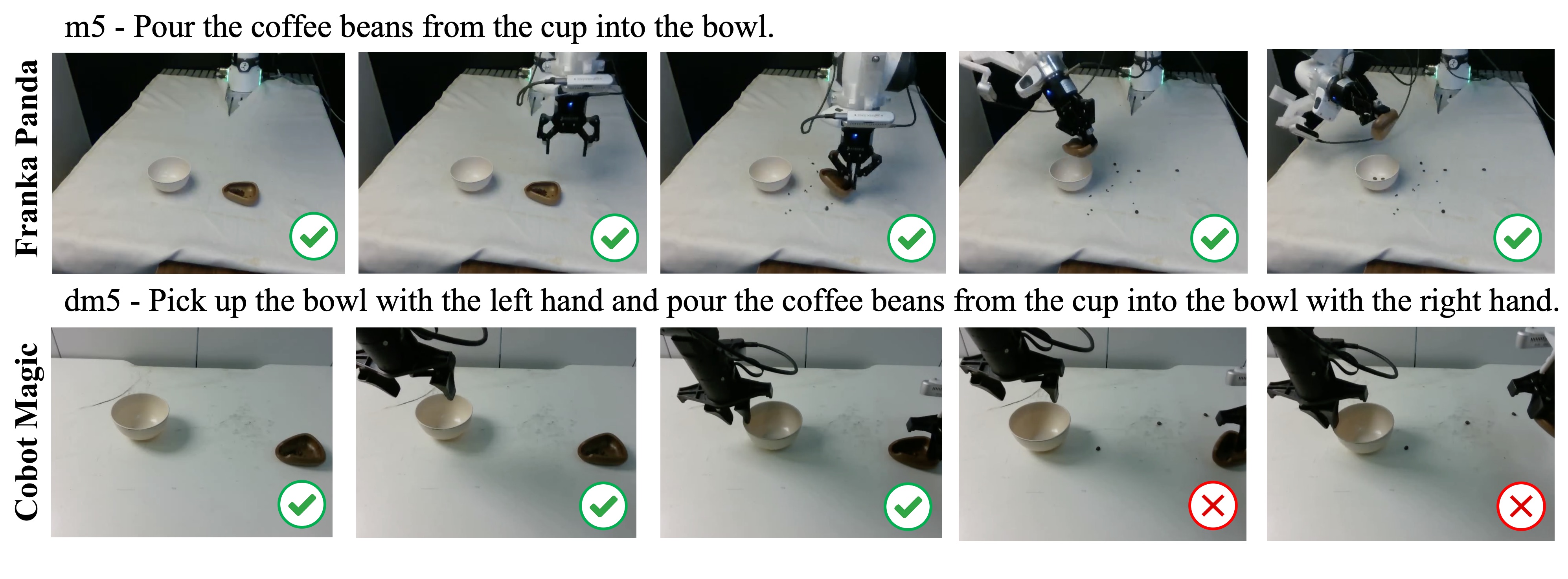}
\caption{Single-arm vs.\ dual-arm failure comparison on a paired task.
         The single-arm rollout completes the task, whereas the dual-arm
         counterpart fails midway.  }%
\label{fig:single_dual_failure}
\end{figure*}

We further analyze failed rollouts to understand where current policies break down. We annotate each failed rollout into five types: \emph{motor failure} (correct object but failed execution), \emph{reference failure} (wrong object selection), \emph{constraint failure} (violating count, exclusion, or other logical constraints), \emph{safety abort} (human intervention due to unsafe behavior), and \emph{timeout} (failure to finish within the time limit).

Across all tasks, motor failure, reference failure, and timeout account for over 90\% of failures. This suggests that most failures come from three bottlenecks: physical execution, target grounding, and task progress. Figure~\ref{fig:failure_examples} shows representative examples of motor failure and reference failure. 
Motor failures are common in tasks that require precise control, such as pouring, reorientation and articulated-object access.
Reference failures are more frequent in tasks involving spatial relations, exclusion, or multiple candidate destinations.

We also compare paired single-arm and dual-arm tasks.
The dual-arm tasks are designed as bimanual counterparts of the single-arm tasks, but they introduce extra requirements such as hand-role assignment, inter-arm coordination, and longer ordered action sequences, which make them more challenging.
As shown in Figure~\ref{fig:single_dual_failure}, the Franka Panda single-arm
rollout completes the pour-beans task, whereas the Cobot Magic dual-arm rollout
fails to grasp the bowl with the left hand, leaving the pour-and-receive
sequence unexecutable.

\subsection{Atomic-transfer vs. Single-task Atomic Fine-tuning}
\label{sec:rq4}

We compare atomic-transfer fine-tuning with single-task atomic fine-tuning on Pi0.5.
For each atomic task, we compute the Transfer Gap (TG) between the single-task checkpoint and the atomic-transfer checkpoint; negative TG indicates that atomic-transfer performs better.
Since the atomic-transfer checkpoint is trained on all 15 atomic tasks, this comparison should be read as a comparison against a 100-demonstration single-task baseline rather than a general verdict on single-task training.

Figure~\ref{fig:rq4_transfer_gap} shows that the effect depends on both platform and atom type.
On Franka Panda, most atoms have negative TG, suggesting that joint atomic training improves many single-arm skills.
This is likely because these tasks share low-level behaviors such as object localization, grasping, moving, and placing, while multi-task training provides more data and scene variation than fitting each task separately.

The pattern is more mixed on Cobot Magic.
Instruction atoms still mostly benefit from atomic-transfer training, likely because they share visual-language grounding and target selection.
In contrast, several bimanual motor atoms have positive TG.
These tasks require hand-role assignment, inter-arm coordination, and precise ordered actions, so single-task fine-tuning can better specialize to the required action pattern.
Overall, atomic-transfer fine-tuning is a strong default for shared instruction grounding and single-arm manipulation, while complex bimanual motor skills remain more sensitive to task-specific practice.

\section{Conclusion}
We introduced ATOM-Bench, a real-world benchmark for evaluating atomic skills and compositional generalization in manipulation policies.
ATOM-Bench contains 30 atomic tasks and 24 held-out compositional tasks across paired single-arm and dual-arm tracks, enabling diagnosis of motor execution, instruction grounding, and compositional reuse. Through 2,700 physical rollouts on five representative policies, we find that current policies can acquire some useful atomic skills, especially simple instruction-grounding skills.
However, fine-grained motor atoms and instruction atoms that require selecting objects from a set, such as counting and exclusion remain difficult, and strong atomic performance does not reliably transfer to held-out compositional tasks.
We hope ATOM-Bench helps the community better understand where manipulation policies fail and supports progress toward more reliable real-world generalization.

\section{Limitations}
ATOM-Bench is limited by the cost of real-robot evaluation and model fine-tuning.
We evaluate five representative policies, but do not cover all available VLA and WAM models.
Our task suite focuses on common tabletop manipulation factors that can be reliably instantiated on real robots, and therefore does not cover many long-tail skills such as deformable-object manipulation, tool use, mobile manipulation, or highly long-horizon tasks.
In addition, evaluation remains labor-intensive: each rollout requires scene setup, safety monitoring, and manual score-point annotation.
Future work will extend the task coverage and explore more automated reset, success detection, and annotation tools to scale real-world diagnostic evaluation.

\bibliography{main}  %
\bibliographystyle{flageval_baai}

\clearpage
\appendix
\section{Task Design}
\label{app:task_design}

ATOM-Bench instantiates the motor and instruction atoms as concrete real-robot tasks on two platforms.
Each platform has three task groups: a \emph{Motor Set} that isolates a single motor atom under a minimal, unambiguous instruction; an \emph{Instruction Set} that isolates a single instruction atom under a simple pick-and-place carrier motor; and a held-out \emph{Composition Set} that combines one or more motor atoms with two or more instruction atoms.
\textbf{The two platforms share a paired task design.}
The single-arm (Franka Panda) tasks are prefixed \texttt{m}/\texttt{i}/\texttt{x} for the three groups, and the dual-arm (Cobot Magic) tasks are prefixed \texttt{dm}/\texttt{di}/\texttt{dx}.
The two tracks are constructed in one-to-one correspondence, the biggest difference is that the dual-arm version additionally requires bimanual coordination. The full task list, with prompts and the atoms each task instantiates, is given in Tables~\ref{tab:tasks-m}--\ref{tab:tasks-dx}.

\begin{longtable}{
  >{\centering\arraybackslash}m{1.4cm}
  >{\raggedright\arraybackslash}m{6.0cm}
  >{\centering\arraybackslash}m{2.2cm}
  >{\centering\arraybackslash}m{4.5cm}
}
\caption{Single-arm Motor Set (\texttt{m1}--\texttt{m8}). Each task isolates one motor atom under a minimal, unambiguous instruction.}\label{tab:tasks-m}\\
\toprule
Task\_id & Prompt & Instruction\_atom & Image \\
\midrule
\endfirsthead

\toprule
Task\_id & Prompt & Instruction\_atom & Image \\
\midrule
\endhead

\midrule
\multicolumn{4}{r}{\small Continued on next page} \\
\endfoot

\bottomrule
\endlastfoot

m1 & Pick up the cube and place it into the basket. & pick\_place &
\includegraphics[width=3.0cm, height=1.7cm, keepaspectratio]{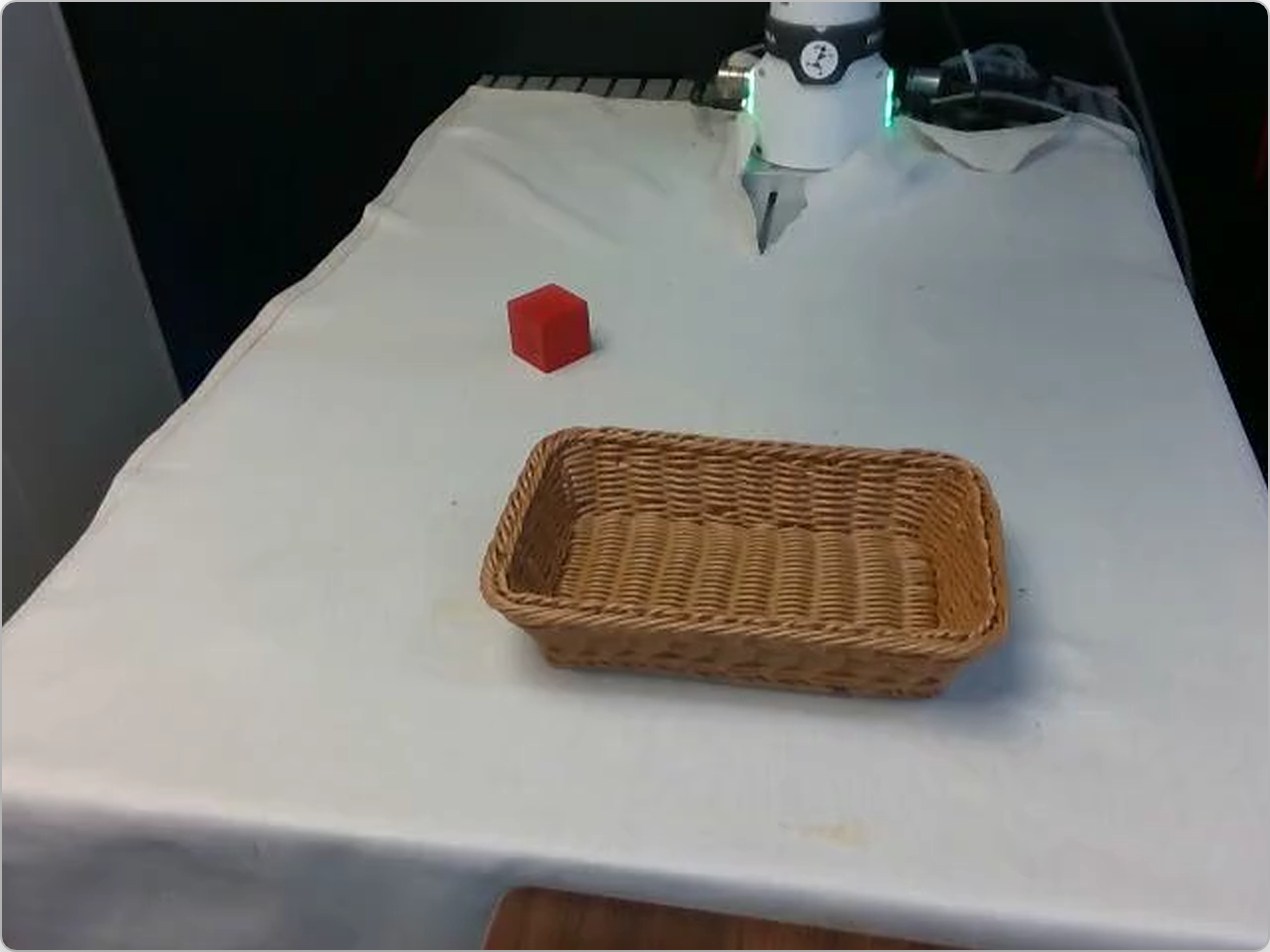} \\

m2 & Pick up the ball and place it into the basket. & pick\_place &
\includegraphics[width=3.0cm, height=1.7cm, keepaspectratio]{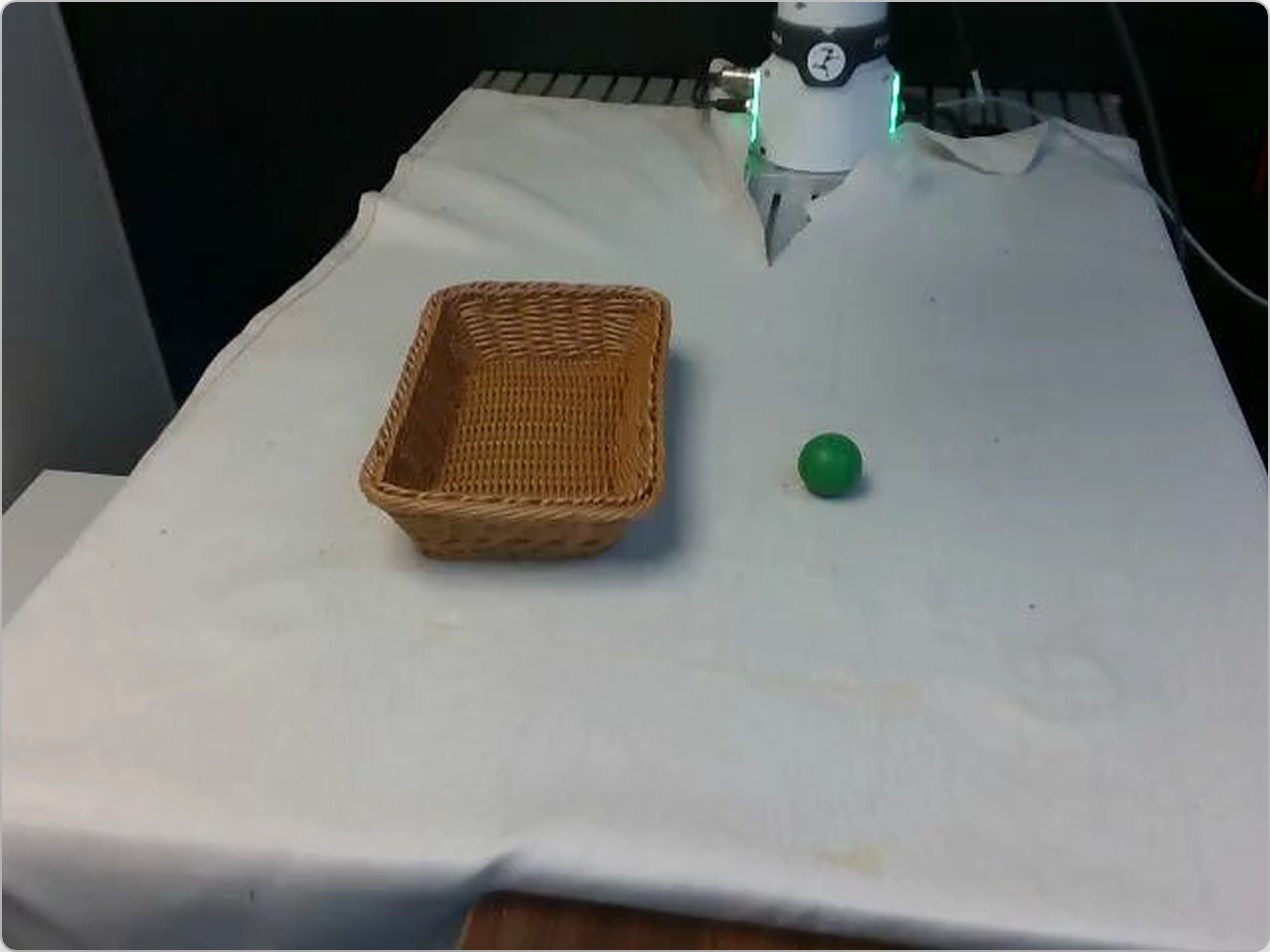} \\

m3 & Place the bottle upright on the tray. & reorient &
\includegraphics[width=3.0cm, height=1.7cm, keepaspectratio]{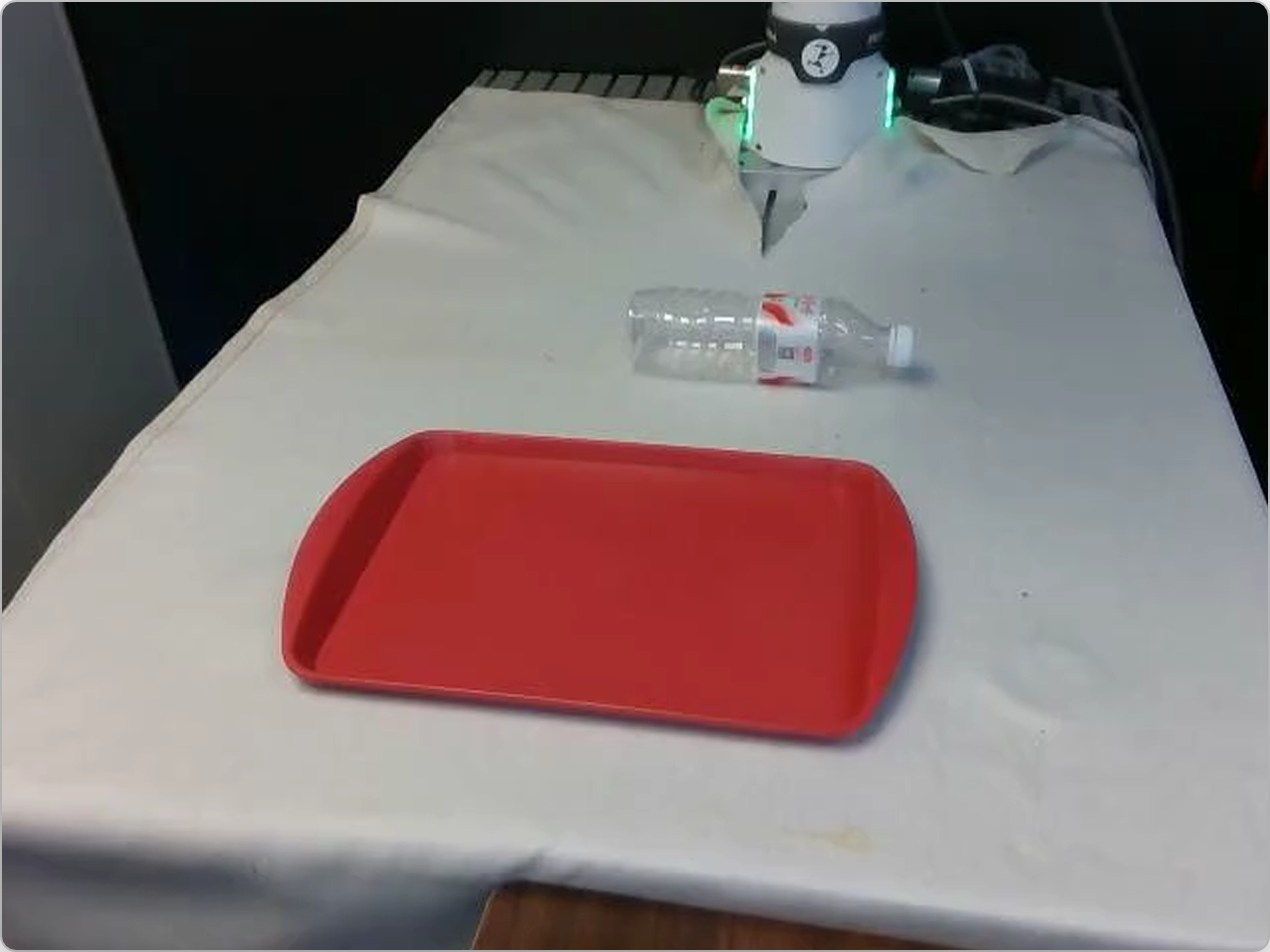} \\

m4 & Push the cube into the marked area. & push &
\includegraphics[width=3.0cm, height=1.7cm, keepaspectratio]{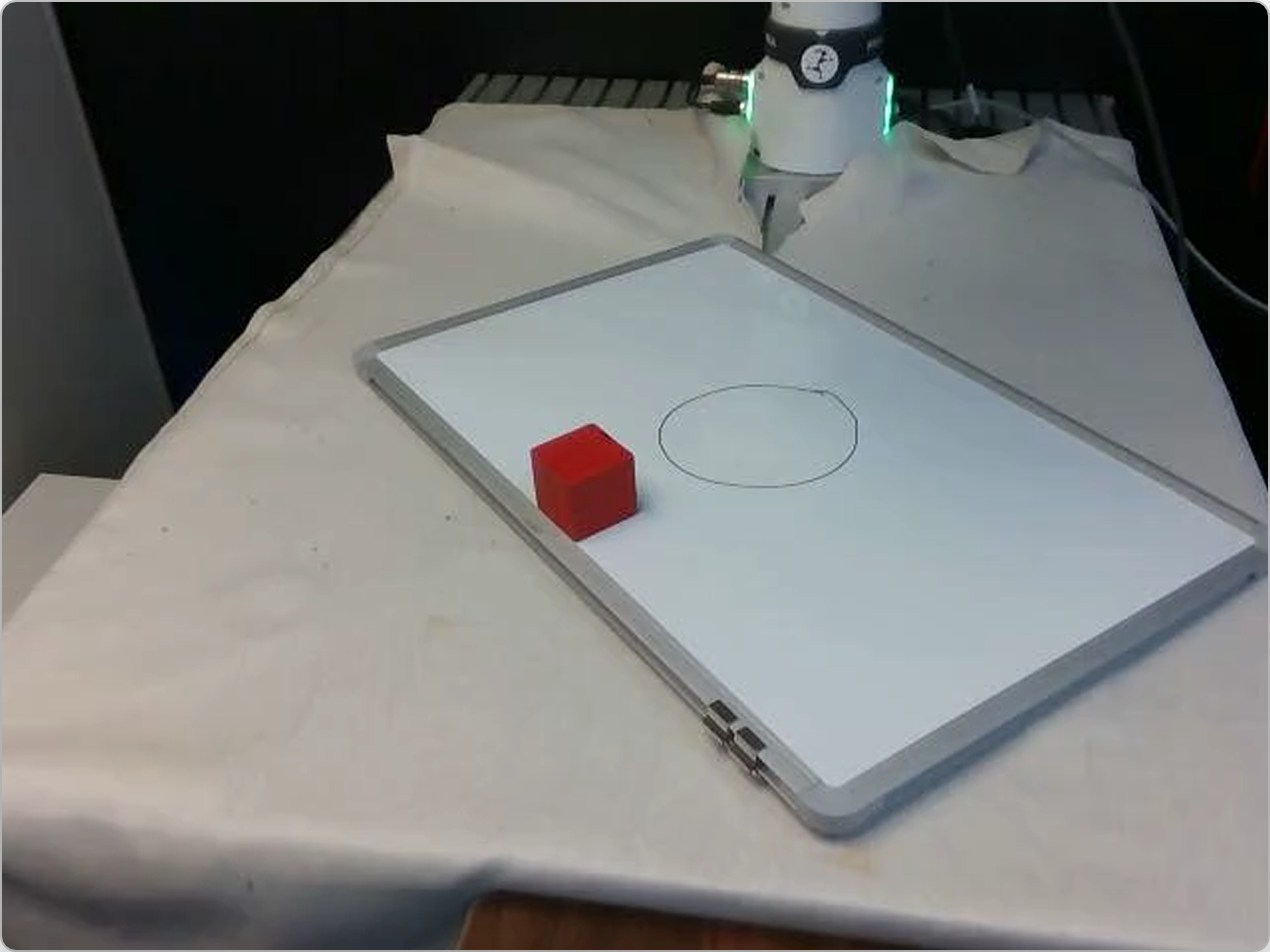} \\

m5 & Pour the coffee beans from the cup into the bowl. & pour &
\includegraphics[width=3.0cm, height=1.7cm, keepaspectratio]{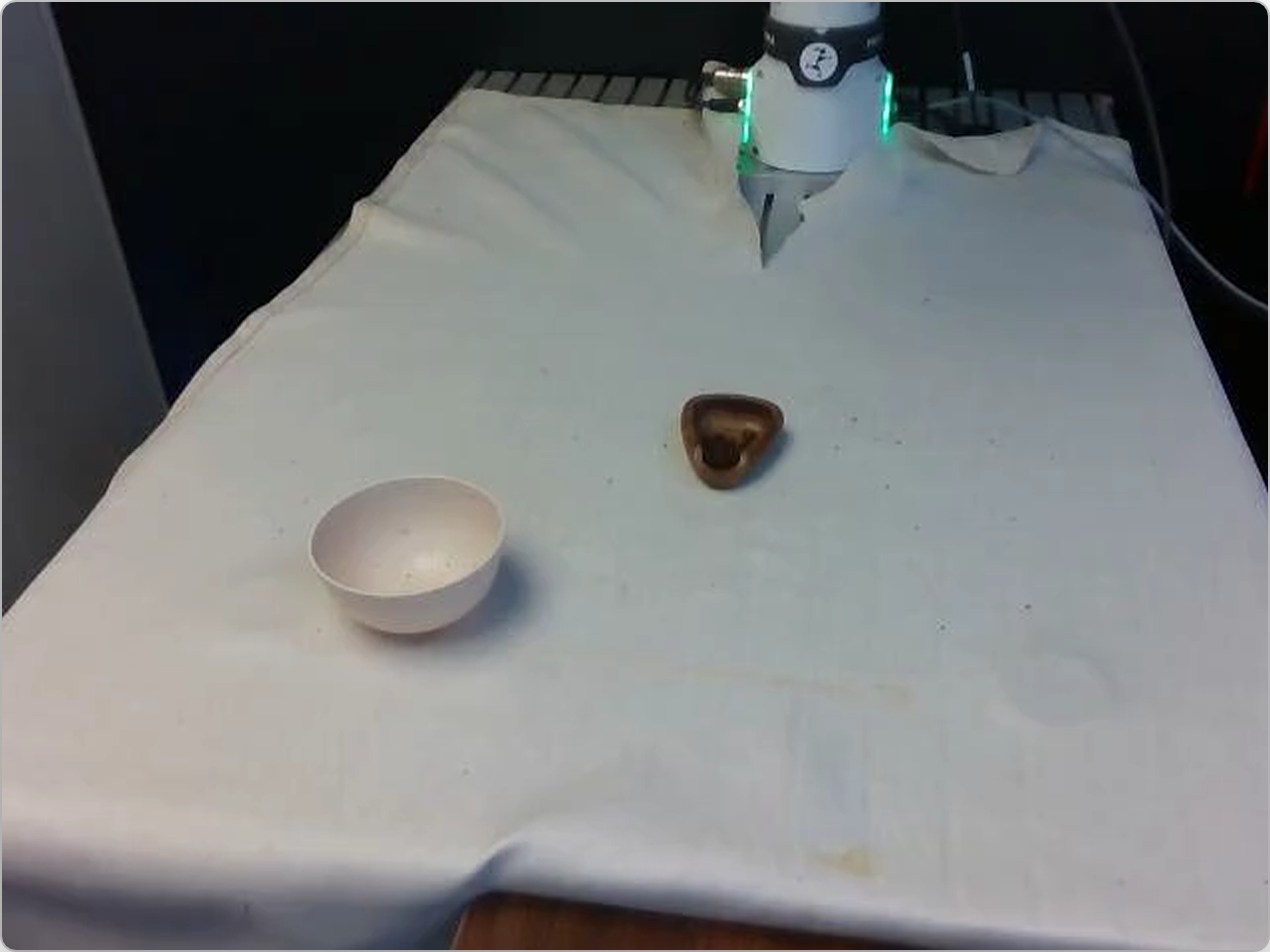} \\

m6 & Open the drawer, then close it. & access &
\includegraphics[width=3.0cm, height=1.7cm, keepaspectratio]{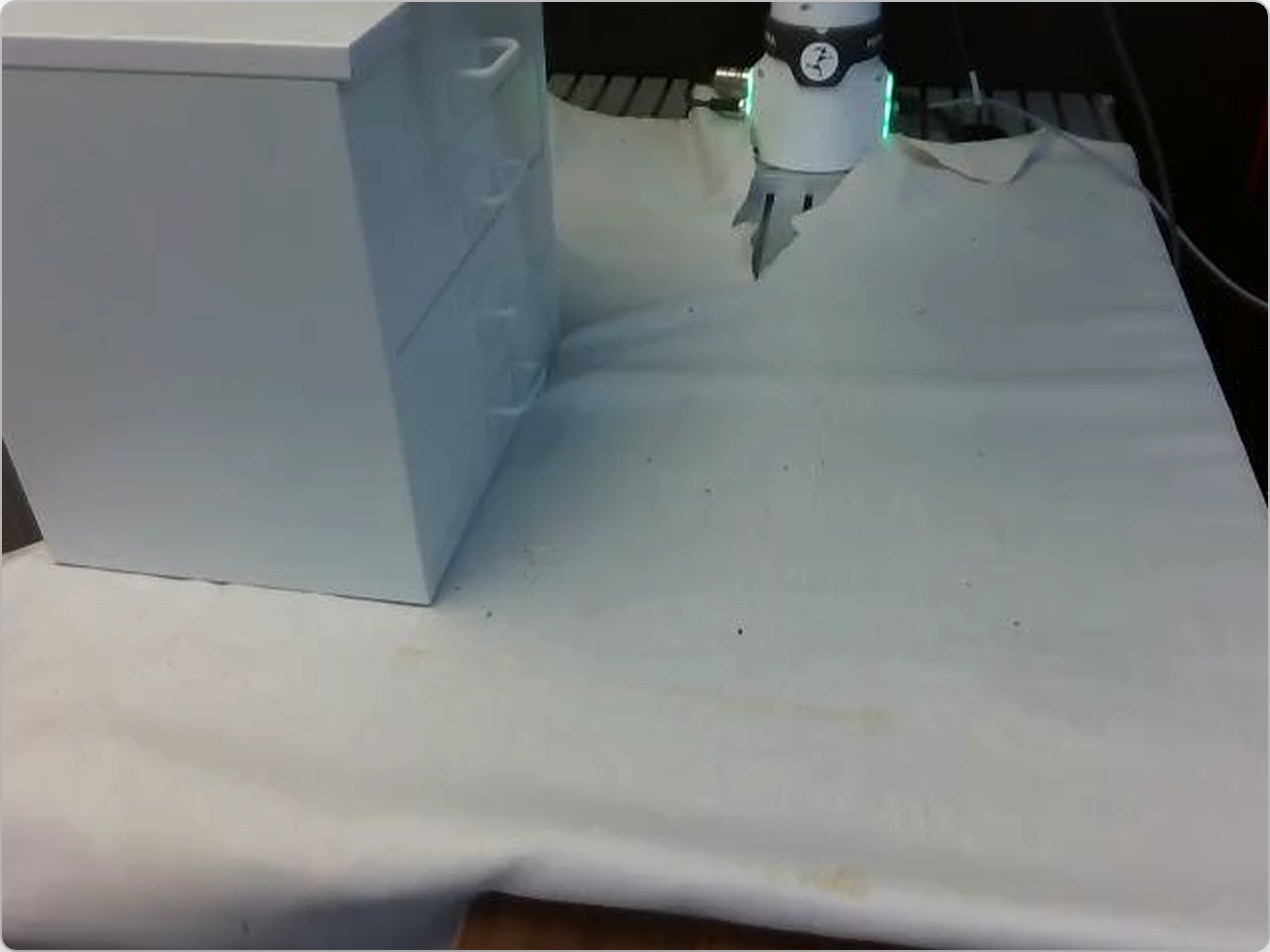} \\

m7 & Stack the two cubes. & stack &
\includegraphics[width=3.0cm, height=1.7cm, keepaspectratio]{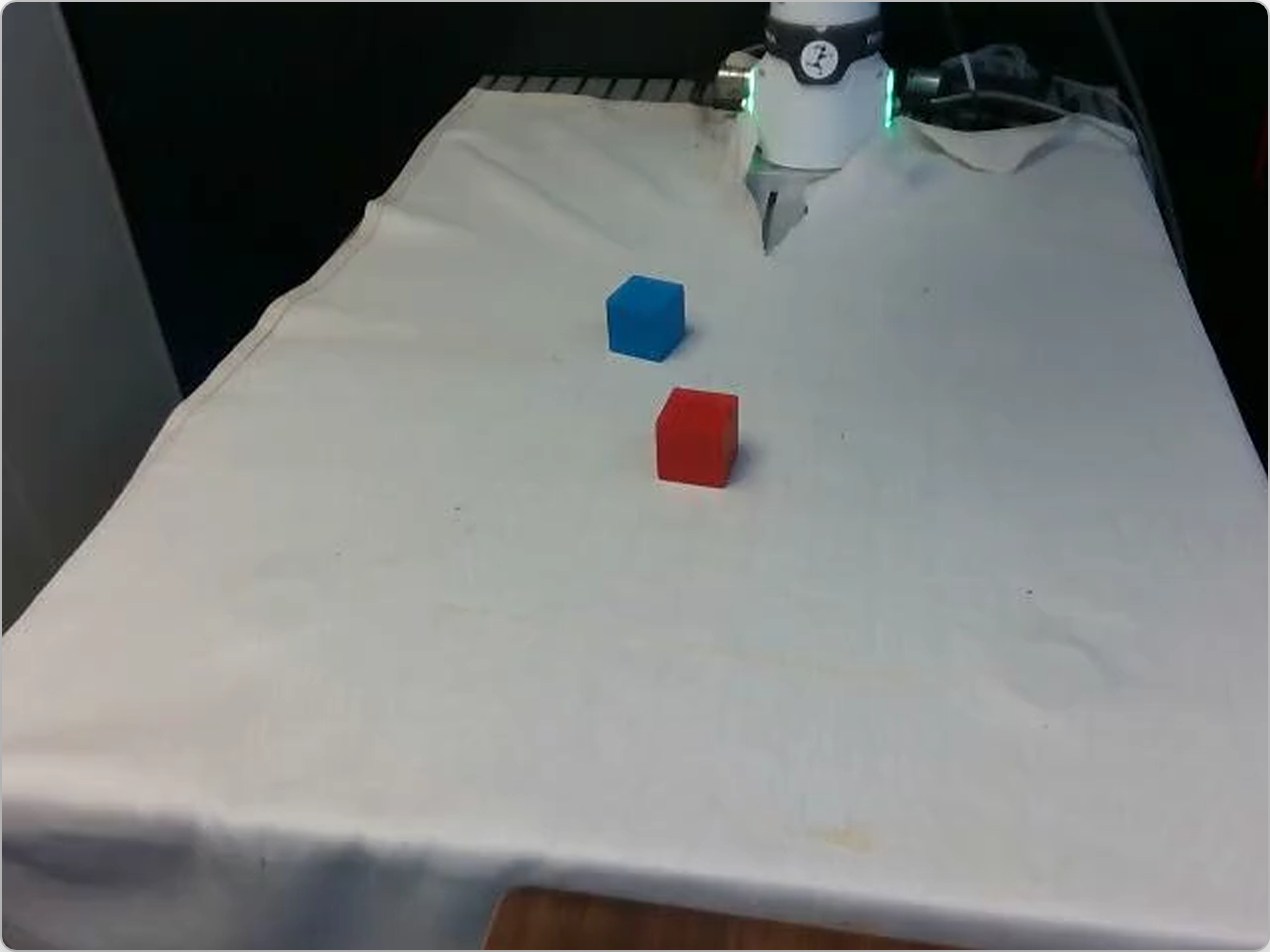} \\

m8 & Open the lid, then close it. & access &
\includegraphics[width=3.0cm, height=1.7cm, keepaspectratio]{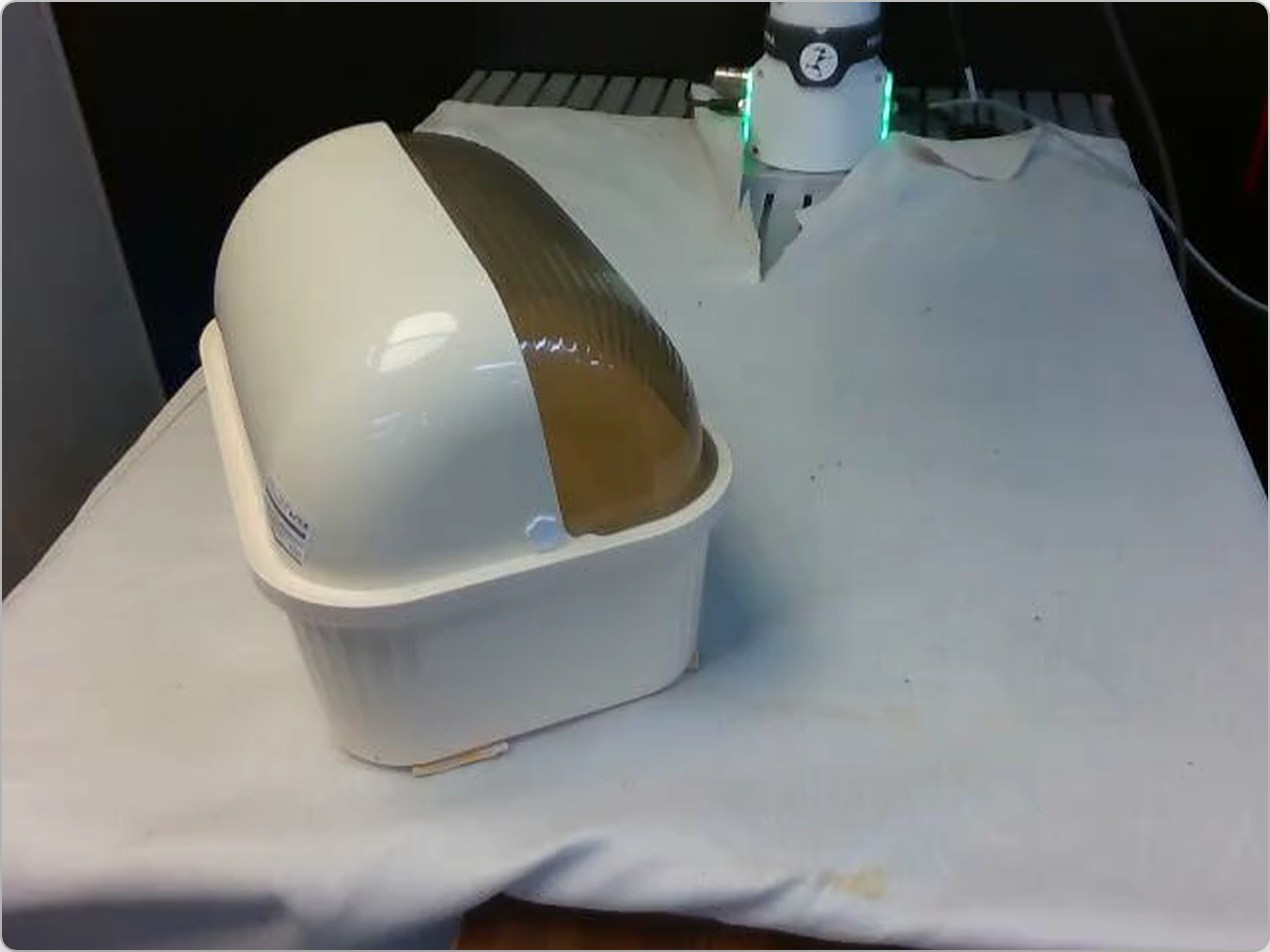} \\

\end{longtable}

\begin{longtable}{
  >{\centering\arraybackslash}m{1.4cm}
  >{\raggedright\arraybackslash}m{6.0cm}
  >{\centering\arraybackslash}m{2.2cm}
  >{\centering\arraybackslash}m{4.5cm}
}
\caption{Single-arm Instruction Set (\texttt{i1}--\texttt{i7}). Each task isolates one instruction atom under a pick-and-place carrier motor.}\label{tab:tasks-i}\\
\toprule
Task\_id & Prompt & Instruction\_atom & Image \\
\midrule
\endfirsthead

\toprule
Task\_id & Prompt & Instruction\_atom & Image \\
\midrule
\endhead

\midrule
\multicolumn{4}{r}{\small Continued on next page} \\
\endfoot

\bottomrule
\endlastfoot

i1 & Pick up the blue cube and place it into the basket. & color &
\includegraphics[width=3.0cm, height=1.7cm, keepaspectratio]{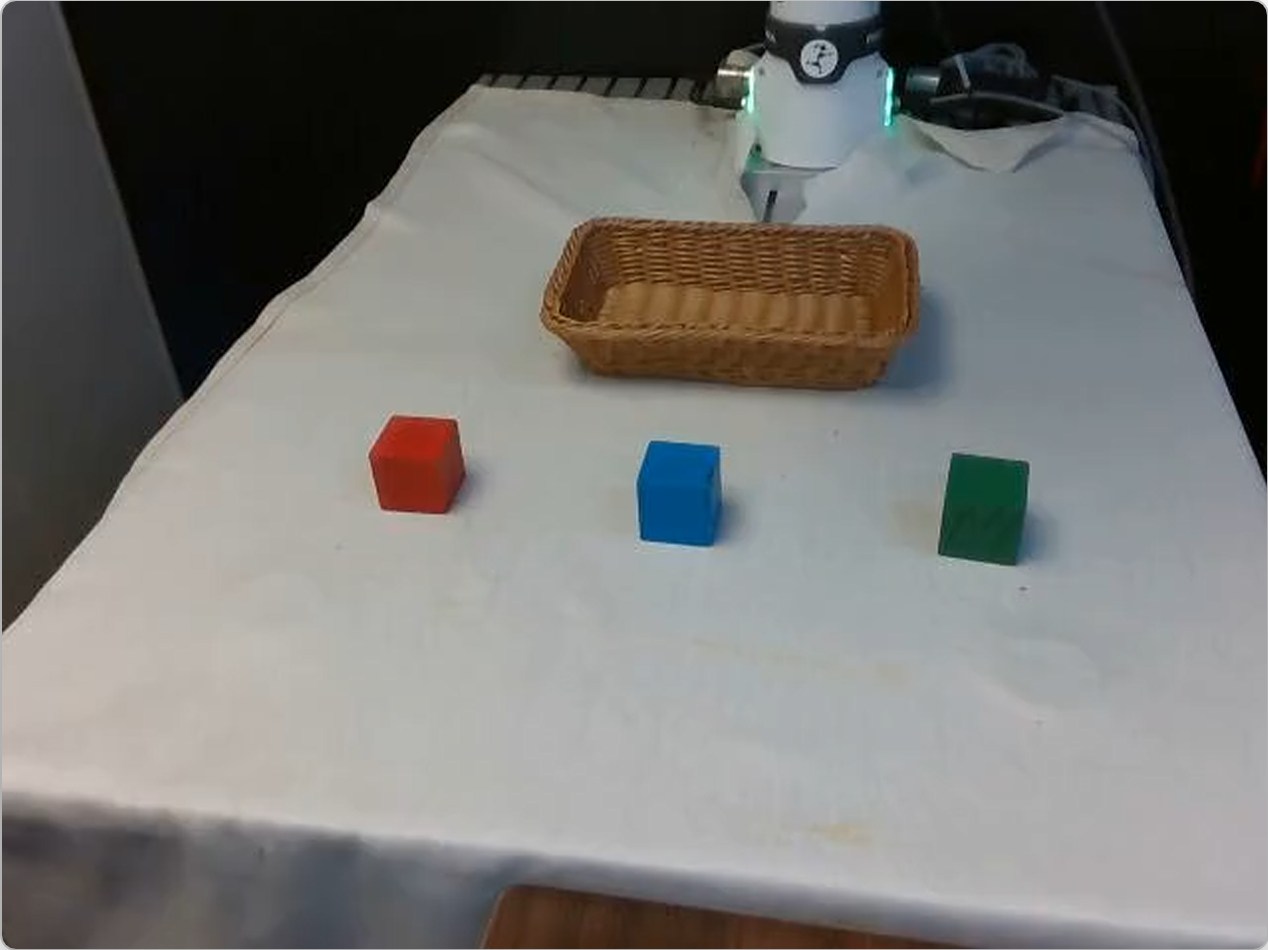} \\

i2 & Pick up the triangular block and place it into the basket. & shape &
\includegraphics[width=3.0cm, height=1.7cm, keepaspectratio]{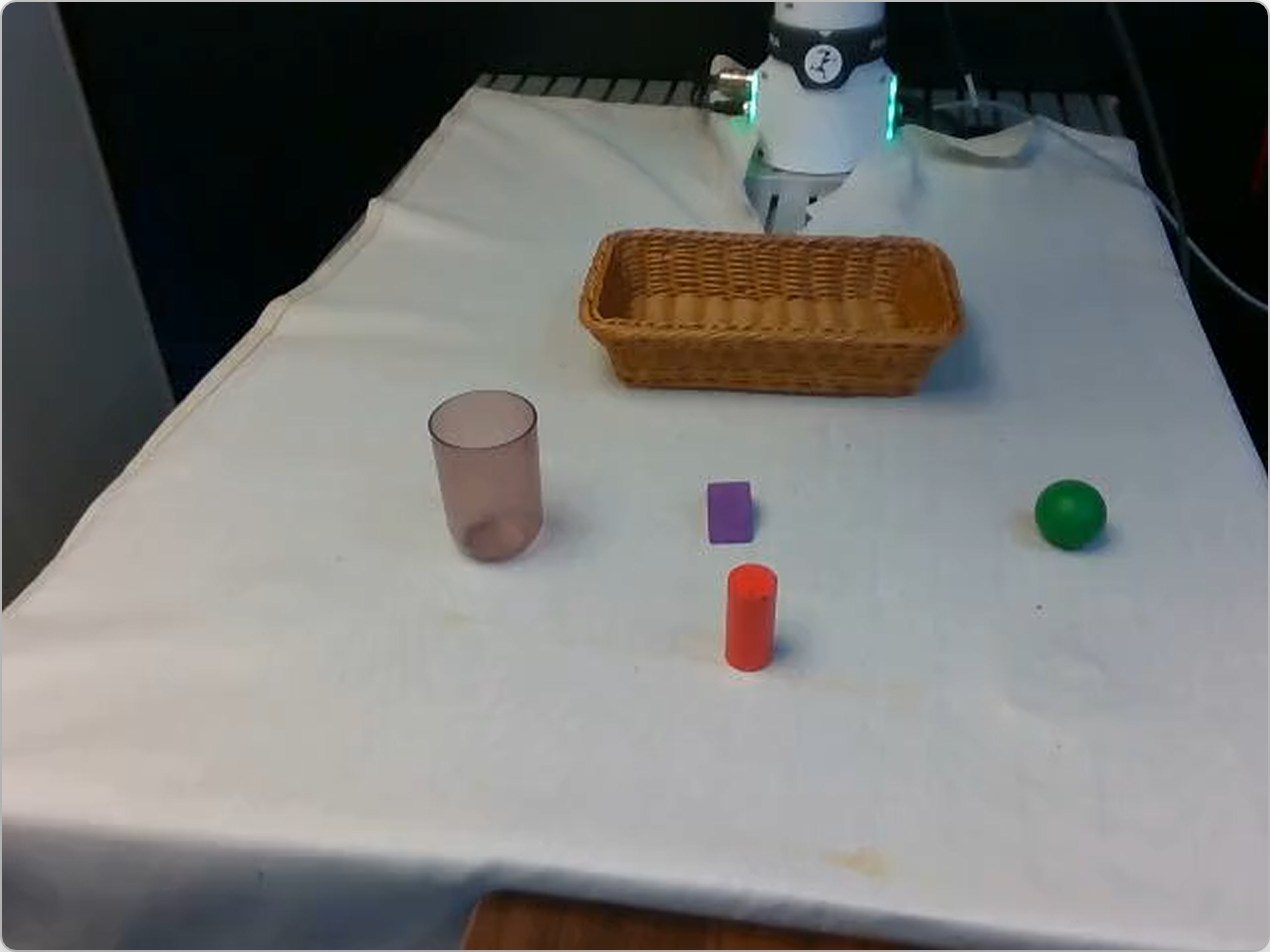} \\

i3 & Pick up the largest block and place it onto the tray. & size &
\includegraphics[width=3.0cm, height=1.7cm, keepaspectratio]{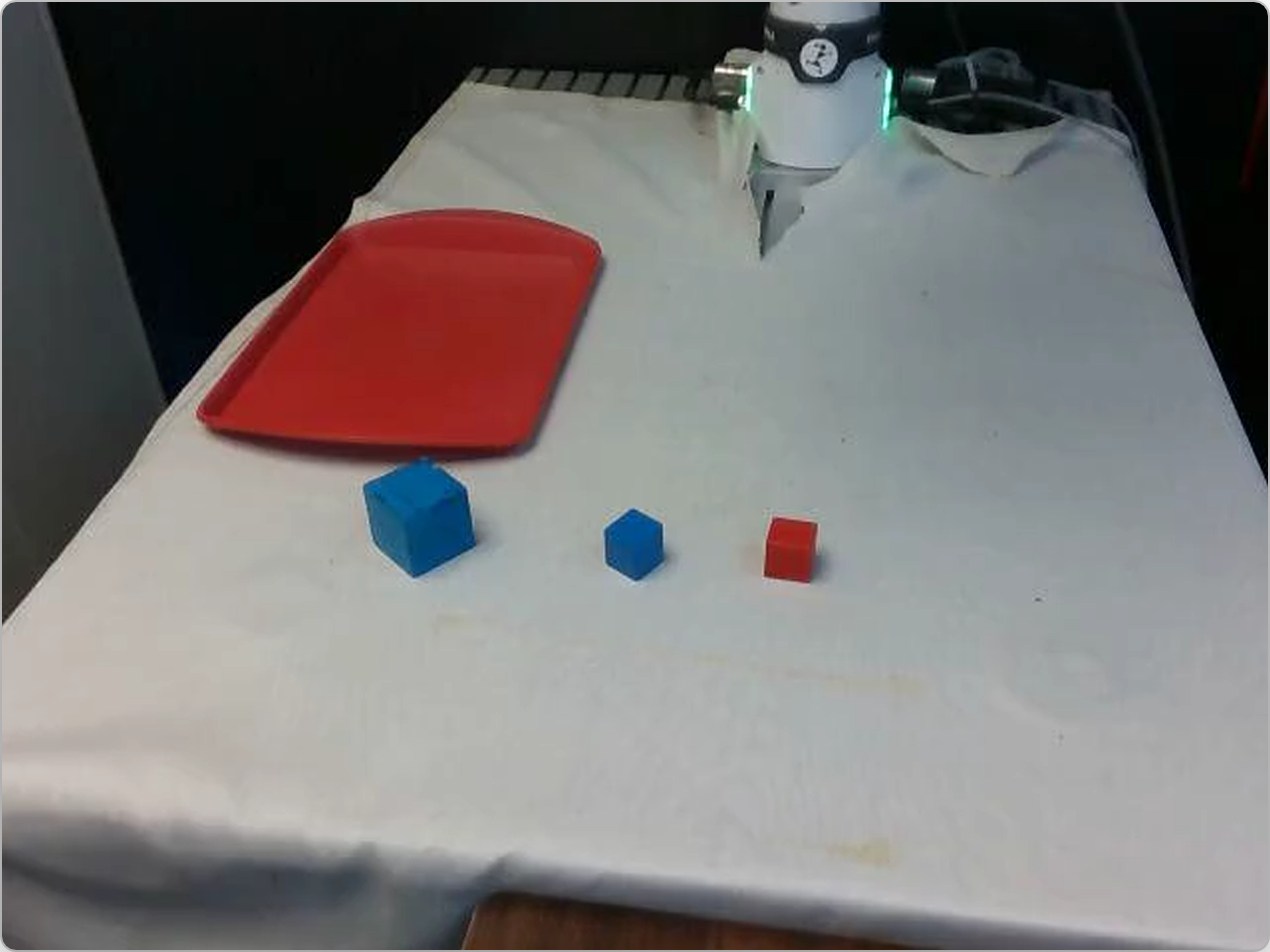} \\

i4 & Pick up the object to the left of the cylinder and place it into the basket. & source\_relation &
\includegraphics[width=3.0cm, height=1.7cm, keepaspectratio]{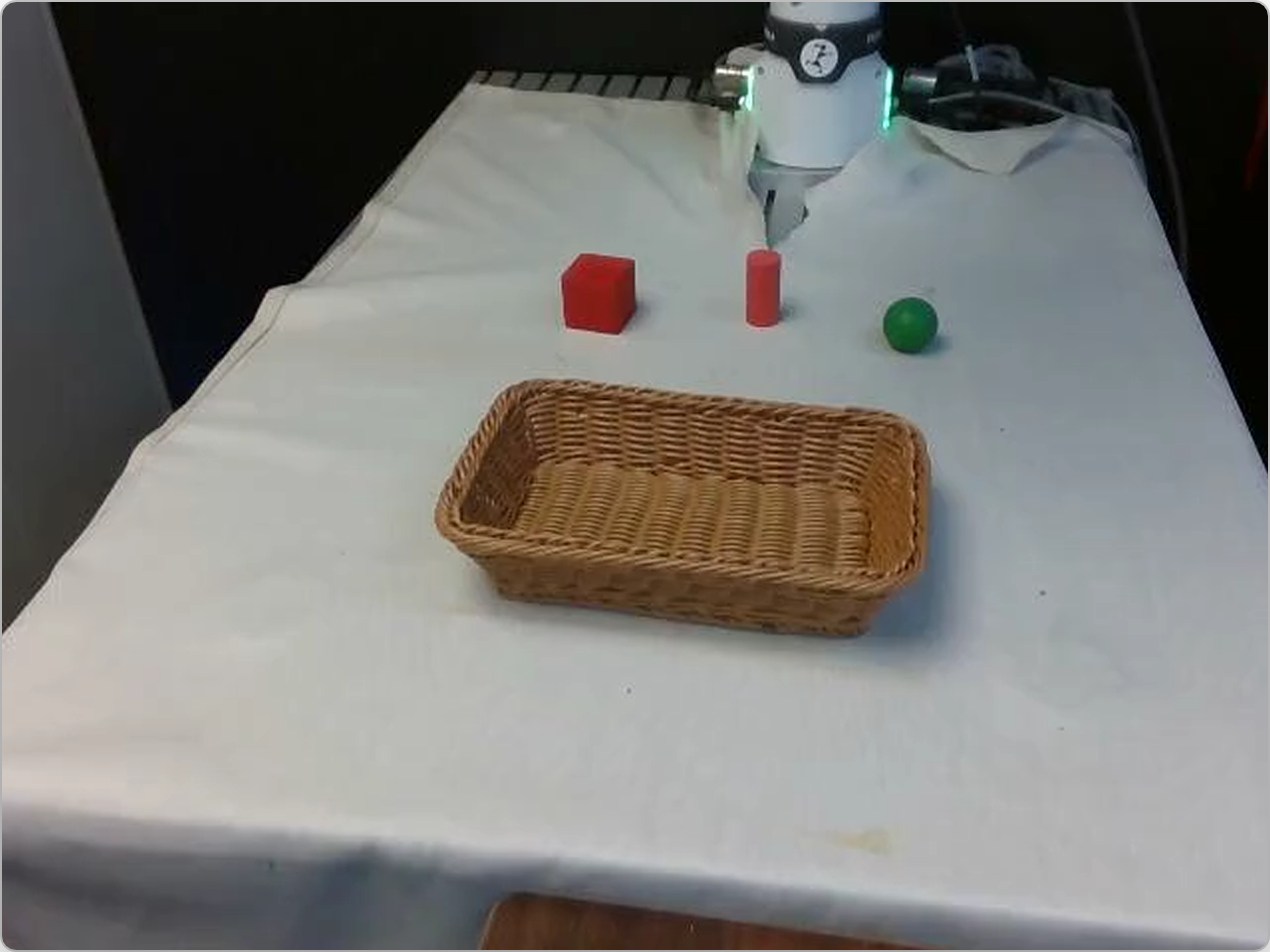} \\

i5 & Pick up exactly two blocks and place them into the basket. & count &
\includegraphics[width=3.0cm, height=1.7cm, keepaspectratio]{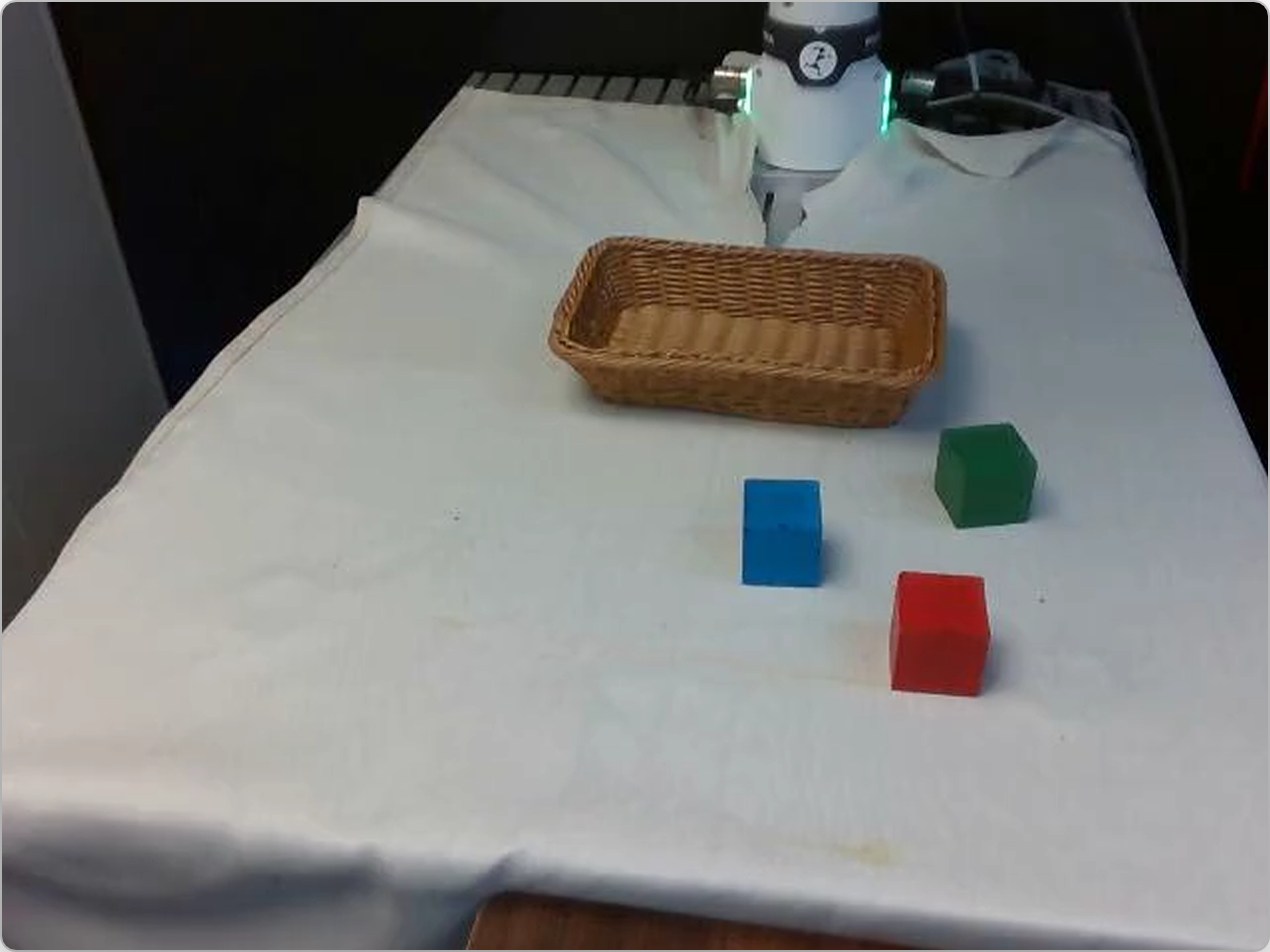} \\

i6 & Pick up the red cube and place it into the right basket. & goal\_destination &
\includegraphics[width=3.0cm, height=1.7cm, keepaspectratio]{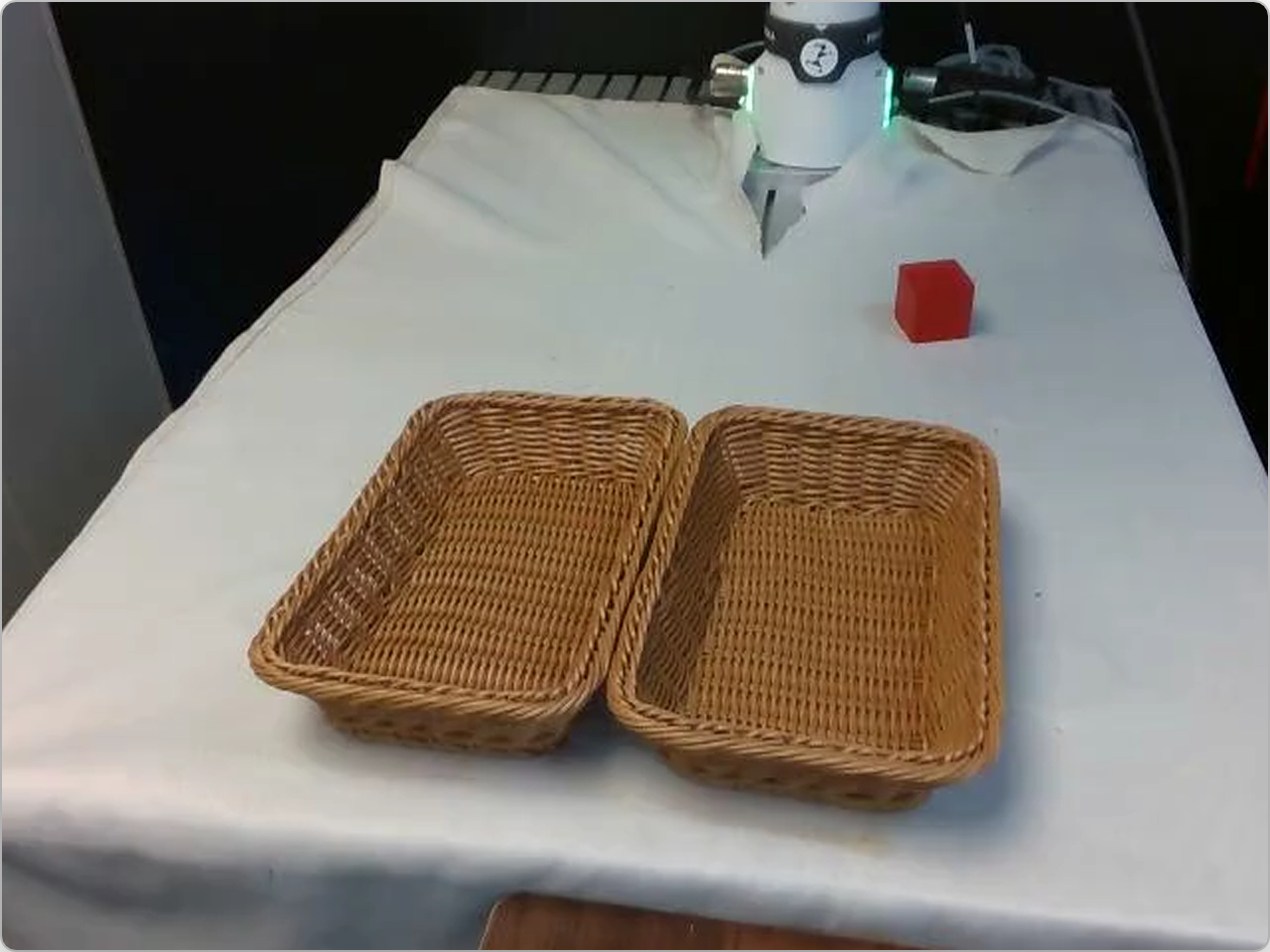} \\

i7 & Place all non-red objects into the basket. & exclusion &
\includegraphics[width=3.0cm, height=1.7cm, keepaspectratio]{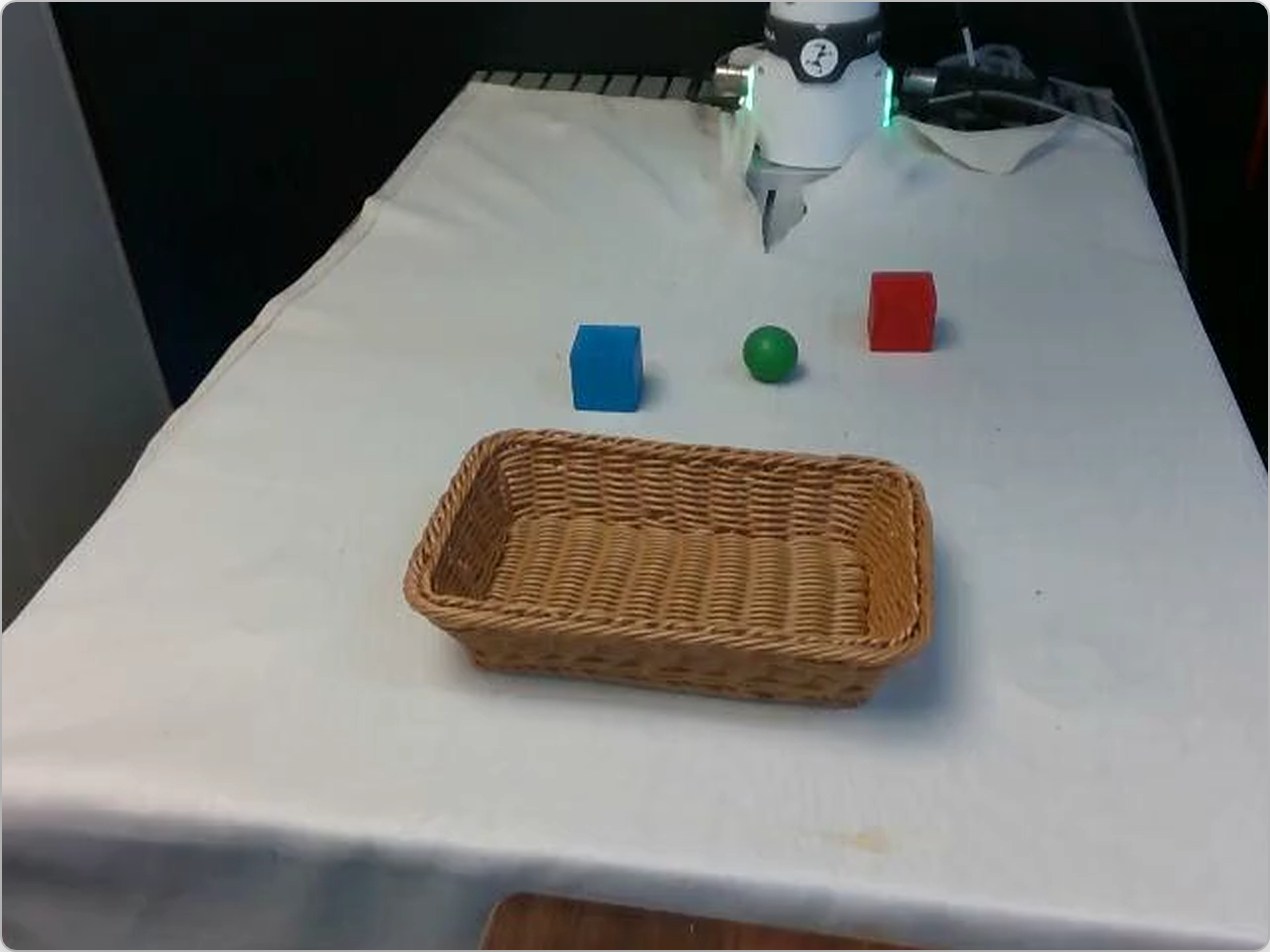} \\

\end{longtable}

\begin{longtable}{
  >{\centering\arraybackslash}m{1.4cm}    %
  >{\raggedright\arraybackslash}m{5.0cm}  %
  >{\centering\arraybackslash}m{1.6cm}    %
  >{\centering\arraybackslash}m{1.6cm}    %
  >{\centering\arraybackslash}m{4.5cm}    %
}
\caption{Single-arm Composition Set (\texttt{x1}--\texttt{x12}). Each held-out task combines one or more motor atoms with two or more instruction atoms.}\label{tab:tasks-x}\\
\toprule
Task\_id & Prompt & Motor\_atom & Instruction\_atoms & Image \\
\midrule
\endfirsthead

\toprule
Task\_id & Prompt & Motor\_atom & Instruction\_atoms & Image \\
\midrule
\endhead

\midrule
\multicolumn{5}{r}{\small Continued on next page} \\
\endfoot

\bottomrule
\endlastfoot

x1 & Pick up exactly two red cubes and place them into the basket. & pick\_place & color + count &
\includegraphics[width=3.0cm, height=1.7cm, keepaspectratio]{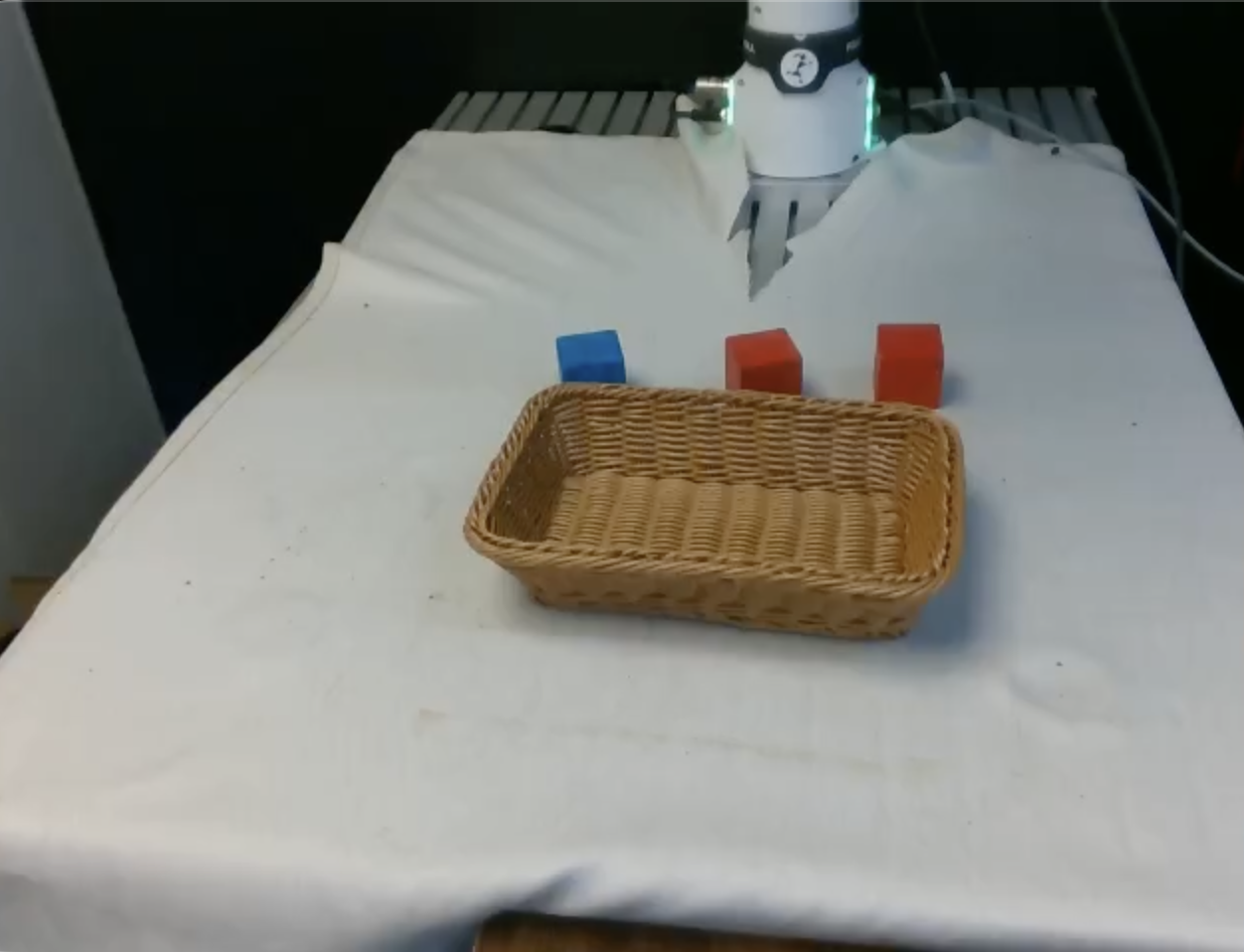} \\  

x2 & Pick up the non-red object to the right of the cylinder and place it into the basket. & pick\_place & source\_relation + exclusion &
\includegraphics[width=3.0cm, height=1.7cm, keepaspectratio]{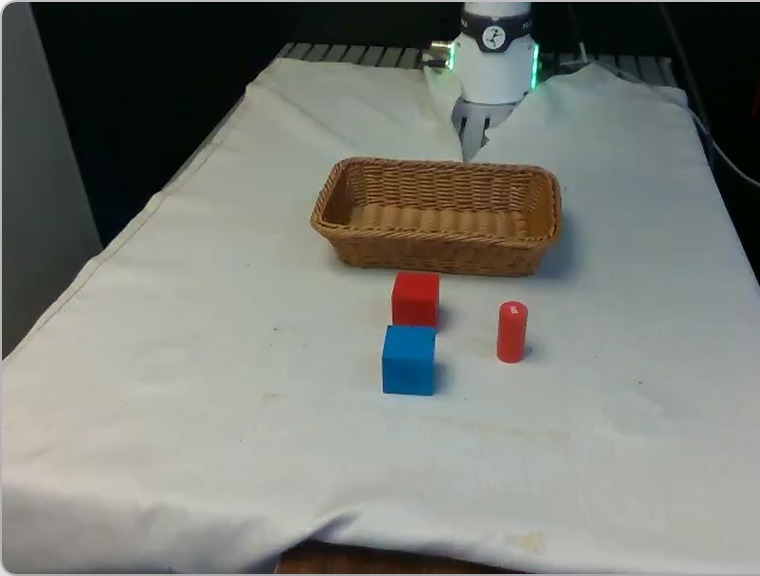} \\  

x3 & Open the drawer, pick up the red object to the left of the cylinder, place it into the basket, then close the drawer. & access + pick\_place & color + source\_relation &
\includegraphics[width=3.0cm, height=1.7cm, keepaspectratio]{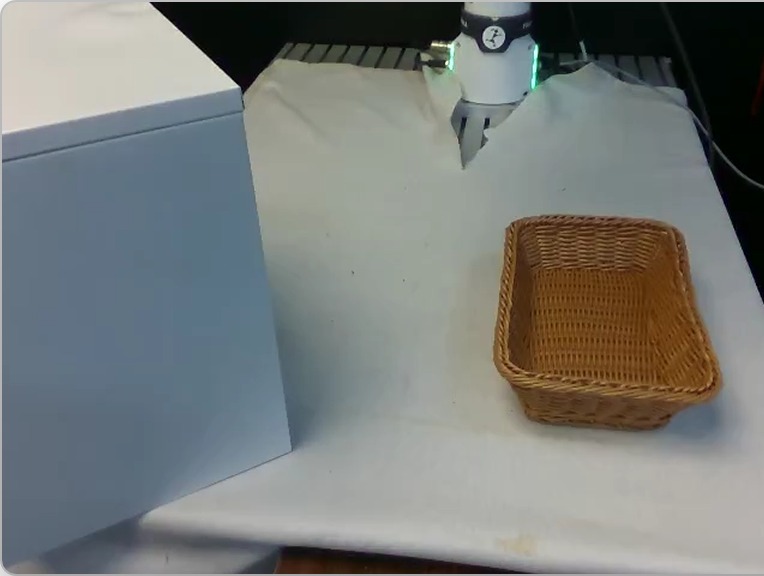} \\  

x4 & Open the drawer, pick up exactly two blocks, place them into the right basket, then close the drawer. & access + pick\_place & count + goal\_destination &
\includegraphics[width=3.0cm, height=1.7cm, keepaspectratio]{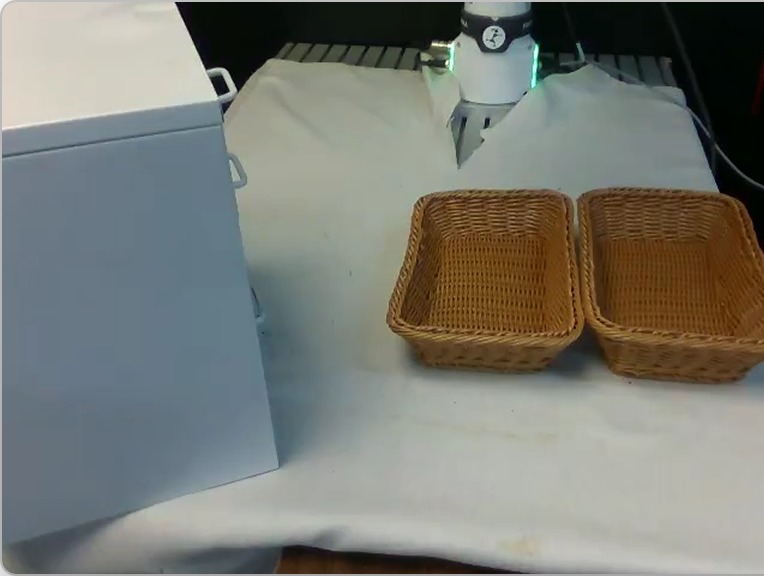} \\  

x5 & Place the two smallest bottles upright into the tray. & reorient & count + size &
\includegraphics[width=3.0cm, height=1.7cm, keepaspectratio]{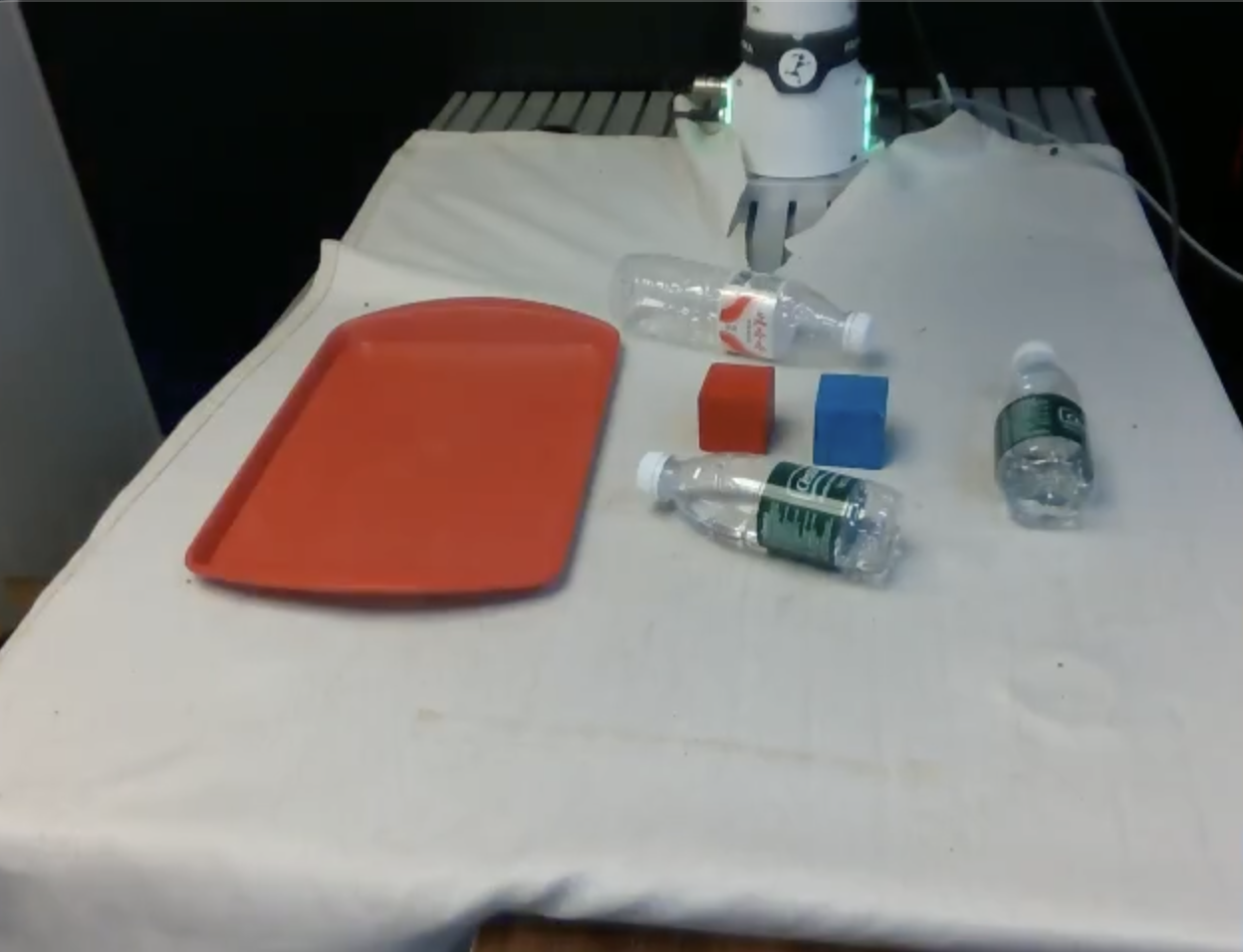} \\  

x6 & Place the green bottle upright onto the right tray. & reorient & color + goal\_destination &
\includegraphics[width=3.0cm, height=1.7cm, keepaspectratio]{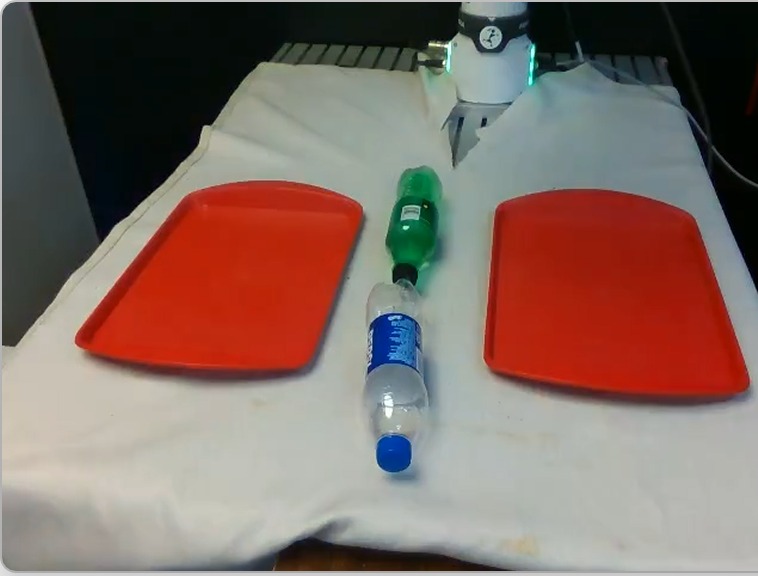} \\  

x7 & Push the non-red block into the right marked area. & push & exclusion + goal\_destination &
\includegraphics[width=3.0cm, height=1.7cm, keepaspectratio]{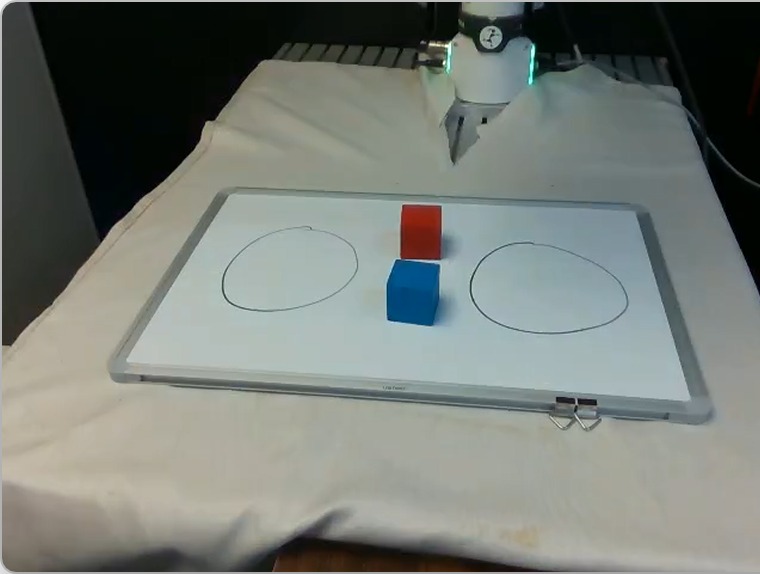} \\  

x8 & Push the triangular block to the left of the cylinder into the marked area. & push & shape + source\_relation &
\includegraphics[width=3.0cm, height=1.7cm, keepaspectratio]{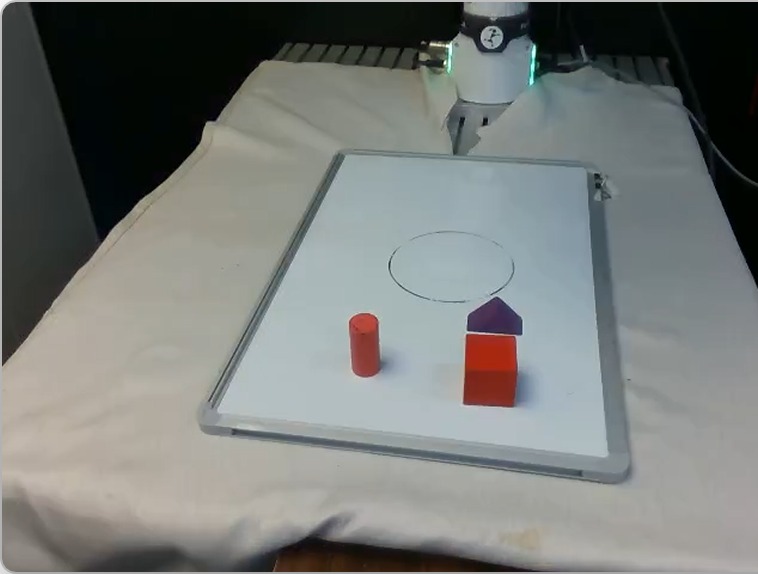} \\  

x9 & Stack the smallest red cube on top of the green cube. & stack & color + size &
\includegraphics[width=3.0cm, height=1.7cm, keepaspectratio]{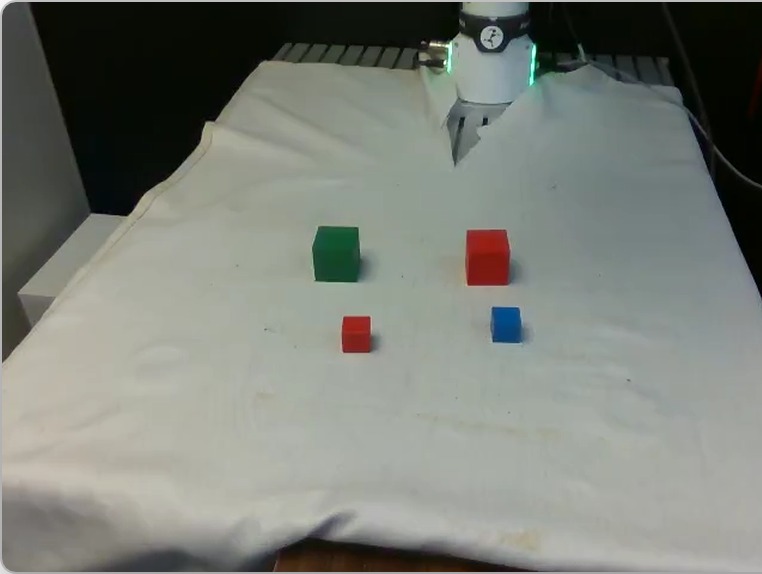} \\  

x10 & Stack the triangular block that is not purple on top of the cube. & stack & shape + exclusion &
\includegraphics[width=3.0cm, height=1.7cm, keepaspectratio]{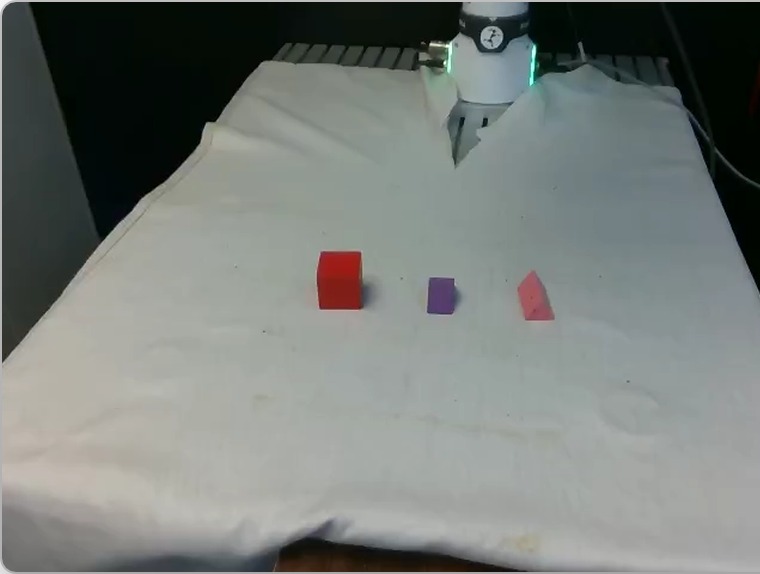} \\  

x11 & Pour the beans from the bean dish into the small bowl on the right. & pour & size + goal\_destination &
\includegraphics[width=3.0cm, height=1.7cm, keepaspectratio]{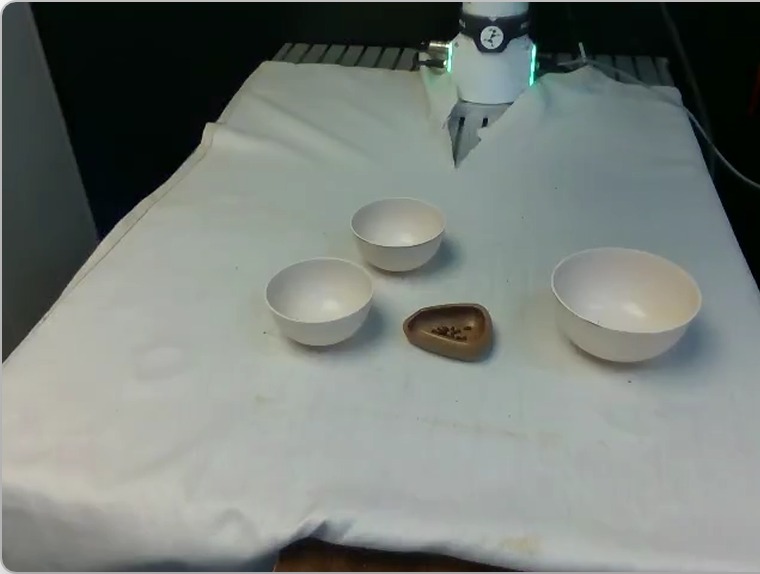} \\  

x12 & Pour the beans from the brown bean dish on the left into the bowl. & pour & color + source\_relation &
\includegraphics[width=3.0cm, height=1.7cm, keepaspectratio]{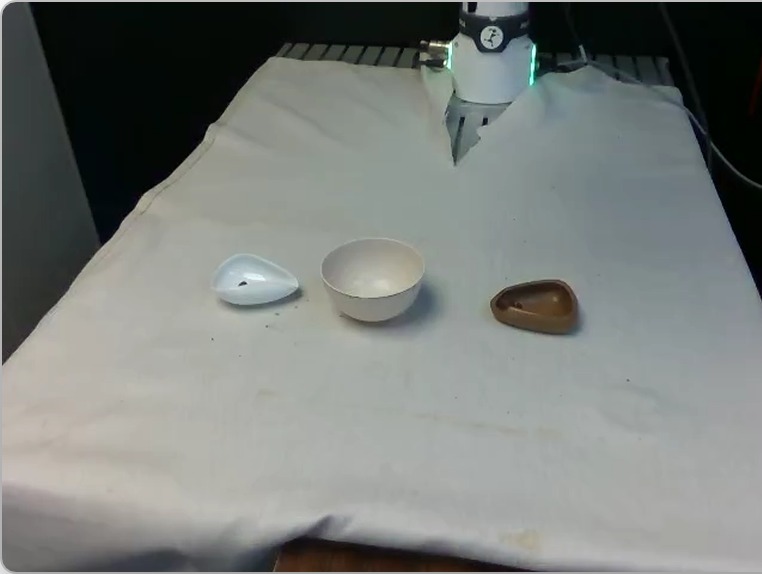} \\  

\end{longtable}

\begin{longtable}{
  >{\centering\arraybackslash}m{1.4cm}
  >{\raggedright\arraybackslash}m{6.0cm}
  >{\centering\arraybackslash}m{2.2cm}
  >{\centering\arraybackslash}m{4.5cm}
}
\caption{Dual-arm Motor Set (\texttt{dm1}--\texttt{dm8}). Bimanual counterparts of the single-arm Motor Set, additionally requiring two-arm coordination.}\label{tab:tasks-dm}\\
\toprule
Task\_id & Prompt & Instruction\_atom & Image \\
\midrule
\endfirsthead

\toprule
Task\_id & Prompt & Instruction\_atom & Image \\
\midrule
\endhead

\midrule
\multicolumn{4}{r}{\small Continued on next page} \\
\endfoot

\bottomrule
\endlastfoot

dm1 & Pick up the basket with the left hand and place the cube into the basket with the right hand. & pick\_place &
\includegraphics[width=3.0cm, height=1.7cm, keepaspectratio]{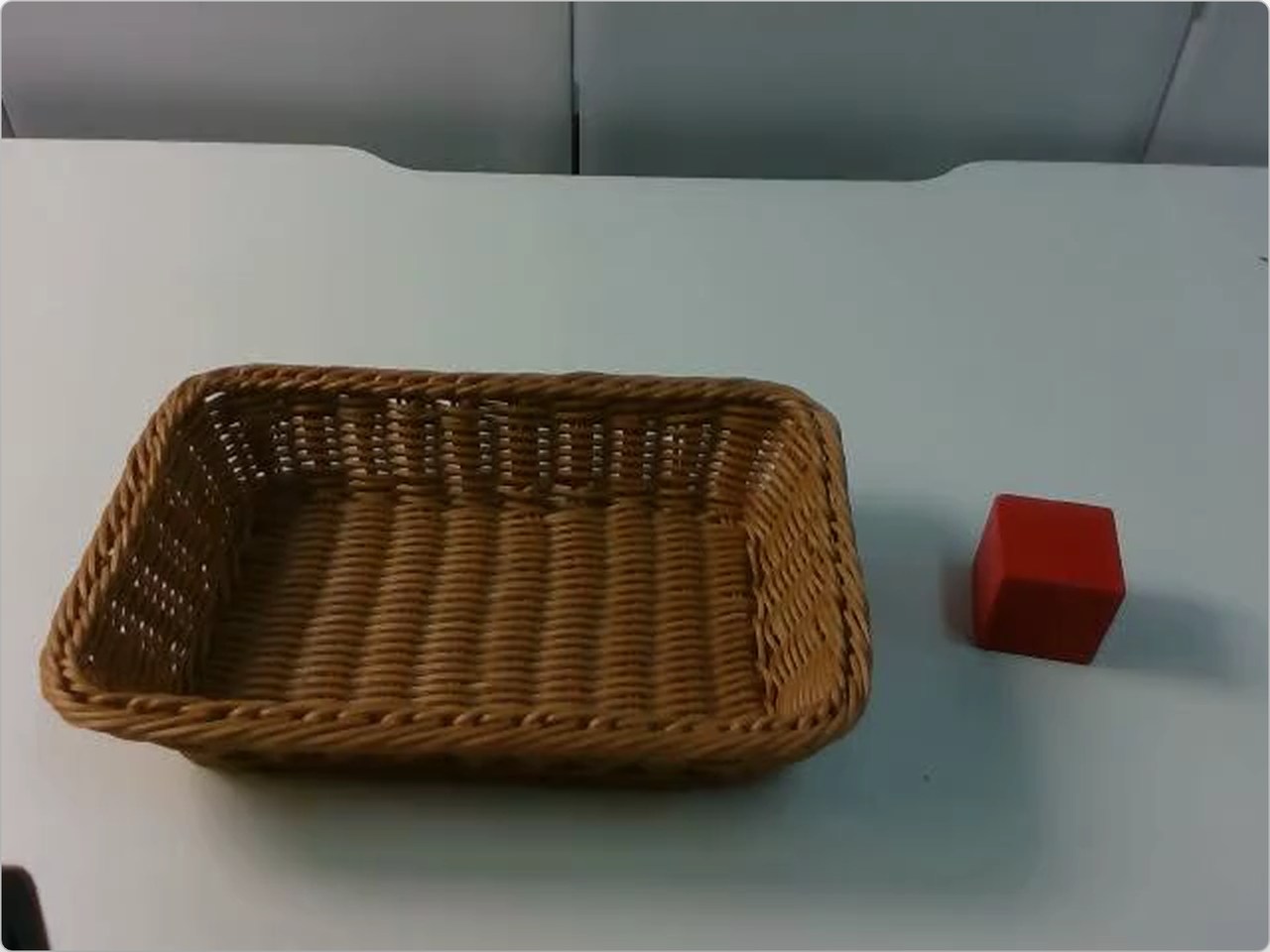} \\

dm2 & Pick up the basket with the left hand and place the ball into the basket with the right hand. & pick\_place &
\includegraphics[width=3.0cm, height=1.7cm, keepaspectratio]{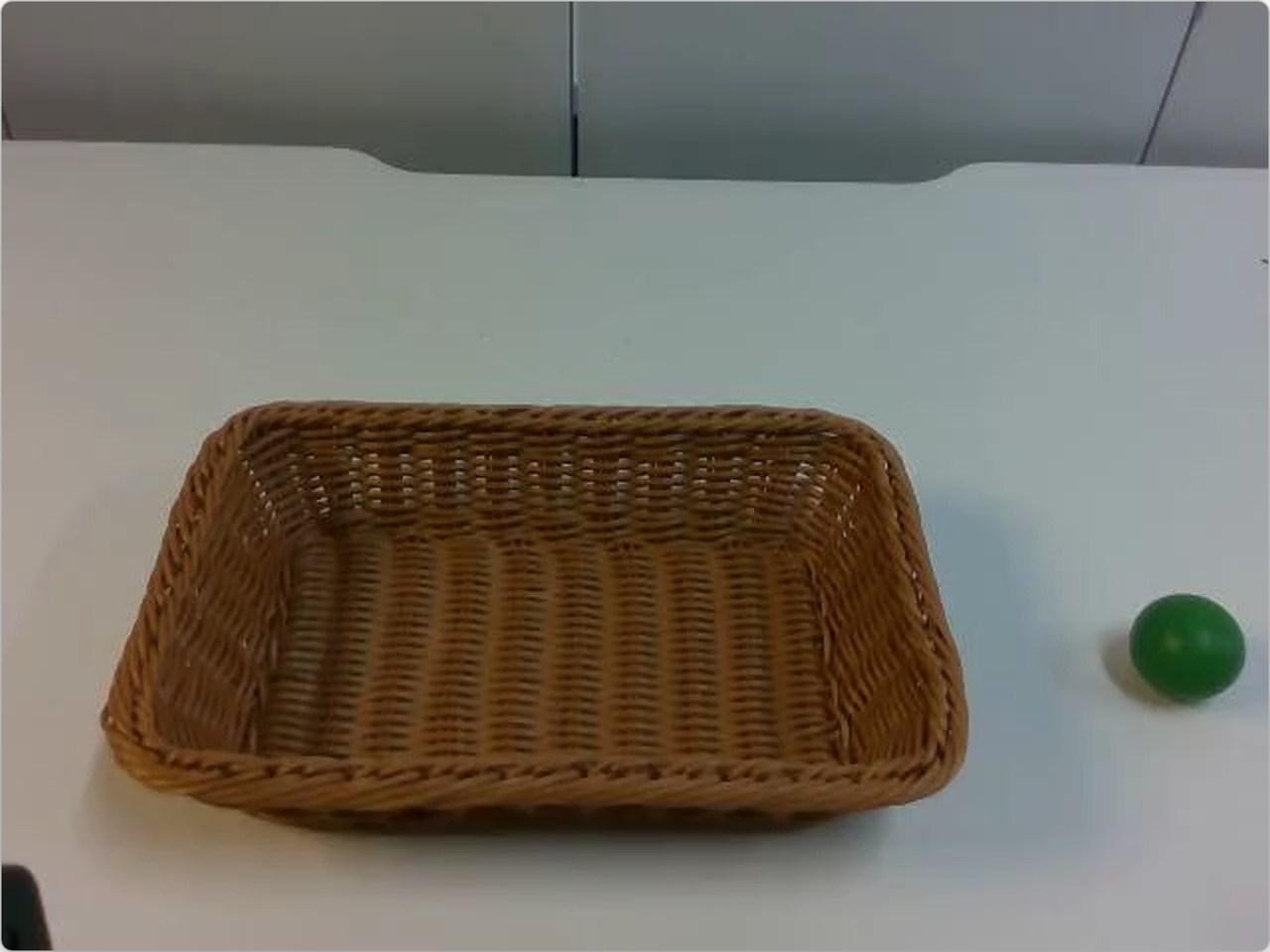} \\

dm3 & Place the bottles upright on the tray. & reorient &
\includegraphics[width=3.0cm, height=1.7cm, keepaspectratio]{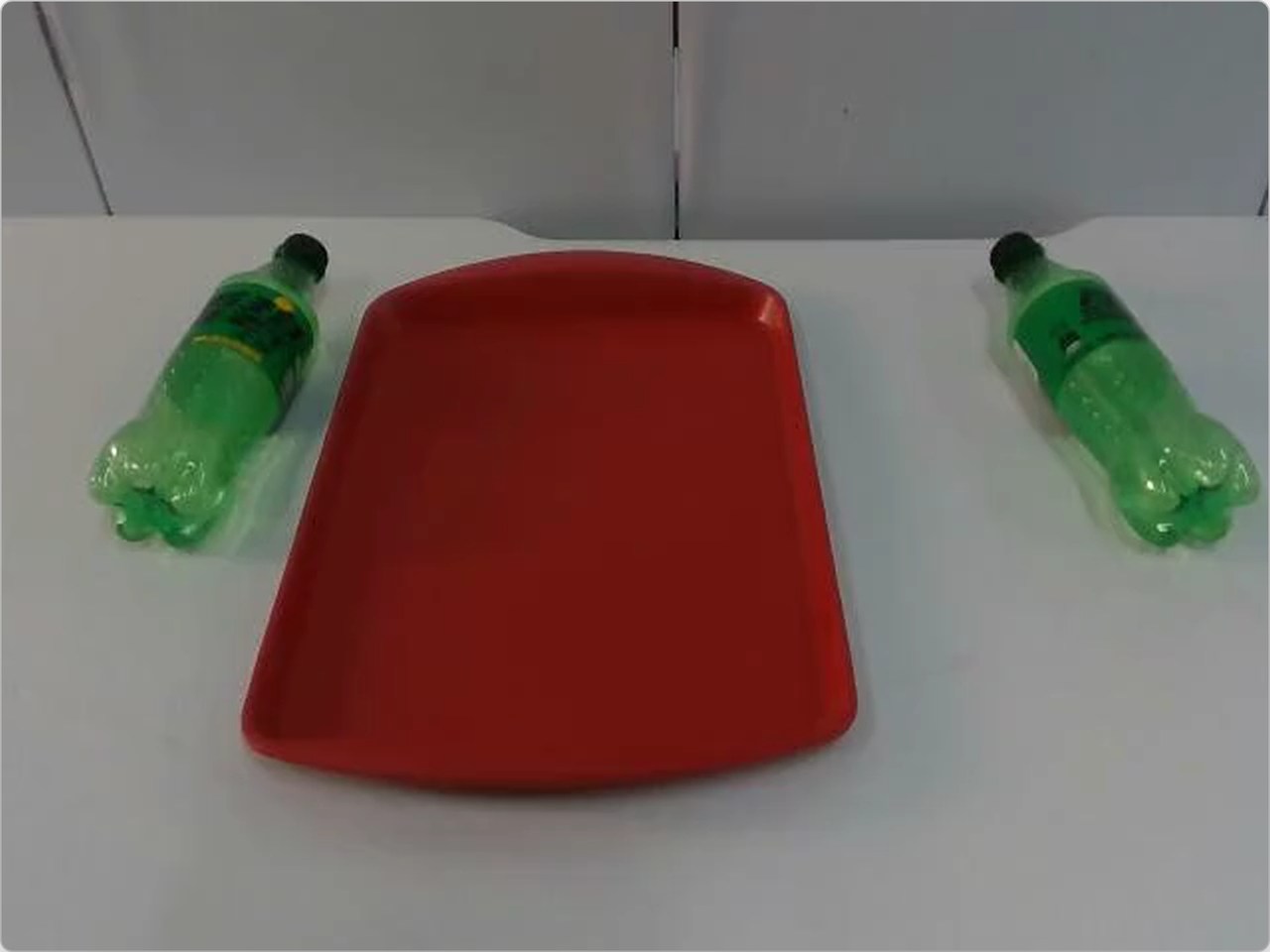} \\

dm4 & Push the two cubes into the marked area. & push &
\includegraphics[width=3.0cm, height=1.7cm, keepaspectratio]{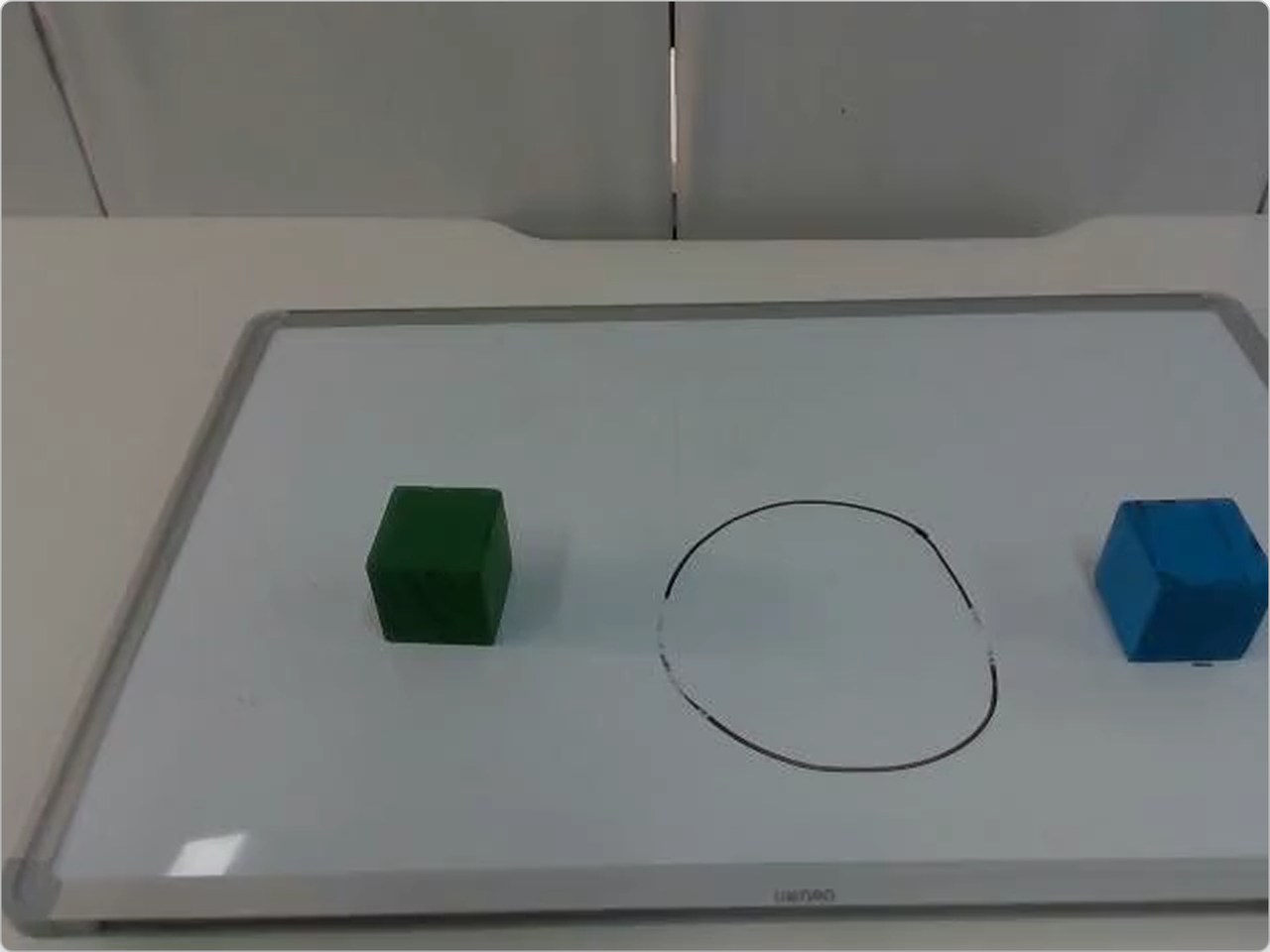} \\

dm5 & Pick up the bowl with the left hand and pour the coffee beans from the cup into the bowl with the right hand. & pour &
\includegraphics[width=3.0cm, height=1.7cm, keepaspectratio]{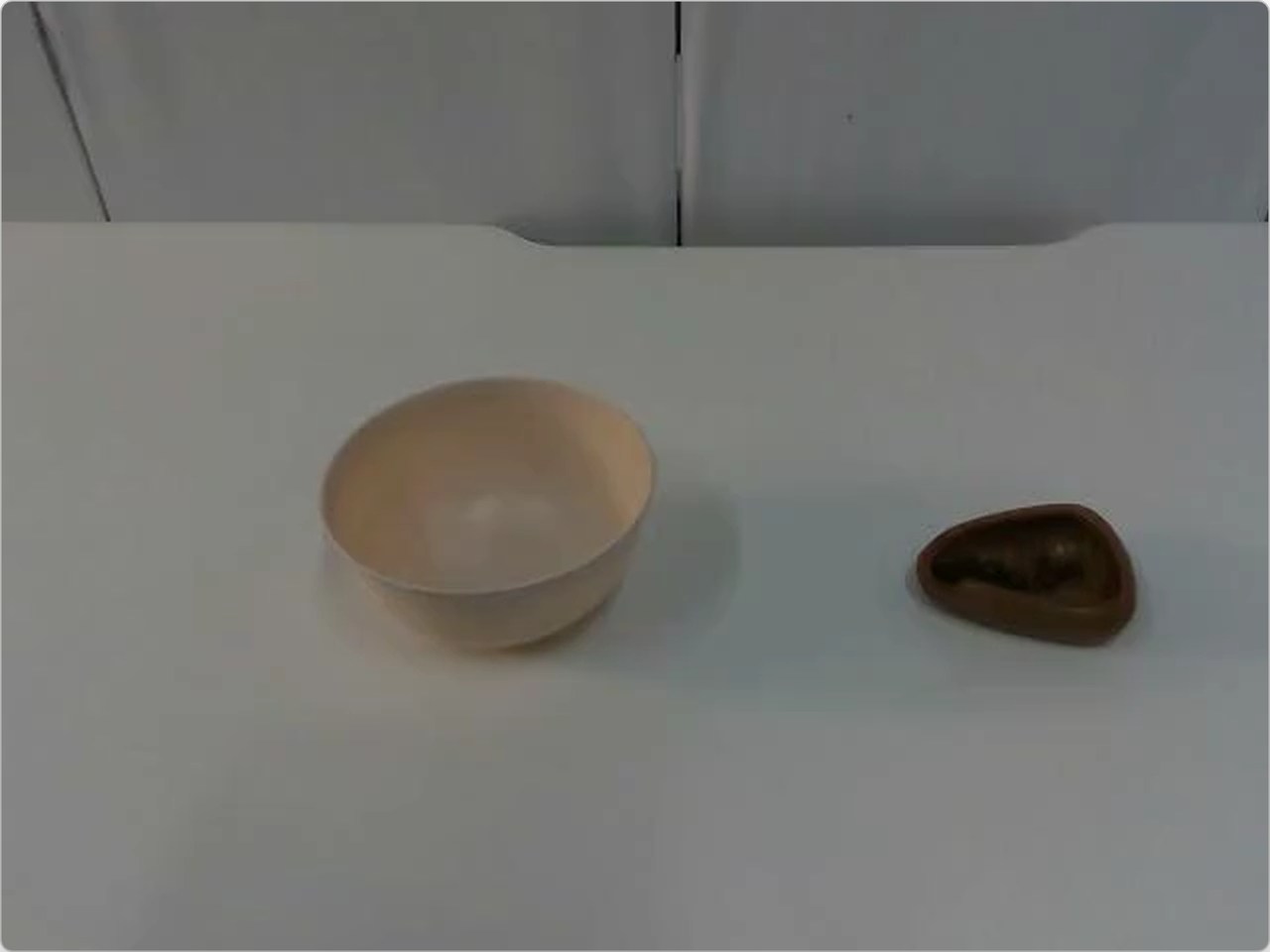} \\

dm6 & Open the drawer with the left hand, place the block into the drawer with the right hand, then close the drawer with the left hand. & access &
\includegraphics[width=3.0cm, height=1.7cm, keepaspectratio]{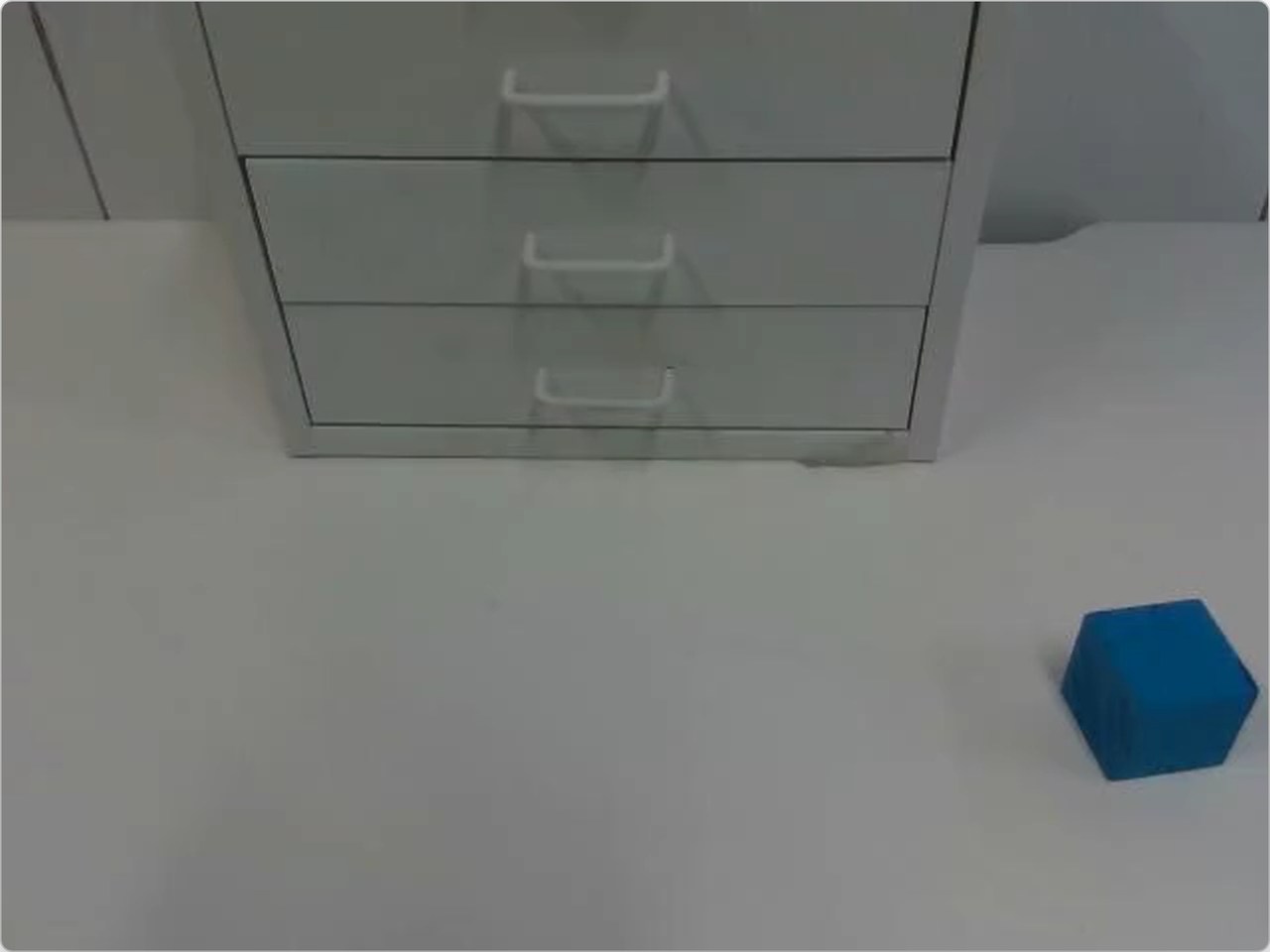} \\

dm7 & Pick up one cube with the left hand and another cube with the right hand. First place the cube in the right hand at the target position, then stack the cube in the left hand on top. & stack &
\includegraphics[width=3.0cm, height=1.7cm, keepaspectratio]{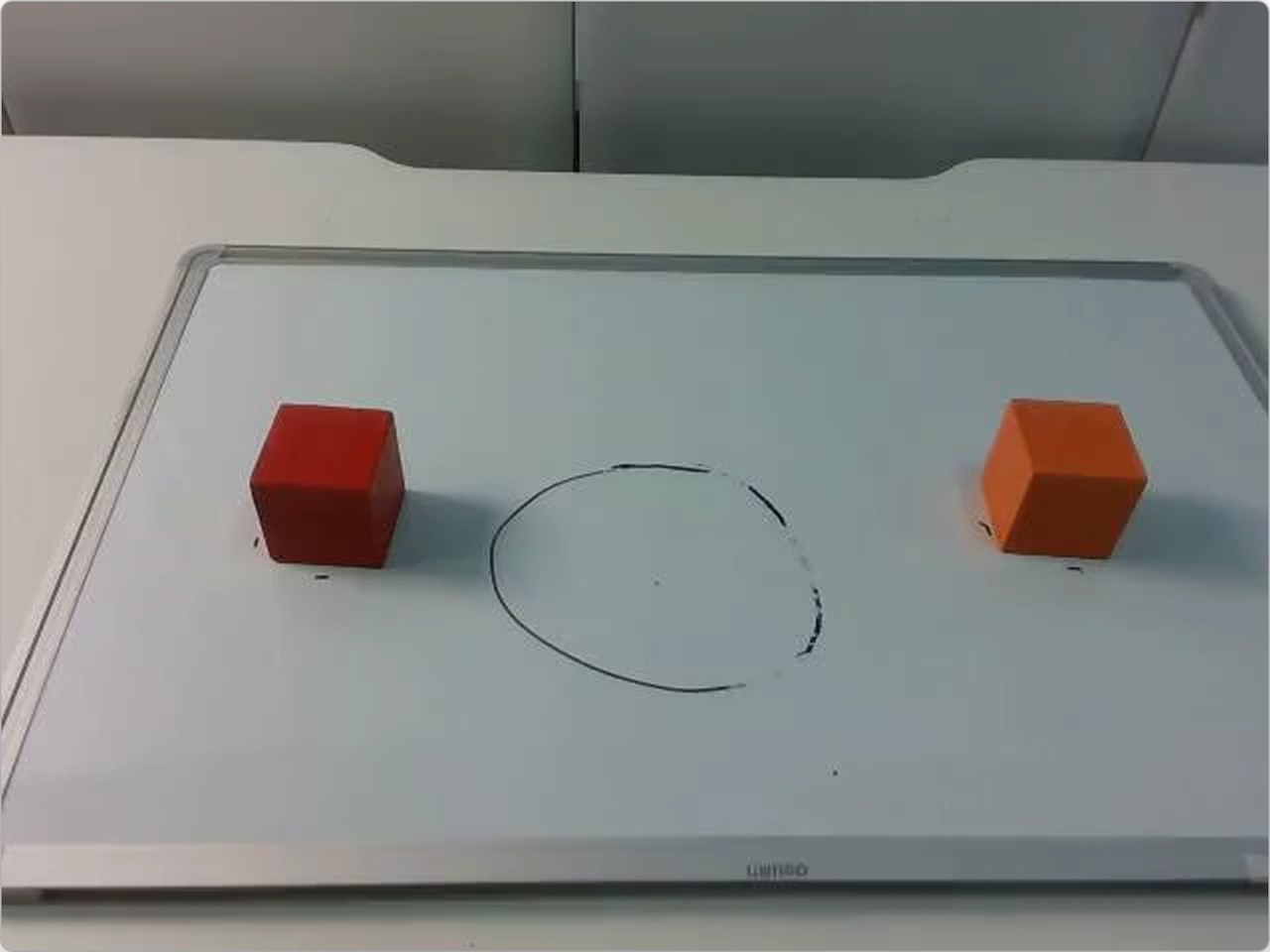} \\

dm8 & Open the box lid with the left hand, place the block into the box with the right hand, then close the box lid with the left hand. & access &
\includegraphics[width=3.0cm, height=1.7cm, keepaspectratio]{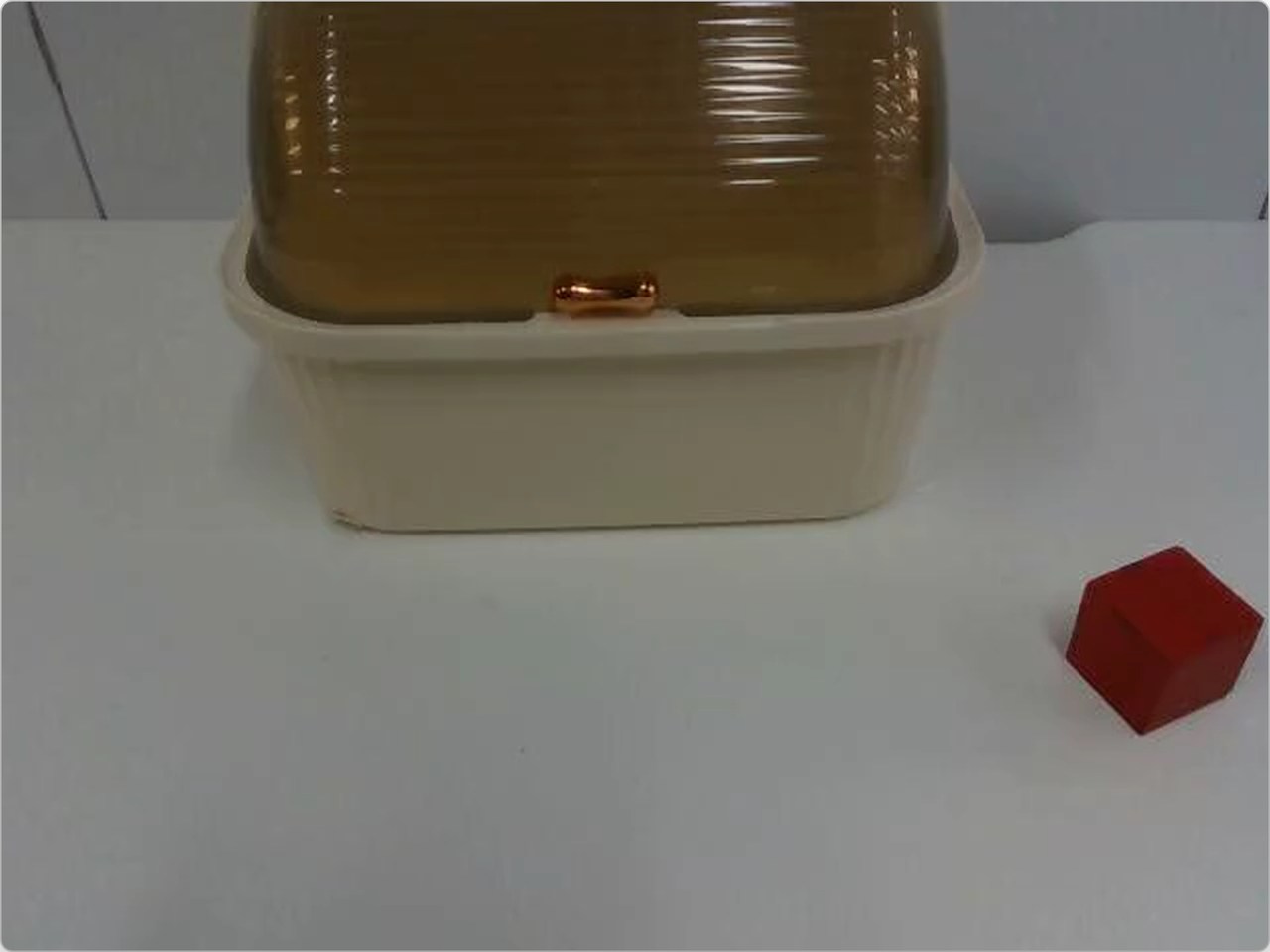} \\

\end{longtable}

\begin{longtable}{
  >{\centering\arraybackslash}m{1.4cm}
  >{\raggedright\arraybackslash}m{6.0cm}
  >{\centering\arraybackslash}m{2.2cm}
  >{\centering\arraybackslash}m{4.5cm}
}
\caption{Dual-arm Instruction Set (\texttt{di1}--\texttt{di7}). Bimanual counterparts of the single-arm Instruction Set, isolating one instruction atom each.}\label{tab:tasks-di}\\
\toprule
Task\_id & Prompt & Instruction\_atom & Image \\
\midrule
\endfirsthead

\toprule
Task\_id & Prompt & Instruction\_atom & Image \\
\midrule
\endhead

\midrule
\multicolumn{4}{r}{\small Continued on next page} \\
\endfoot

\bottomrule
\endlastfoot

di1 & Pick up the basket with the left hand and place the blue cube into the basket with the right hand. & color &
\includegraphics[width=3.0cm, height=1.7cm, keepaspectratio]{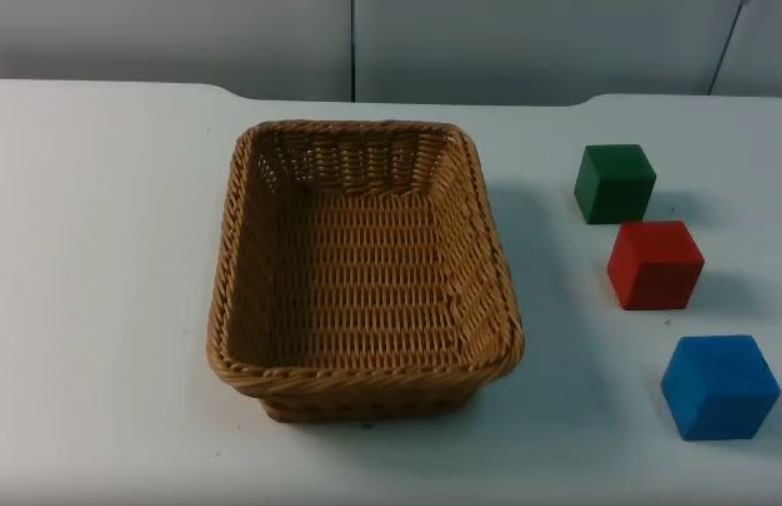} \\

di2 & Pick up the basket with the left hand and place the triangular block into the basket with the right hand. & shape &
\includegraphics[width=3.0cm, height=1.7cm, keepaspectratio]{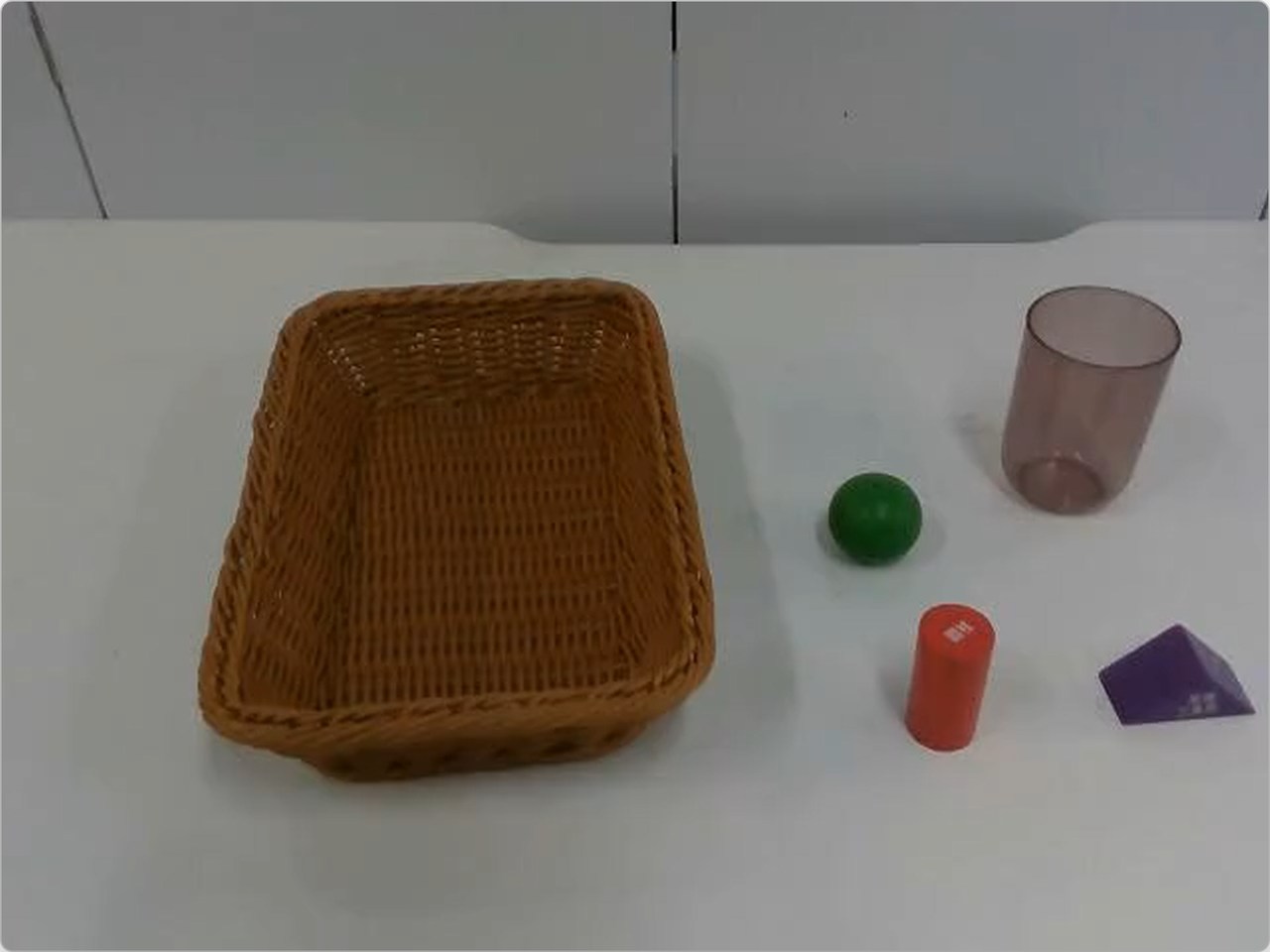} \\

di3 & Pick up the basket with the left hand and place the largest block into the basket with the right hand. & size &
\includegraphics[width=3.0cm, height=1.7cm, keepaspectratio]{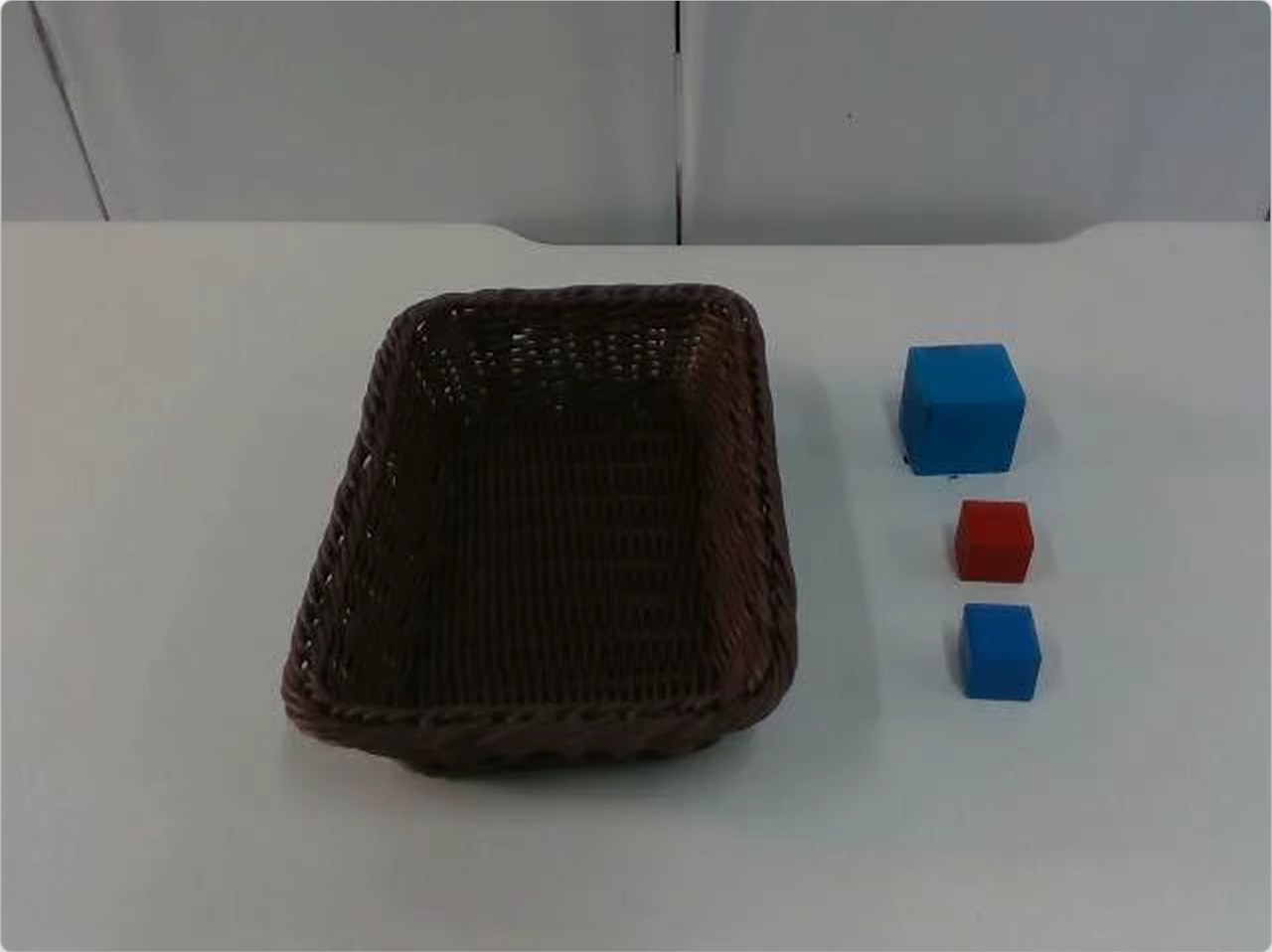} \\

di4 & Pick up the basket with the left hand and place the object to the left of the red cylinder into the basket with the right hand. & source\_relation &
\includegraphics[width=3.0cm, height=1.7cm, keepaspectratio]{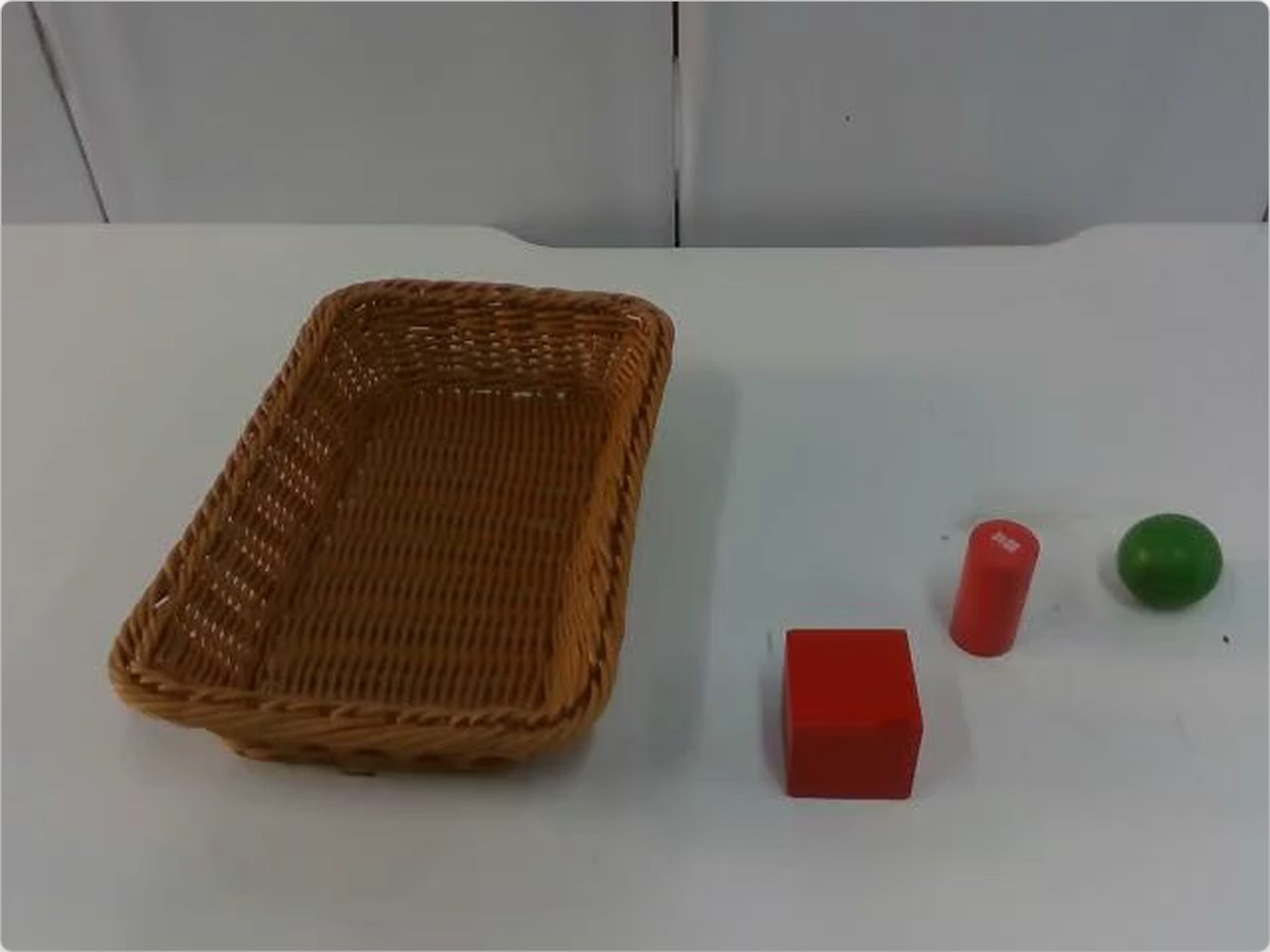} \\

di5 & Pick up one cube with each hand and place exactly two cubes into the same basket, leaving the remaining cube outside. & count &
\includegraphics[width=3.0cm, height=1.7cm, keepaspectratio]{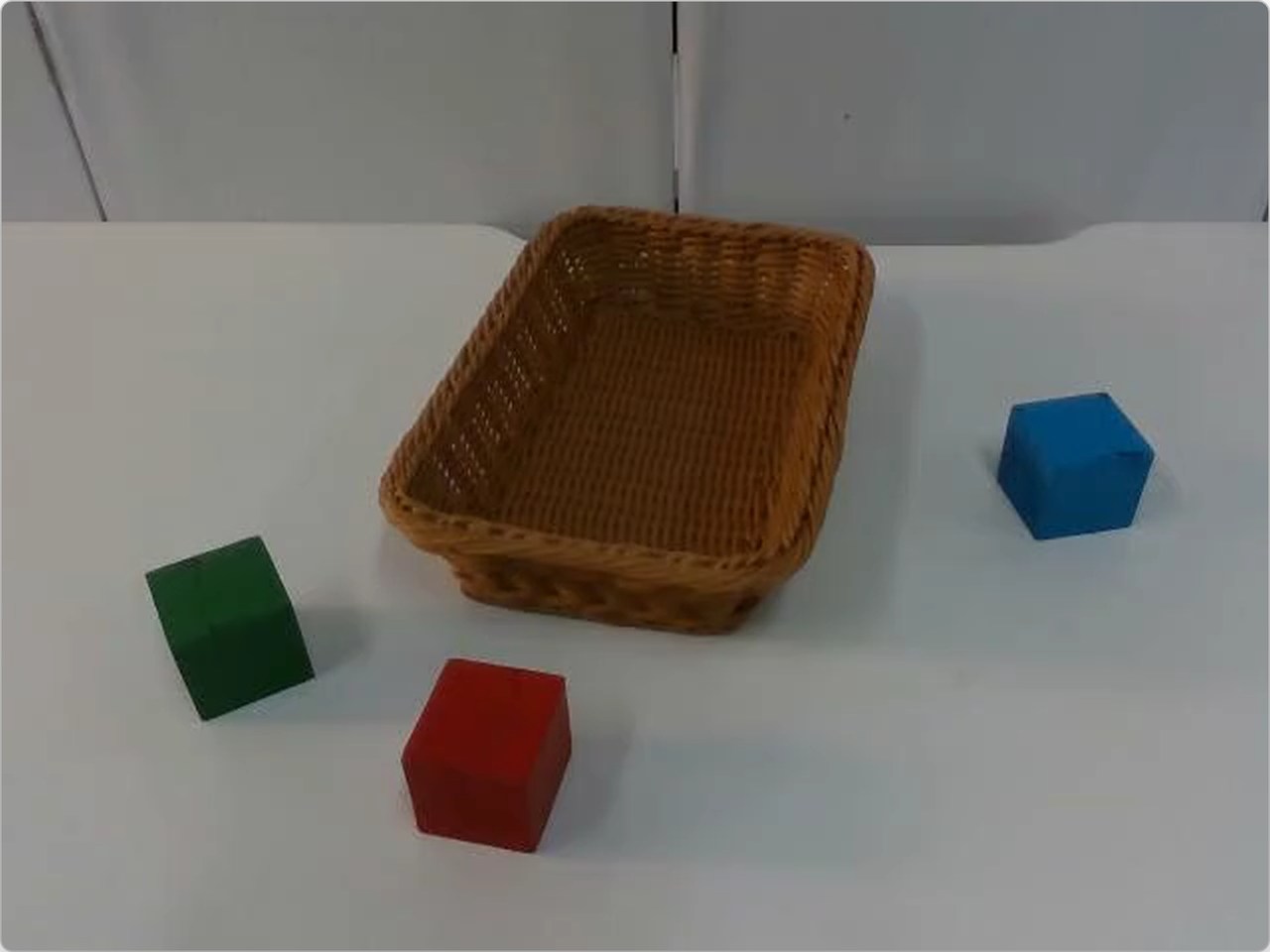} \\

di6 & Pick up one cube with the left hand and place it into the upper basket. Pick up one cube with the right hand and place it into the lower basket. & goal\_destination &
\includegraphics[width=3.0cm, height=1.7cm, keepaspectratio]{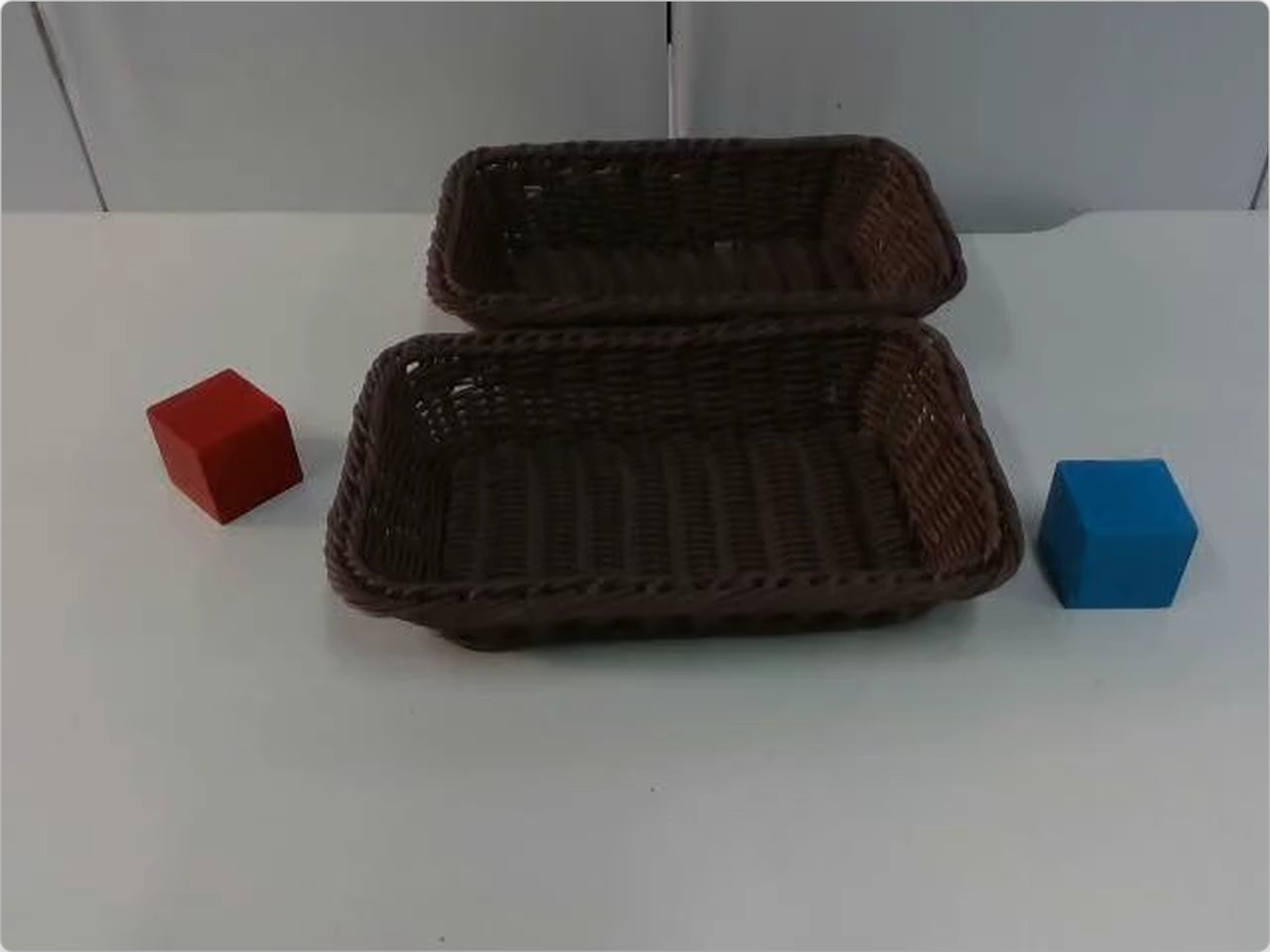} \\

di7 & Place all geometric objects except the red cube into the basket. & exclusion &
\includegraphics[width=3.0cm, height=1.7cm, keepaspectratio]{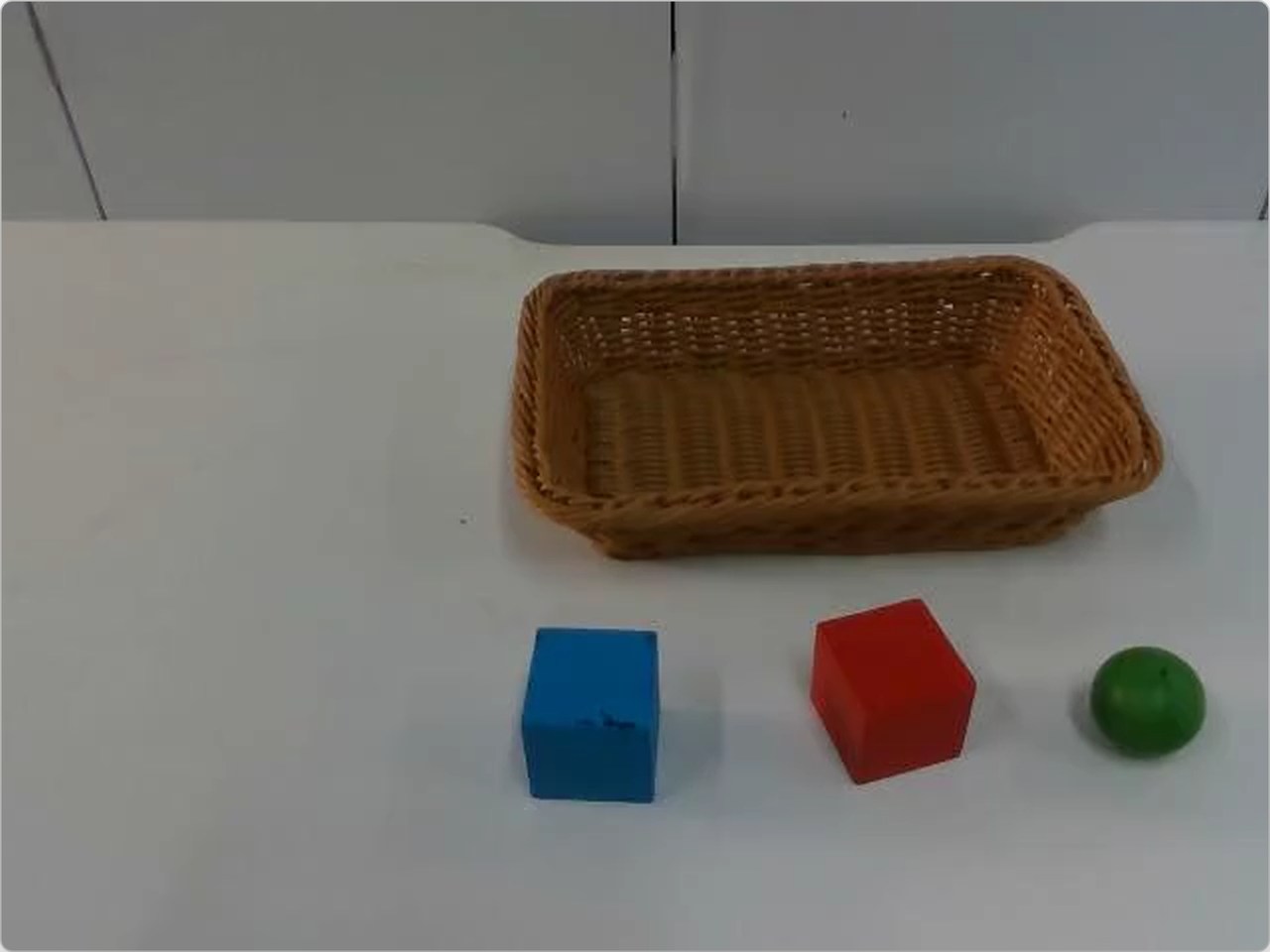} \\

\end{longtable}

\begin{longtable}{
  >{\centering\arraybackslash}m{1.4cm}    %
  >{\raggedright\arraybackslash}m{5.0cm}  %
  >{\centering\arraybackslash}m{1.6cm}    %
  >{\centering\arraybackslash}m{1.6cm}    %
  >{\centering\arraybackslash}m{4.5cm}    %
}
\caption{Dual-arm Composition Set (\texttt{dx1}--\texttt{dx12}). Bimanual held-out compositions combining one or more motor atoms with two or more instruction atoms.}\label{tab:tasks-dx}\\
\toprule
Task\_id & Prompt & Motor\_atom & Instruction\_atoms & Image \\
\midrule
\endfirsthead

\toprule
Task\_id & Prompt & Motor\_atom & Instruction\_atoms & Image \\
\midrule
\endhead

\midrule
\multicolumn{5}{r}{\small Continued on next page} \\
\endfoot

\bottomrule
\endlastfoot

dx1 &Pick up one red cube with each hand and place exactly two red cubes into the basket.\newline & pick\_place & color + count &
\includegraphics[width=3.0cm, height=1.7cm, keepaspectratio]{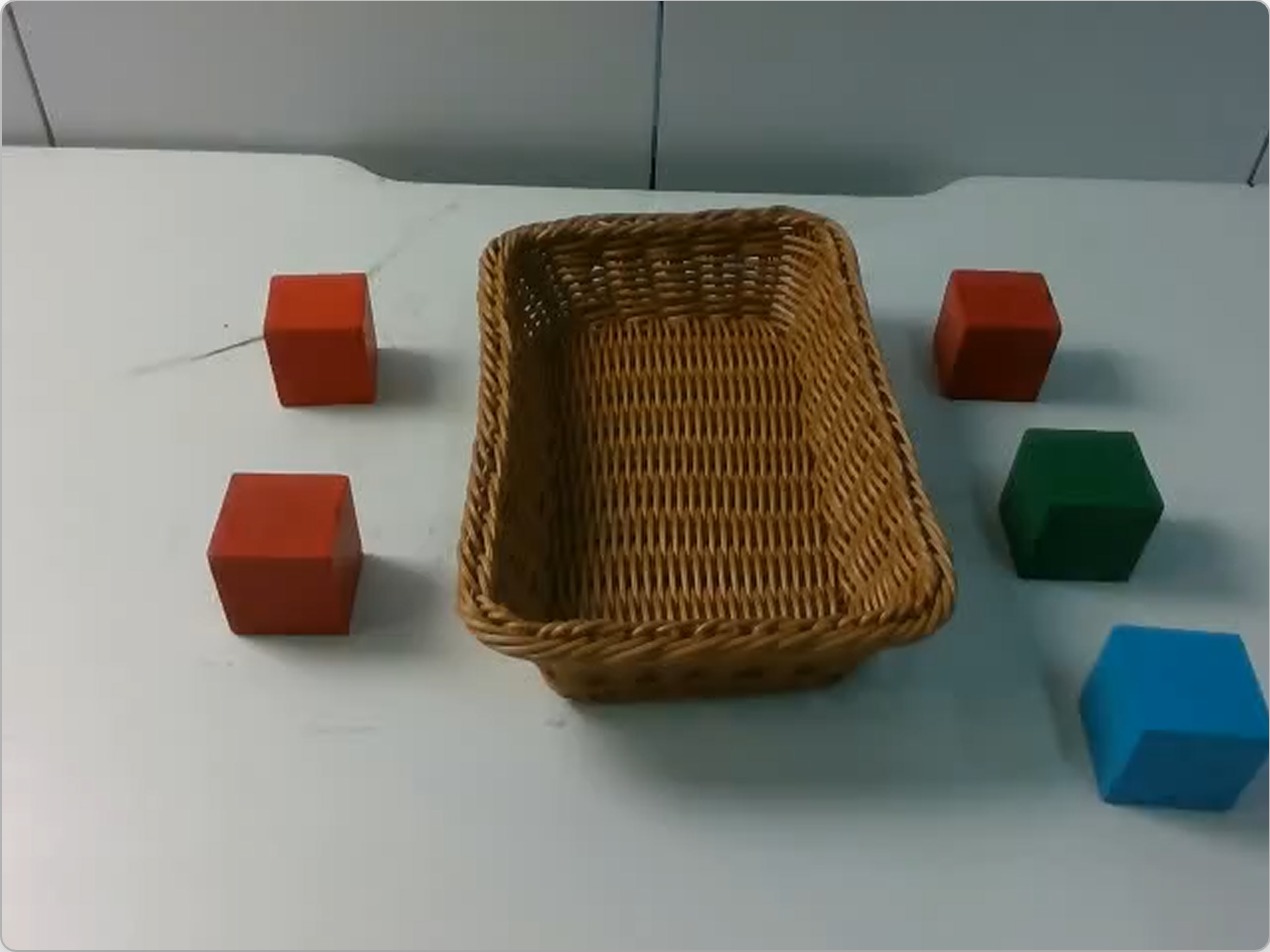} \\ 

dx2 & Pick up the basket with the left hand and place the all non-red objects to the right of the cylinder into the basket with the right hand. \newline& pick\_place & source\_relation + exclusion &
\includegraphics[width=3.0cm, height=1.7cm, keepaspectratio]{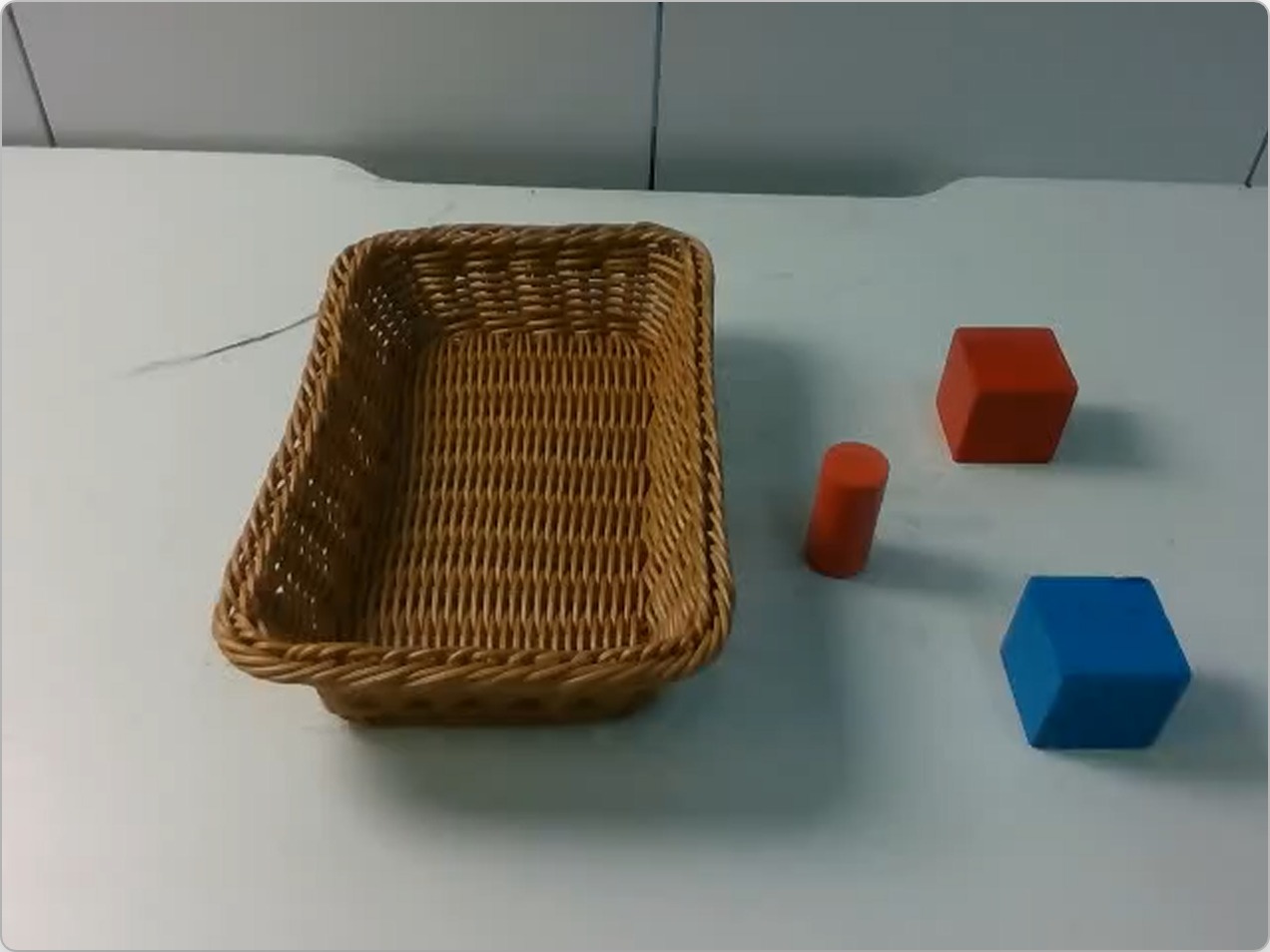} \\ 

dx3 & Open the drawer with the left hand, pick up the red cube to the left of the cylinder with the right hand, place it into the basket, then close the drawer with the left hand.\newline & access + pick\_place & color + source\_relation &
\includegraphics[width=3.0cm, height=1.7cm, keepaspectratio]{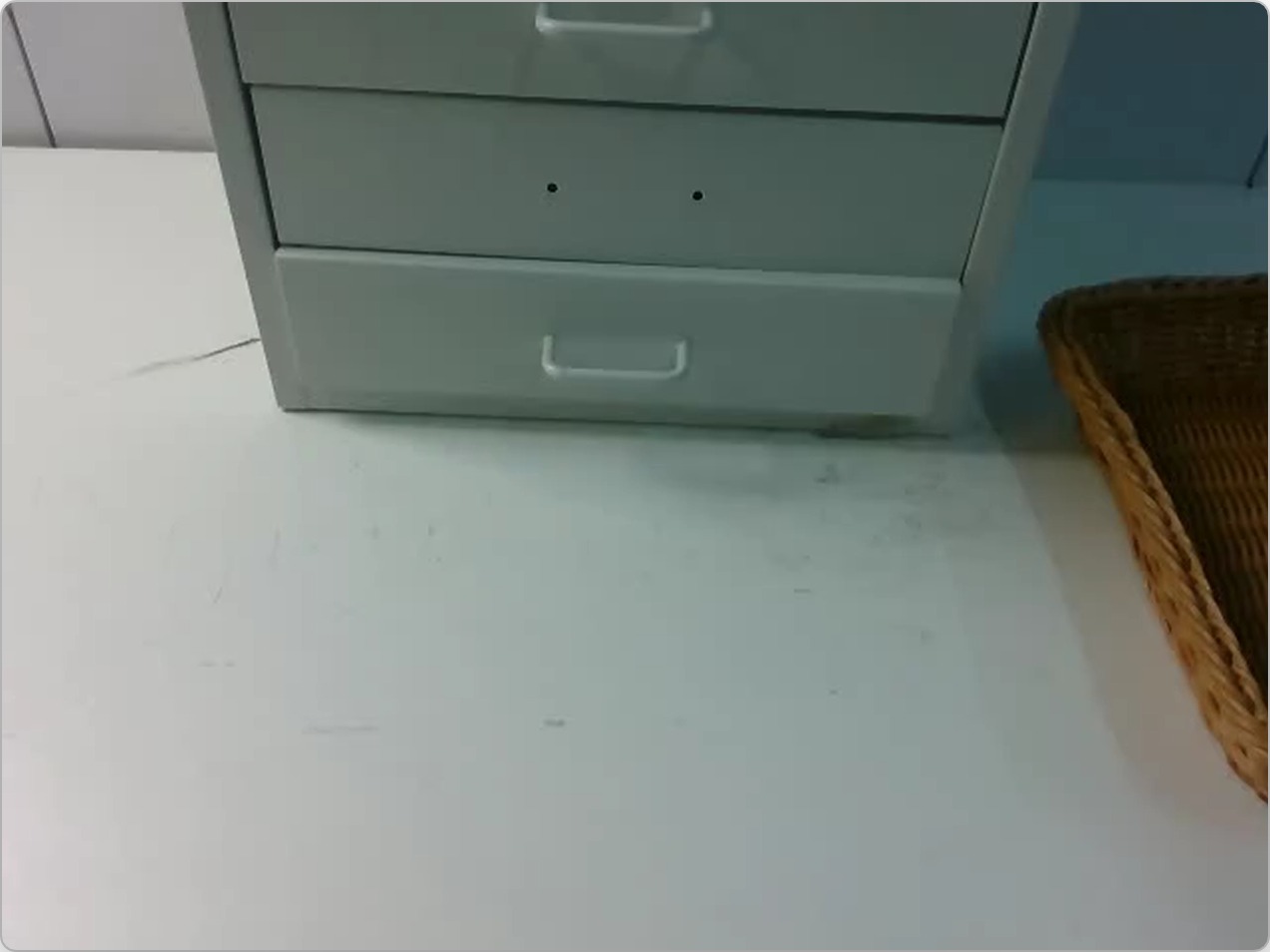} \\ 

dx4 &Open the drawer with the left hand, pick up one block with each hand, place exactly two blocks into the right basket, then close the drawer with the left hand.\newline & access + pick\_place & count + goal\_destination &
\includegraphics[width=3.0cm, height=1.7cm, keepaspectratio]{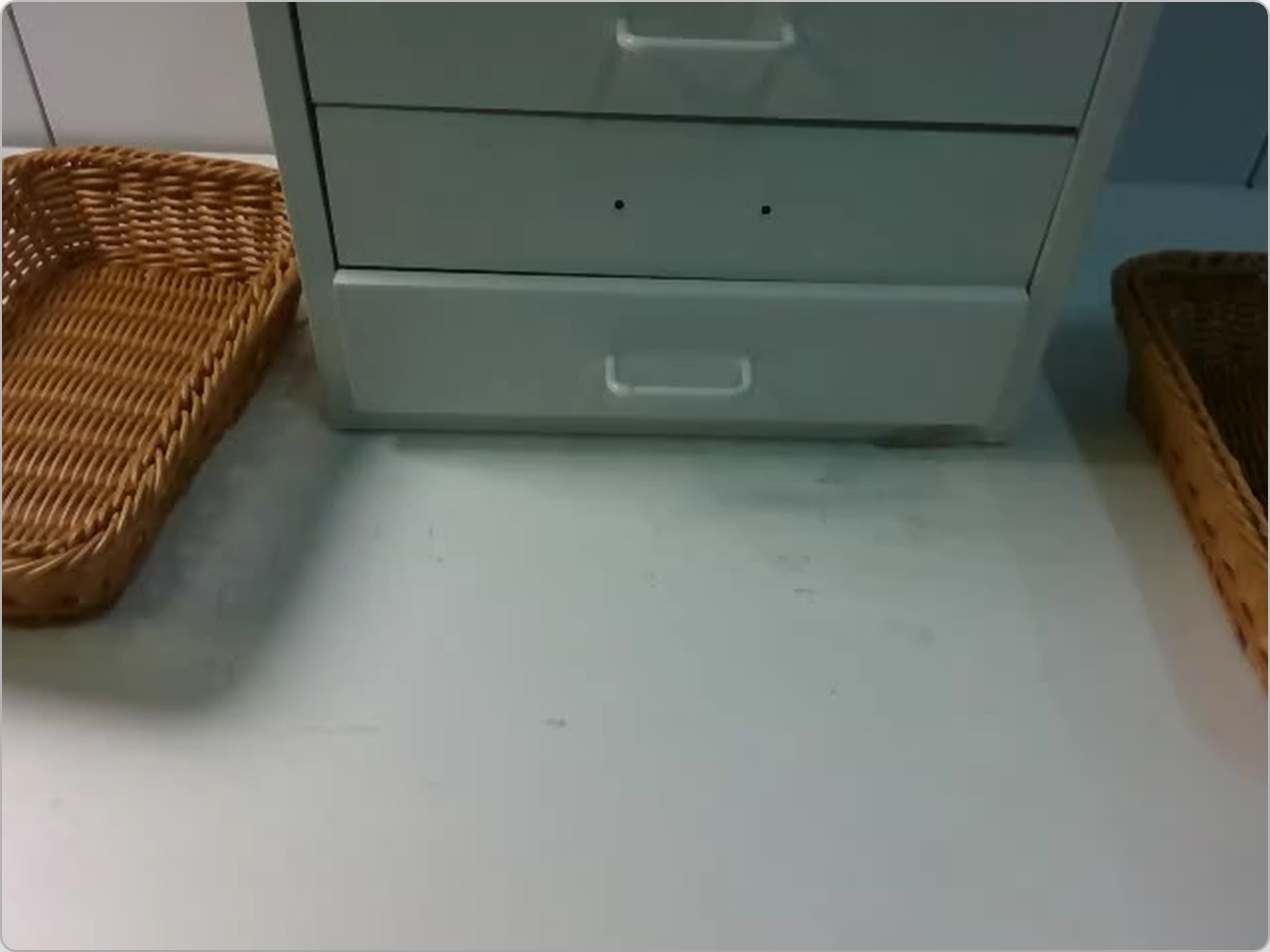} \\ 

dx5 & Place the two smallest bottles upright into the tray. \newline& reorient & count + size &
\includegraphics[width=3.0cm, height=1.7cm, keepaspectratio]{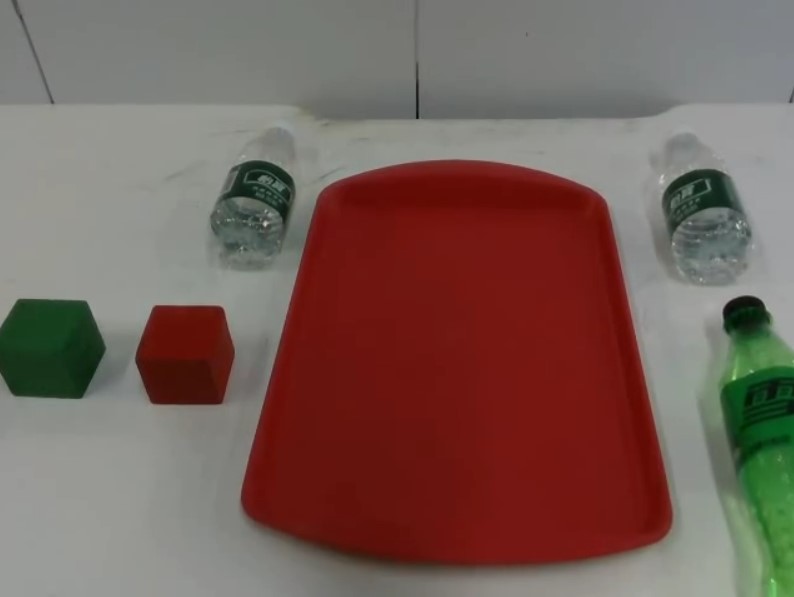} \\ 

dx6 &Pick up the two green bottles lying down, one with each hand, and place them upright in the right target area, leaving the bottles of other colors in place. \newline& reorient & color + goal\_destination &
\includegraphics[width=3.0cm, height=1.7cm, keepaspectratio]{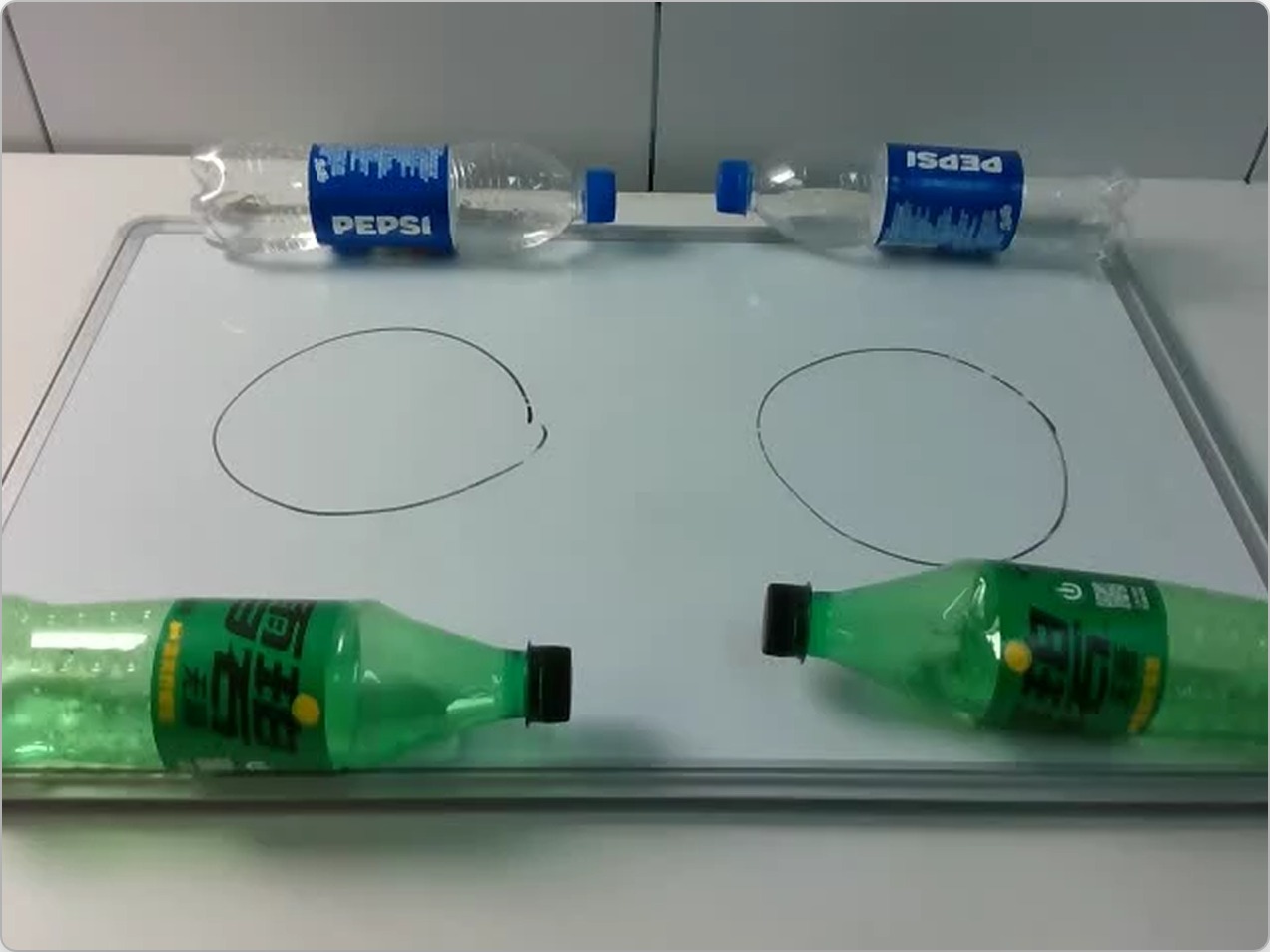} \\ 

dx7 &Push the block that is not red into the right marked area.\newline & push & exclusion + goal\_destination &
\includegraphics[width=3.0cm, height=1.7cm, keepaspectratio]{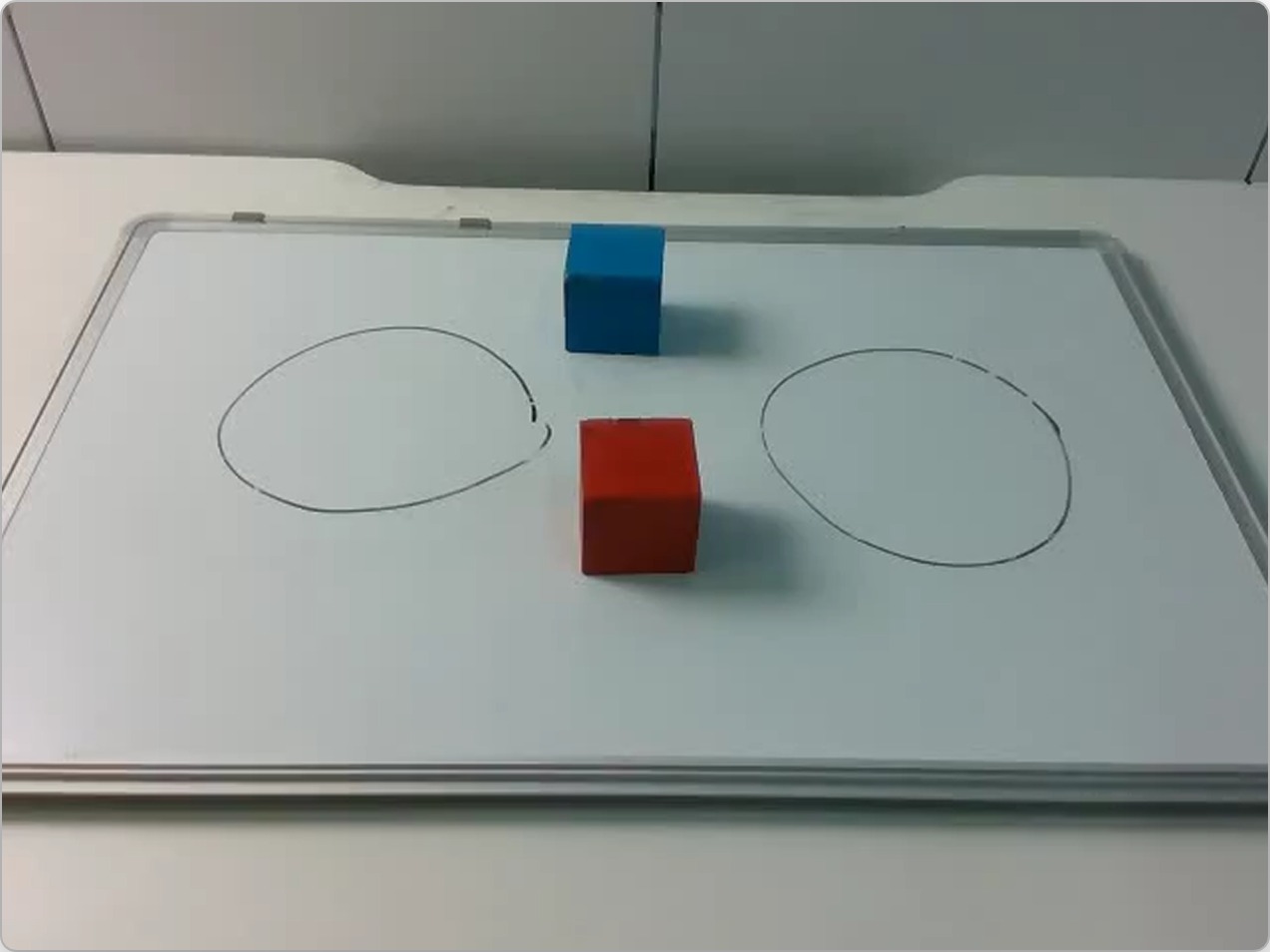} \\ 

dx8 &Push the triangular block to the left of the cylinder into the marked area.\newline & push & shape + source\_relation &
\includegraphics[width=3.0cm, height=1.7cm, keepaspectratio]{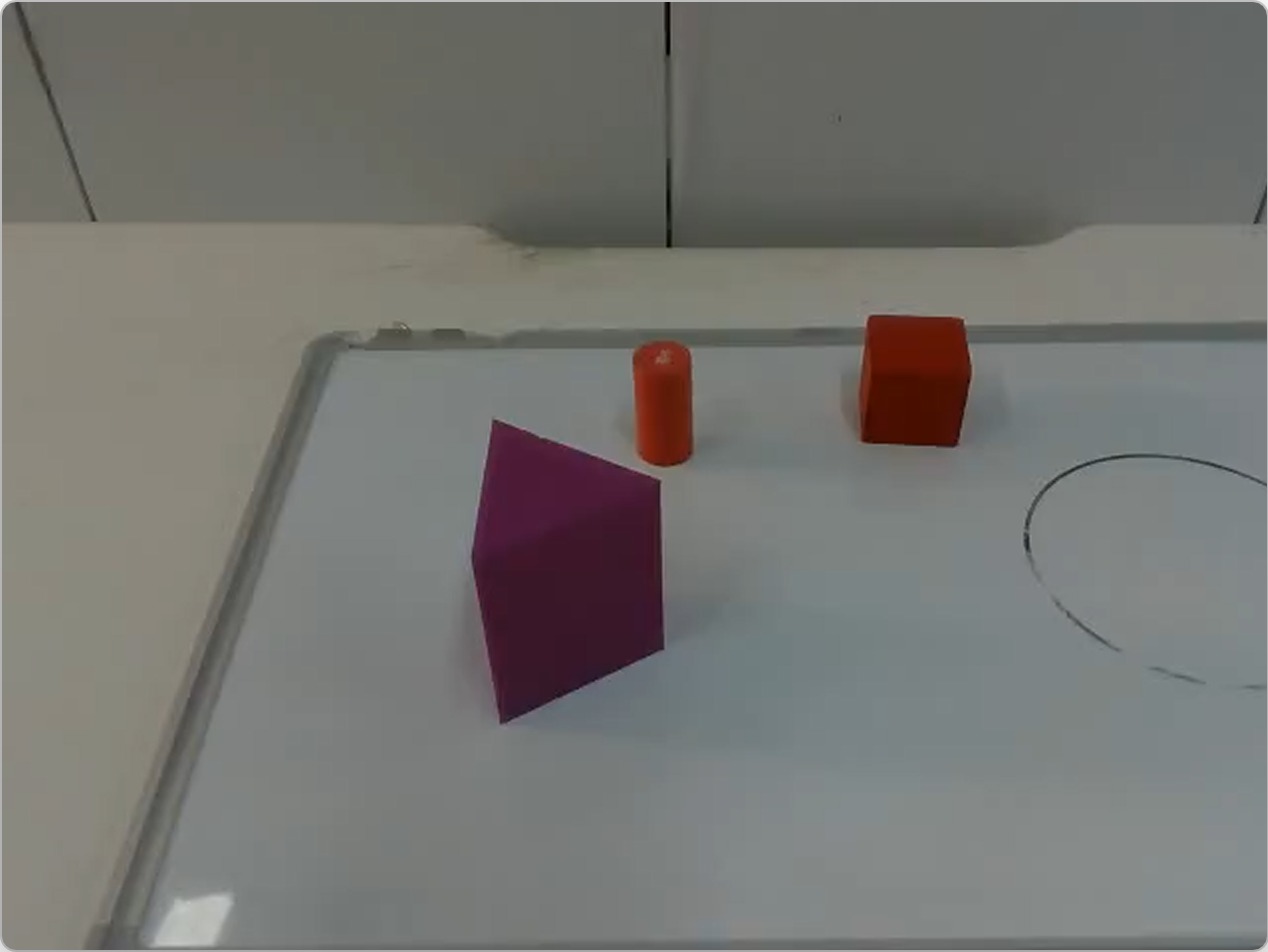} \\ 

dx9 &Place the green cube on top of the largest blue cube with your left hand, and put the smallest red cube onto the green cube with your right hand.\newline & stack & color + size &
\includegraphics[width=3.0cm, height=1.7cm, keepaspectratio]{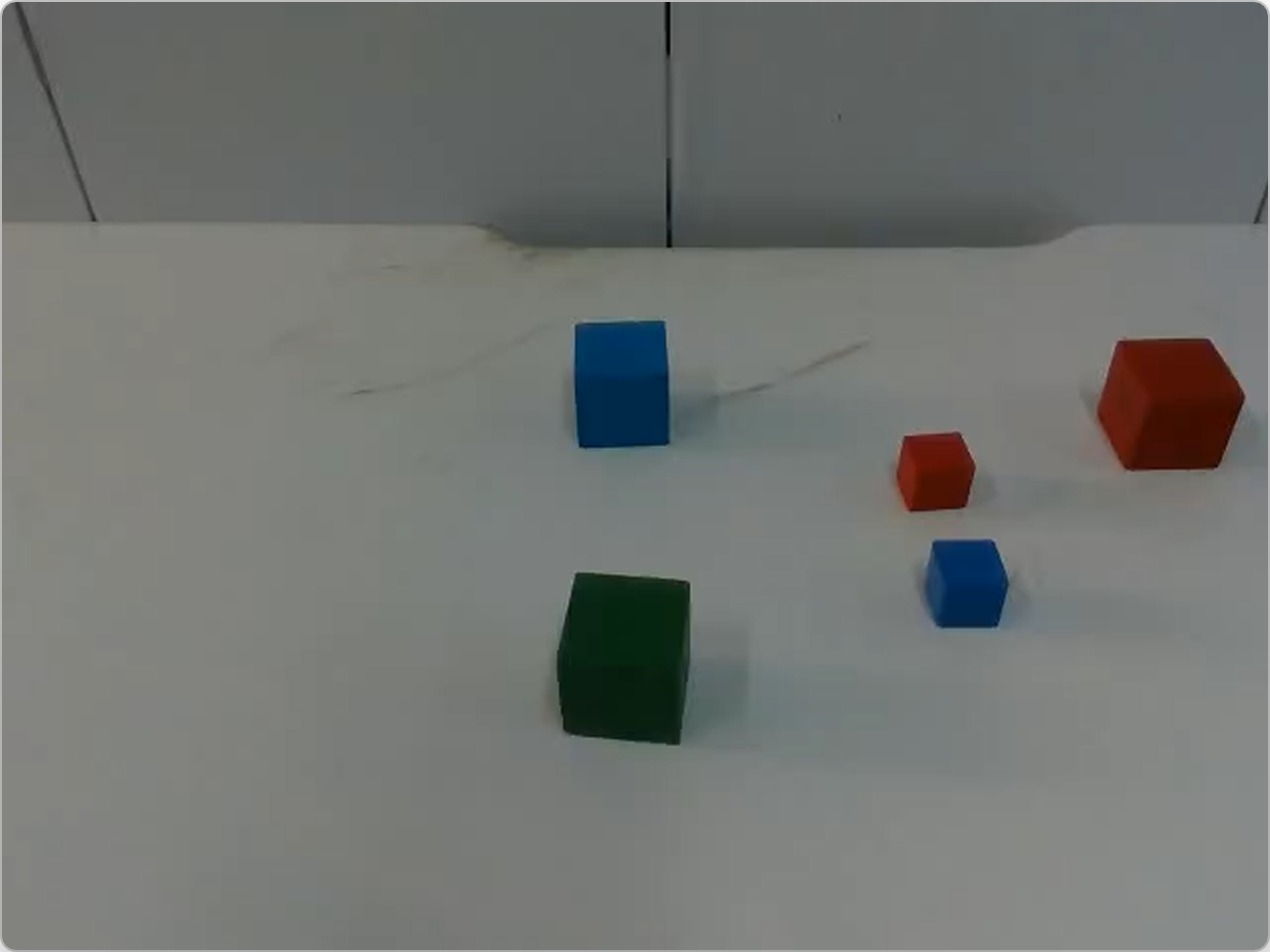} \\ 

dx10 &Pick up the triangular block with the left hand and pick up a cube that is not red with the right hand. First place the cube at the target position, then stack the triangular block on top of it.\newline & stack & color + shape + exclusion &
\includegraphics[width=3.0cm, height=1.7cm, keepaspectratio]{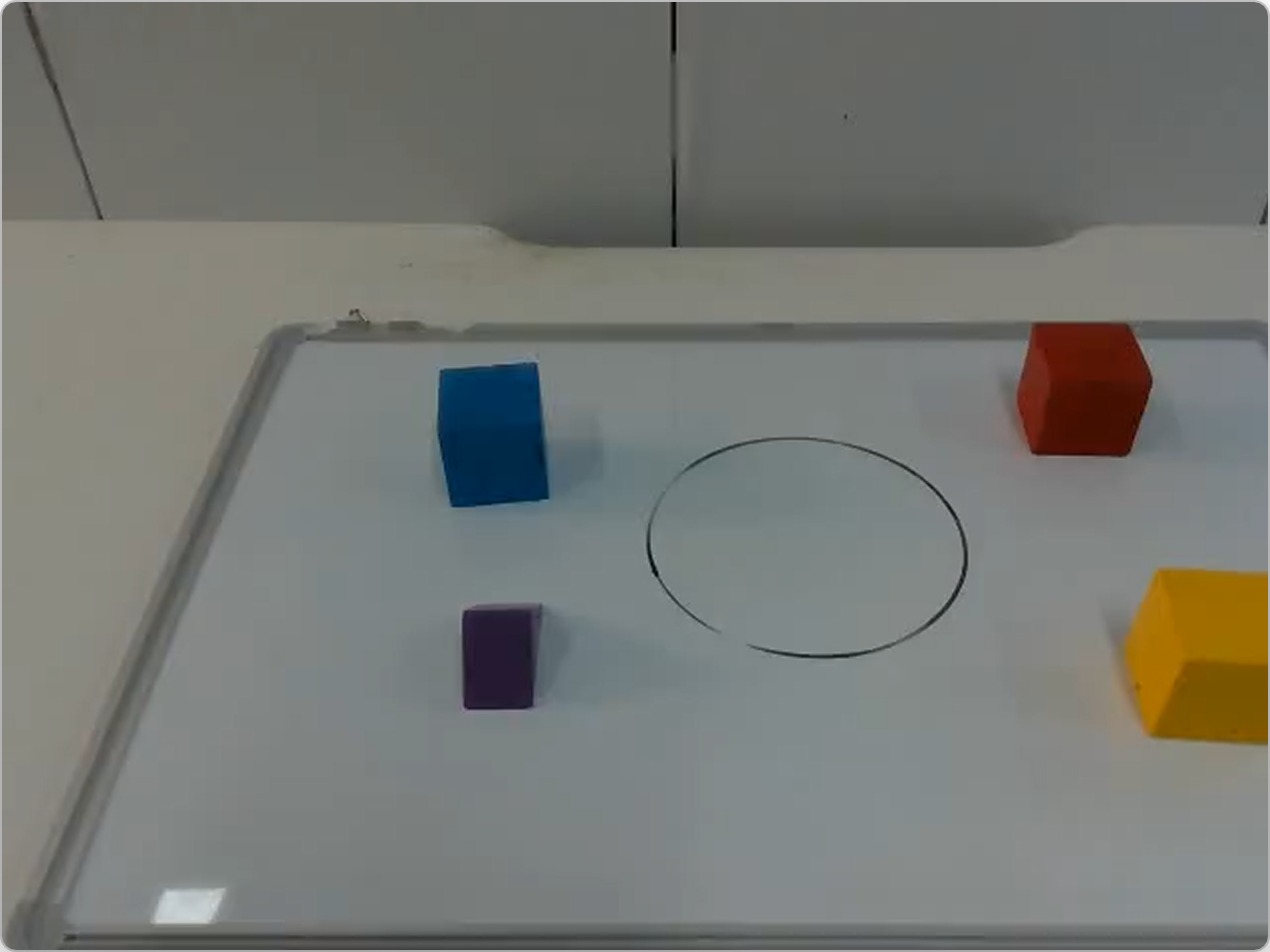} \\ 

dx11 & Pick up the smaller of the two bowls with the left hand, pick up the bean dish containing beans with the right hand, and pour the beans from the bean dish into that smaller bowl.\newline & pour & size + goal\_destination &
\includegraphics[width=3.0cm, height=1.7cm, keepaspectratio]{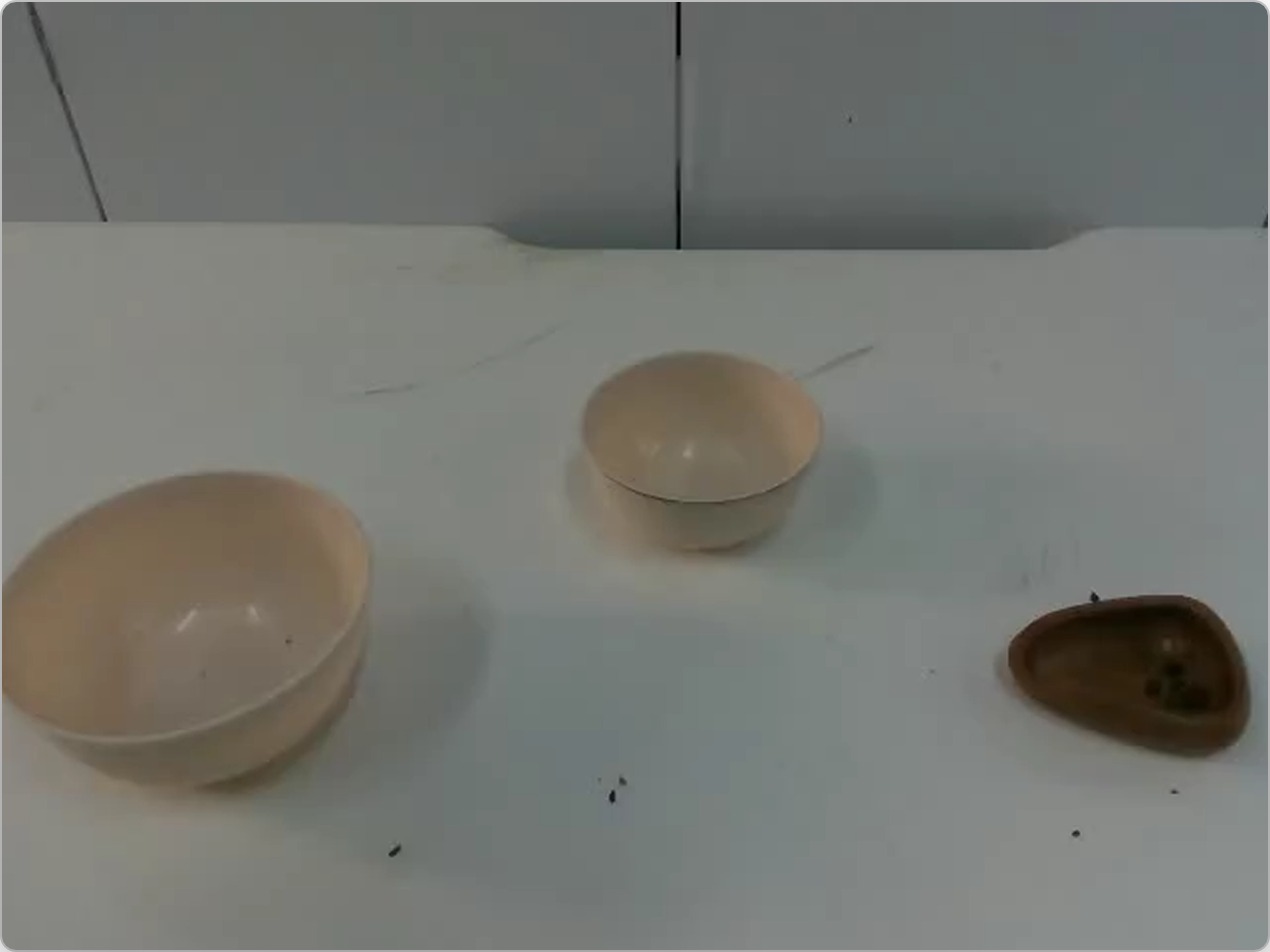} \\ 

dx12 &Pick up the bowl with the left hand, pick up the brown bean dish with the right hand, and pour the beans from the brown bean dish into the bowl.\newline & pour & color + source\_relation &
\includegraphics[width=3.0cm, height=1.7cm, keepaspectratio]{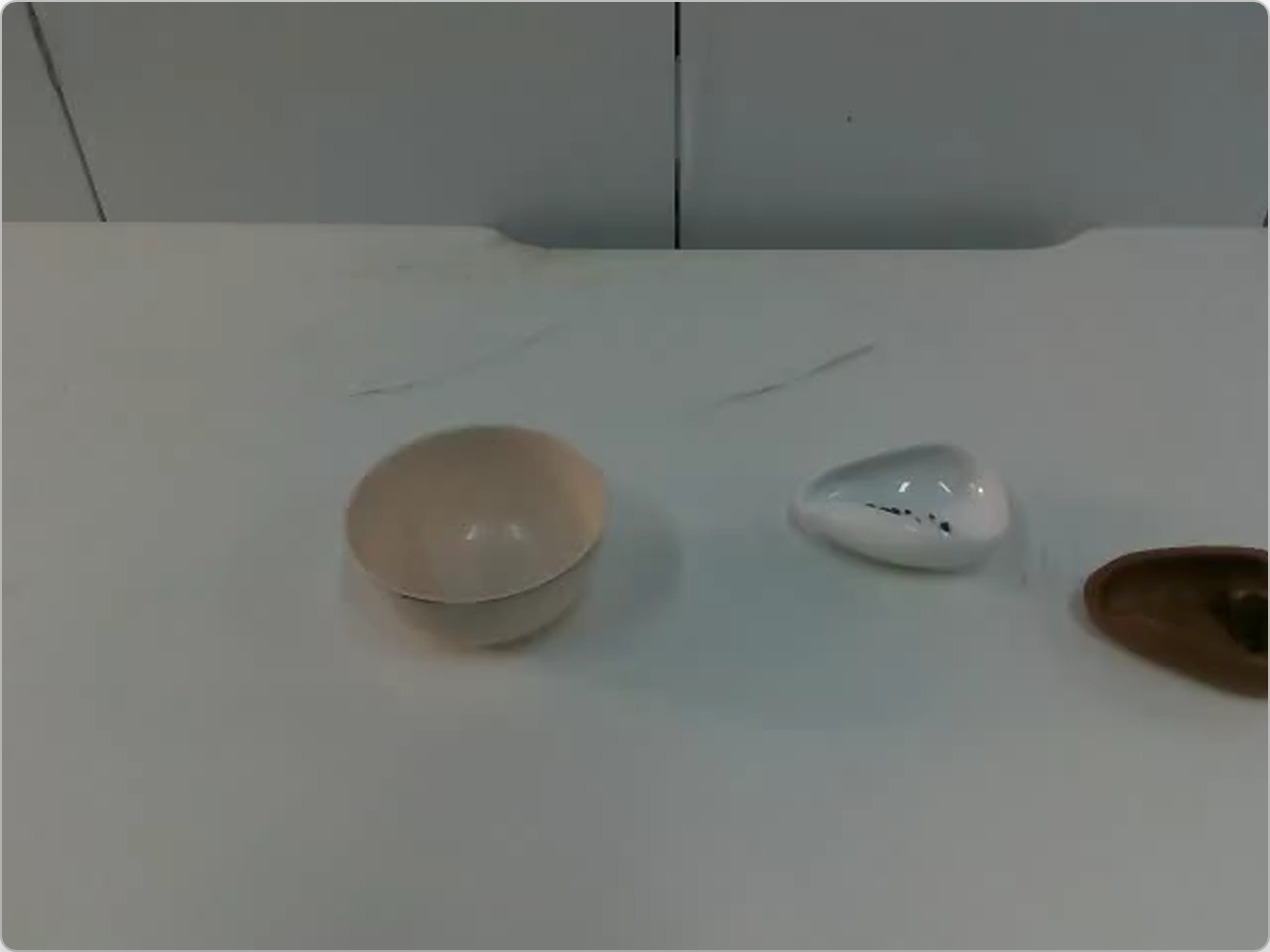} \\ 

\end{longtable}

\section{Physical Test Seed Placement}
\label{app:seed_placement}

\begin{figure}[htbp]
    \centering
    \includegraphics[width=0.95\linewidth]{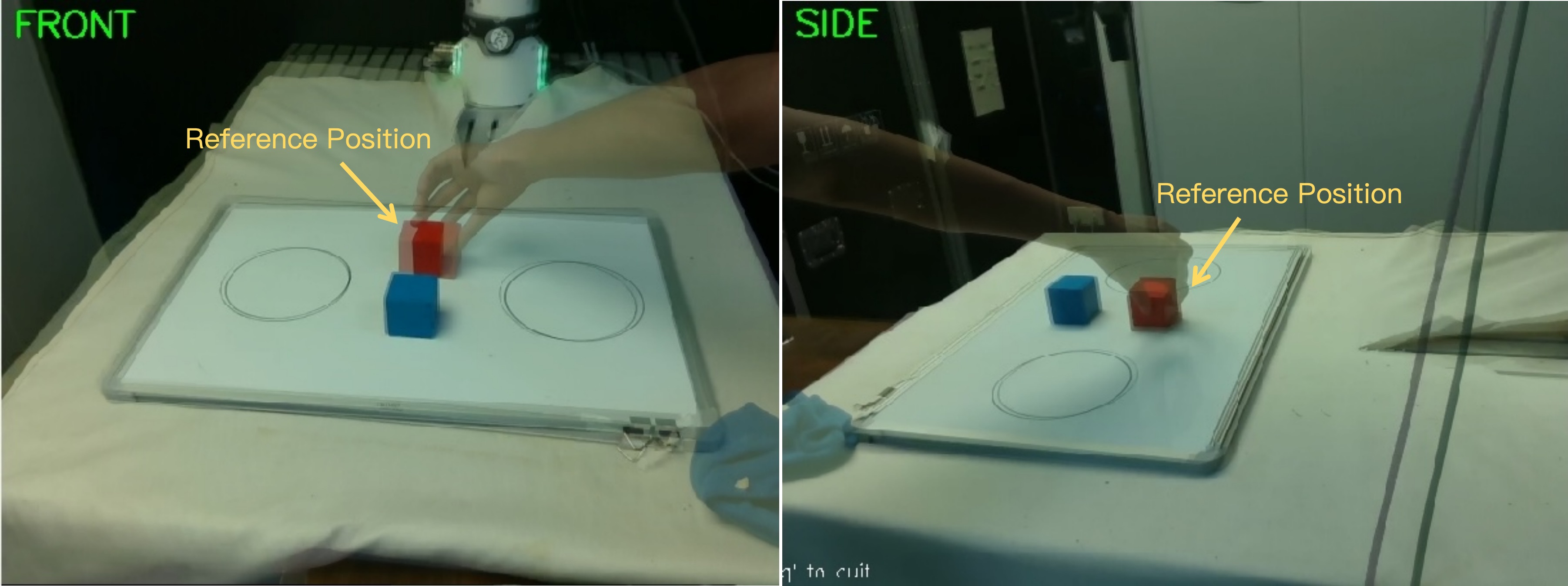}
    \caption{
        Mask-guided object placement. The reference scene of a physical test seed (overlaid as a mask) lets the operator reproduce the same initial object positions across policies and evaluation runs.
    }
    \label{fig:cam_align}
\end{figure}

Each task is evaluated under 10 fixed physical test seeds, and the same initial object placements must be reproduced when testing different policies.
To achieve this without precise measurement, the reference scene of each seed is shown as a semi-transparent mask overlaid on the operator's camera view (Figure~\ref{fig:cam_align}).
Before every rollout, the operator arranges the physical objects until they align with the mask, so that all policies are evaluated under approximately the same initial conditions.

\section{Robustness to Language and Visual Perturbations}
\label{app:perturbation}

Beyond the atomic-transfer and single-task protocols, we probe whether successful policies remain robust under simple perturbations.
We reuse each policy's atomic-transfer checkpoint without further fine-tuning, select the tasks on which it already achieves a high success rate, and re-evaluate it under two perturbation types: a visual perturbation (changing the background while keeping the scene layout) and a language perturbation (rewording the instruction while preserving its meaning).
This isolates whether a policy relies on the specific visual and linguistic surface form seen during fine-tuning rather than on the underlying skill.

\paragraph{Visual perturbations (Perturbed-Vis).}
We change the tabletop background from the default white cloth to a brown one while keeping all objects, layout, and lighting fixed. The instruction, score-point rubric, and physical test seeds remain unchanged.

\paragraph{Language perturbations (Perturbed-Lan).}
For each selected task, we author a single paraphrase that preserves the underlying skill but changes the surface form.
Atom-defining tokens (color, shape, and count words such as ``blue,'' ``triangular,'' ``exactly two''), the left/right role assignment, and the temporal ordering of multi-stage actions are all preserved verbatim.
Surface form is varied along three axes: verb synonyms within the same motor family (e.g., \emph{pick up}~$\rightarrow$~\emph{grasp/lift/hold}; \emph{place}~$\rightarrow$~\emph{drop/put}; \emph{open/close}~$\rightarrow$~\emph{pull open/push shut}; \emph{pour}~$\rightarrow$~\emph{tip}), role-assignment phrasing (\emph{with the left hand}~$\rightarrow$~\emph{using your left hand}), and clause structure (sentence merging or splitting, pronoun replacement, explicit sequencers).
The score-point rubric and physical test seeds remain unchanged. The paraphrases are listed in Table~\ref{tab:perturbed_lan_prompts}.

\begin{table}[ht]
\centering
\small
\caption{Original and language-perturbed (Perturbed-Lan) instructions for the evaluated tasks.}
\label{tab:perturbed_lan_prompts}
\begin{tabularx}{\linewidth}{l X X}
\toprule
\textbf{Task} & \textbf{Original prompt} & \textbf{Perturbed-Lan prompt} \\
\midrule
\texttt{di1} & Pick up the basket with the left hand and place the blue cube into the basket with the right hand. & Use your left hand to hold the basket, and use your right hand to drop the blue cube into it. \\
\addlinespace
\texttt{di2} & Pick up the basket with the left hand and place the triangular block into the basket with the right hand. & Hold the basket with the left hand; with the right hand, put the triangular block inside. \\
\addlinespace
\texttt{dm2} & Pick up the basket with the left hand and place the ball into the basket with the right hand. & Grasp the basket using the left hand, then drop the ball into it using the right hand. \\
\addlinespace
\texttt{dm6} & Open the drawer with the left hand, place the block into the drawer with the right hand, then close the drawer with the left hand. & Pull the drawer open with the left hand, put the block inside with the right hand, then push the drawer shut with the left hand. \\
\addlinespace
\texttt{dx1} & Pick up one red cube with each hand and place exactly two red cubes into the basket. & Use each hand to grasp one red cube, then place exactly two red cubes into the basket. \\
\addlinespace
\texttt{dx12} & Pick up the bowl with the left hand, pick up the brown bean dish with the right hand, and pour the beans from the brown bean dish into the bowl. & Lift the bowl with the left hand and the brown bean dish with the right hand, then tip the beans from the brown bean dish into the bowl. \\
\bottomrule
\end{tabularx}
\end{table}

\begin{figure}[htbp]
    \centering
    \includegraphics[width=0.95\linewidth]{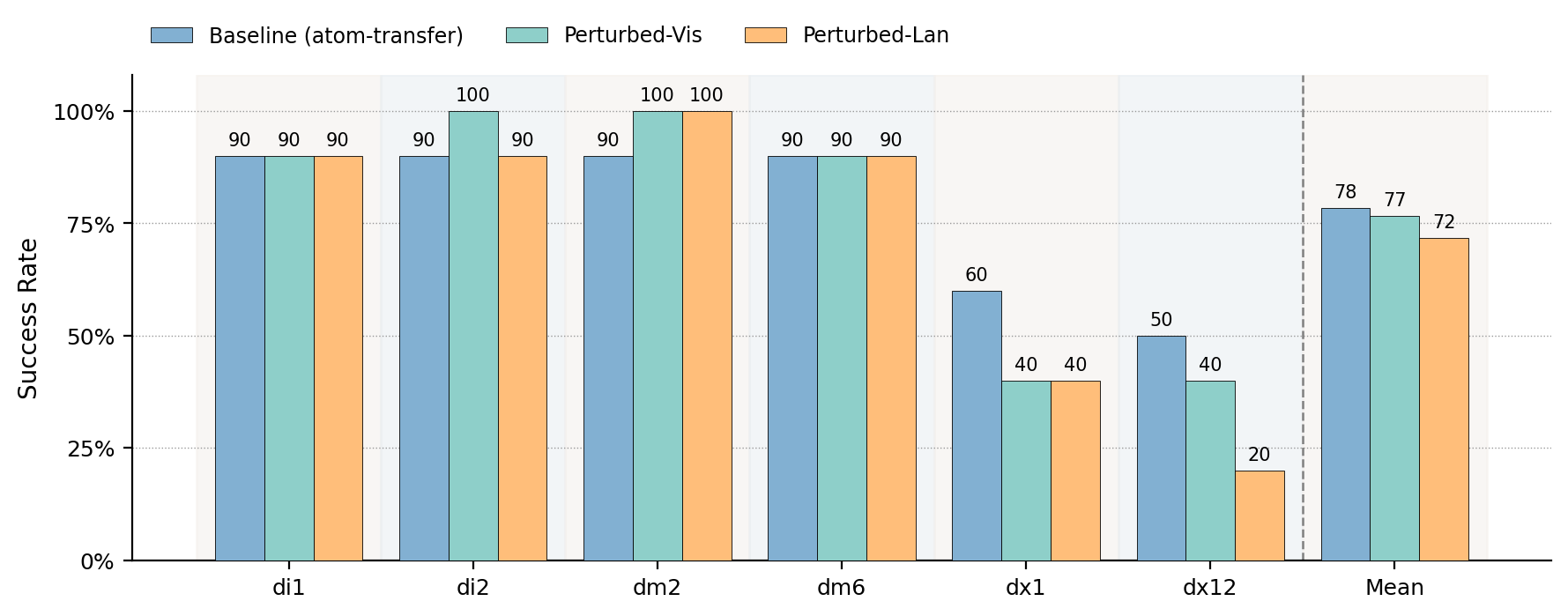}
    \caption{Pi0.5 success rate on Cobot Magic under the baseline atomic-transfer protocol, Perturbed-Vis (brown tablecloth), and Perturbed-Lan (paraphrased instruction). The right-most group is the mean SR across the six tasks.}
    \label{fig:perturbation_sr}
\end{figure}

\paragraph{Results.}
We re-evaluate the Pi0.5 atomic-transfer checkpoint on Cobot Magic under both perturbation types for six tasks --- four atomic tasks the policy already solves reliably (\texttt{di1}, \texttt{di2}, \texttt{dm2}, \texttt{dm6}) and two compositional tasks (\texttt{dx1}, \texttt{dx12}). Figure~\ref{fig:perturbation_sr} reports the per-task SR.

On the four atomic tasks, both perturbations have a minimal effect: SR stays at $90$--$100\%$ across all settings, indicating that Pi0.5 has learned the underlying atomic skills rather than overfitting to the training background or wording.
On the two compositional tasks, however, robustness drops sharply: \texttt{dx1} falls from $60\%$ to $40\%$ under both perturbations, and \texttt{dx12} falls from $50\%$ to $40\%$ (Perturbed-Vis) and $20\%$ (Perturbed-Lan).
Averaged across the six tasks, the mean SR decreases from $78.3\%$ (baseline) to $76.7\%$ under visual perturbation and to $71.7\%$ under language perturbation, with the language perturbation being the more damaging of the two.
This pattern suggests that compositional skills are markedly more brittle to surface-form changes than atomic skills, and that language paraphrasing is a stronger stress test for compositional generalization than background perturbation.

\section{SFT Configurations for Models in ATOM-Bench}
\label{app:sft_config}

We fine-tune five models using their official codebases, following the publicly released training recipes whenever possible.
In the atomic-transfer setting, each policy is jointly fine-tuned on all 15 atomic tasks, with 100 demonstrations per task and 1,500 demonstrations in total.
In the single-task setting, which is reported only for Pi0.5, we fine-tune an independent checkpoint for each atomic task using its 100 demonstrations. 

All fine-tuning runs are conducted on NVIDIA H100 GPUs. SmolVLA was fine-tuned with LoRA due to its smaller model size.
Table~\ref{tab:sft_config} lists the per-model details in atomic-transfer setting, including the base checkpoint, fine-tuning mode, optimizer, learning rate, batch size, and number of training steps.

\begin{table}[h]
\centering
\small
\caption{Per-model SFT configurations in atomic-transfer.}
\label{tab:sft_config}
\begin{tabular}{l l l l l l l}
\toprule
\textbf{Model} & \textbf{Base ckpt.} & \textbf{Mode} & \textbf{Optimizer} & \textbf{LR} & \textbf{Batch} & \textbf{Steps} \\
\midrule
\multicolumn{7}{c}{\textit{Franka Panda}} \\
\midrule
Pi0.5      & pi05\_base & full & AdamW & 2.0e-5 (cosine, warmup steps 1000) & 256 & 3 epochs \\
LingBot-VLA & lingbot-vla-4b & full & AdamW & 2.5e-5 (cosine, warmup ratio 0.25) & 128 & 3 epochs \\
GROOT N1.6 & GR00T N1.6 & full & AdamW & 1.0e-4 (cosine, warmup ratio 0.05) & 256 & 3 epochs \\
SmolVLA    & smolvla\_base & LoRA & AdamW & 1.0e-4 (cosine, warmup steps 1000) & 256 & 9 epochs \\
Motus      & Motus & full & AdamW & 2.0e-5 (linear, warmup steps 1000) & 64 & 6 epochs \\
\midrule
\multicolumn{7}{c}{\textit{Cobot Magic}} \\
\midrule
Pi0.5      & pi05\_base & full & AdamW & 2.0e-5 (cosine, warmup steps 1000) & 256 & 3 epochs \\
LingBot-VLA & lingbot-vla-4b & full & AdamW & 5.0e-5 (constant) & 256 & 3 epochs \\
GROOT N1.6 & GR00T N1.6 & full & AdamW & 1.0e-4 (cosine, warmup ratio 0.05) & 256 & 6 epochs \\
SmolVLA    & smolvla\_base & LoRA & AdamW & 1.0e-4 (cosine, warmup steps 1000) & 256 & 5 epochs \\
Motus      & Motus & full & AdamW & 2.0e-5 (linear, warmup steps 1000) & 64 & 5 epochs \\
\bottomrule
\end{tabular}
\end{table}

\section{Atom Taxonomy and Design Rationale}
\label{app:atom_taxonomy}

\subsection{Motor Atoms}
\label{app:motor_atoms}

Table~\ref{tab:app_motor_atoms} summarizes the motor atoms used in ATOM-Bench. We select these atoms from common physical operation families in prior manipulation benchmarks and motion taxonomies~\cite{yu2019metaworld,nasiriany2024robocasa,luo2025fmb,wang2025roboeval,zhang2024vlabench,paulius2019manipulation,paulius2020motion}. They provide a compact set of real-robot-executable skills for evaluating atomic skill acquisition and compositional reuse.

\begin{table*}[t]
\centering
\small
\caption{
Motor atoms in ATOM-Bench. 
Each atom summarizes a recurring physical operation family used for real-robot adaptation and held-out compositional evaluation.
}
\label{tab:app_motor_atoms}
\begin{tabularx}{\textwidth}{lXX}
\toprule
\textbf{Motor Atom} 
& \textbf{Definition} 
& \textbf{Design Rationale} \\
\midrule

\texttt{pick\_place}
& Grasp an object and place it into a target container or region.
& Foundational tabletop operation and default carrier motor for many instruction atoms. \\

\texttt{reorient}
& Change an object's pose or orientation before completing placement, such as placing a bottle upright.
& Tests object pose control beyond basic transport. \\

\texttt{push}
& Move an object along a support surface to a target region without necessarily grasping it.
& Captures canonical non-prehensile tabletop manipulation. \\

\texttt{stack}
& Place one object on top of another object.
& Tests relative-pose precision and contact stability. \\

\texttt{pour}
& Transfer granular or liquid-like contents from one container to another.
& Tests functional container use and material transfer. \\

\texttt{access}
& Open, maintain, or close an articulated container, such as a drawer, box, or lid.
& Tests articulated-object interaction and phase transitions. \\

\bottomrule
\end{tabularx}
\end{table*}

We separate \texttt{stack} from \texttt{pick\_place} because stacking requires stable contact and accurate relative pose control. We also define \texttt{access} as a functional operation family, covering tasks such as opening, using, and closing an articulated container.

\subsection{Instruction Atoms}
\label{app:instruction_atoms}

Table~\ref{tab:app_instruction_atoms} summarizes the instruction atoms used in ATOM-Bench. We select these atoms from common semantic and logical constraints in referring-expression grounding, compositional reasoning, and embodied manipulation benchmarks~\cite{kazemzadeh2014referitgame,mao2016refcocog,yu2016modelingcontext,johnson2017clevr,srivastava2022behavior,nasiriany2024robocasa,zhang2024vlabench}. Each instruction atom is evaluated through an executable robot task, usually with pick-and-place or its bimanual counterpart as the carrier action.

\begin{table*}[t]
\centering
\small
\caption{
Instruction atoms in ATOM-Bench. 
These atoms summarize recurring semantic and logical constraints in language-conditioned manipulation.
}
\label{tab:app_instruction_atoms}
\begin{tabularx}{\textwidth}{lXX}
\toprule
\textbf{Instruction Atom} 
& \textbf{Definition} 
& \textbf{Design Rationale} \\
\midrule

\texttt{color}
& Select the target object according to color.
& Basic attribute for object-level referring expressions. \\

\texttt{shape}
& Select the target object according to geometric shape.
& Stable visual attribute for tabletop object selection. \\

\texttt{size}
& Select the target object according to relative size, such as the largest or smallest object.
& Tests comparative attribute grounding. \\

\texttt{count}
& Manipulate an exact number of objects.
& Tests set-level reasoning and multi-object selection. \\

\texttt{exclusion}
& Select objects by excluding a specified attribute or category.
& Tests logical filtering over candidate objects. \\

\texttt{source\_relation}
& Select the target object according to its spatial relation to another reference object.
& Tests relation-based object reference. \\

\texttt{goal\_destination}
& Bind the action endpoint to one among multiple candidate destinations or target regions.
& Tests binding between instruction and action endpoint. \\

\bottomrule
\end{tabularx}
\end{table*}

These atoms cover three roles in a manipulation instruction. \texttt{color}, \texttt{shape}, \texttt{size}, \texttt{source\_relation}, and \texttt{exclusion} specify which object should be manipulated. \texttt{count} specifies how many objects should be manipulated. \texttt{goal\_destination} specifies where the object should be delivered.
\texttt{goal\_destination} is a goal-binding atom, which determines where the manipulated object should be delivered.

\end{document}